\documentclass{article}

\usepackage{microtype}
\usepackage{graphicx}
\usepackage{subfigure}
\usepackage{booktabs} 

\usepackage{comment}

\usepackage{notations}
\usepackage{amsmath}
\usepackage{amssymb}
\usepackage{bbm}
\usepackage{amsthm}
\usepackage{bm}
\usepackage[dvipsnames]{xcolor}

\usepackage{soul}

\DeclareMathOperator{\Rad}{Rad}
\makeatletter
\newenvironment{sqcases}{%
  \matrix@check\sqcases\env@sqcases
}{%
  \endarray\right.%
}
\def\env@sqcases{%
  \let\@ifnextchar\new@ifnextchar
  \left\lbrack
  \def\arraystretch{1.2}%
  \array{@{}l@{\quad}l@{}}%
}
\makeatother

\usepackage{hyperref}

\usepackage[accepted]{arxiv2025}

\usepackage{amsmath}
\usepackage{amssymb}
\usepackage{mathtools}
\usepackage{amsthm}

\usepackage[capitalize,noabbrev]{cleveref}

\theoremstyle{plain}
\newtheorem{theorem}{Theorem}[section]

\newtheorem{result}[theorem]{Result}

\theoremstyle{definition}

\theoremstyle{remark}
\newtheorem{remark}[theorem]{Remark}

\usepackage[textsize=tiny]{todonotes}

\arxivtitlerunning{Optimal generalisation and learning transition in extensive-width shallow neural networks near interpolation}

\begin{document}

\twocolumn[
\arxivtitle{Optimal generalisation and learning transition in \texorpdfstring{\\}{}extensive-width shallow neural networks near interpolation}

\arxivsetsymbol{equal}{*}

\begin{arxivauthorlist}
\arxivauthor{Jean Barbier}{equal,ictp}
\arxivauthor{Francesco Camilli}{equal,ictp}
\arxivauthor{Minh-Toan Nguyen}{equal,ictp}
\arxivauthor{Mauro Pastore}{equal,ictp}
\arxivauthor{Rudy Skerk}{equal,sissa}
\end{arxivauthorlist}

\arxivaffiliation{ictp}{The Abdus Salam International Centre for Theoretical Physics (ICTP), Strada Costiera 11, 34151 Trieste, Italy}
\arxivaffiliation{sissa}{International School for Advanced Studies (SISSA), Via Bonomea 265, 34136 Trieste, Italy}

\arxivcorrespondingauthor{Mauro Pastore}{mpastore@ictp.it}

\arxivkeywords{Machine Learning}

\vskip 0.3in
]



\printAffiliationsAndNotice{\arxivEqualContribution} 

\begin{abstract}
We consider a teacher-student model of supervised learning with a fully-trained two-layer neural network whose width $k$ and input dimension $d$ are large and proportional. We provide an effective theory for approximating the Bayes-optimal generalisation error of the network for any activation function in the regime of sample size $n$ scaling quadratically with the input dimension, i.e., around the interpolation threshold where the number of trainable parameters $kd+k$ and of data $n$ are comparable. Our analysis tackles generic weight distributions. We uncover a discontinuous phase transition separating a ``universal'' phase from a ``specialisation'' phase. In the first, the generalisation error is independent of the weight distribution and decays slowly with the sampling rate $n/d^2$, with the student learning only some non-linear combinations of the teacher weights. In the latter, the error is weight distribution-dependent and decays faster due to the alignment of the student towards the teacher network. We thus unveil the existence of a highly predictive solution near interpolation, which is however potentially hard to find by practical algorithms.
\end{abstract}

\section{Introduction}

Understanding the expressive power and generalisation capabilities of neural networks is not only a stimulating intellectual activity, producing surprising results that seem to defy established common sense in statistics and optimisation~\citep{bartlett2021}, but has important practical implications in cost-benefit planning whenever a model is deployed.
E.g., from a fruitful research line that spanned three decades, we now know that deep fully-connected Bayesian neural networks with $O(1)$ read-out weights and $L_2$ regularisation behave as kernel machines (the so-called Neural Network Gaussian processes, NNGPs) in the heavily overparametrised, infinite-width regime~\citep{neal1996,williams1996,lee2018gaussian,matthews2018gaussian,hanin2023infinite}, and so suffer from these models' limitations. Indeed, kernel machines infer the decision rule by first embedding the data in a fixed a priori feature space, the renowned \emph{kernel trick}, then operating linear regression/classification over the features. In this respect, they do not learn features (in the sense of statistics relevant for the decision rule) from the data, so they need larger and larger feature spaces and training sets to fit their higher order statistics~\citep{yoon1998poly,dietrich1999svm,gerace2021,bordelon2021kernel,canatar2021spectral,xiao2022precise}.

Many efforts have been devoted to studying Bayesian neural networks in a regime where they could learn a better feature map from the data. In the so-called proportional regime, when the width of the network is large and proportional to the size of the training set, recent studies showed how a limited amount of feature learning makes the network equivalent to optimally regularised kernels~\citep{li2021,pacelli2023,camilli2023fundamental,cui2023bayes,baglioni2024}. This effect could be a consequence of the fully-connected architecture, as, e.g., convolutional neural networks learn more informative features in this regime~\citep{naveh2021,seroussi2023,aiudi2023,bassetti2024}. 
Another scenario recently proposed is the mean-field scaling, i.e., 
when the read-out weights are small:
in this case too a Bayesian network can learn features in the proportional regime~\citep{rubin2024unified,vanmeegen2024}.

In this paper, we consider instead the generalisation performance of a fully-connected two-layer Bayesian network of extensive width trained end-to-end near the interpolation threshold, when the sample size $n$ is scaling like the number of trainable parameters: for input dimension $d$ and width $k$, both large and proportional, $n= \Theta(d^2) =\Theta(kd)$. We consider i.i.d. standard Gaussian input vectors with labels generated by a teacher network with matching architecture, in order to study the Bayes-optimal performance of the model. Therefore, the results we report not only enable to approximate the generalisation error of Bayesian students, but can serve as benchmark for the performance of \emph{any} model trained on the same dataset. The activation of the hidden layer is only required to admit a decomposition in the basis of Hermite polynomials.

\paragraph{Our contributions and related works}
The aforementioned setting is related to the recent paper \citet{maillard2024bayes}, with however two major differences: said work considers only Gaussian distributed weights and quadratic activation. These hypotheses allow numerous simplifications for the analysis, exploited in a series of works \cite{du2018power,soltanolkotabi2018theoretical,venturi2019spurious,sarao2020optimization,gamarnik2024stationary,martin2024impact,arjevani2025geometry}. Thanks to this, \citet{maillard2024bayes} map the learning task onto a generalised \emph{linear} model (GLM) where the goal is to infer a Wishart matrix from linear observations, which is analysable using known results on the GLM \cite{barbier2019glm} and matrix denoising \cite{barbier2022statistical,maillard2022perturbative,matrix_inference_Barbier,semerjian2024}.

Our main contribution is a general statistical mechanics framework for characterising the prediction performance of shallow Bayesian neural networks, able to handle arbitrary activation functions and different distributions of i.i.d. weights. In particular, we show that there is not always universality in the teacher weights,
and that the prior over the inner weights and the choice of activation function play an important role in learning. Our theory draws a rich picture with two phases separated by a learning phase transition when tuning the sample rate $\alpha= n/d^2$: 

${\mathbf{(i)}}$ For low $\alpha$, feature learning occurs only because the student tunes its weights to match non-linear combinations of the teacher's ones, rather than aligning to those weights themselves. This phase is \emph{universal} in the law of the i.i.d. teacher inner weights: our numerics obtained with binary inner weights match well the theory valid for Gaussian ones.  

\vspace{-6pt}
${\mathbf{(ii)}}$ For high enough $\alpha$, a \emph{specialisation transition} occurs, where the student can align its weights to the actual teacher ones. We predict this transition to occur for binary inner weights and generic activation, or for Gaussian inner weights and more-than-quadratic activation; in general, we write a criterion to assess if the transition will occur at given prior and activation function. We provide a description of the two phases and identify the relevant order parameters (sufficient statistics) needed to deduce the generalisation error through scalar systems of equations.

The picture that emerges is closely connected to recent findings in the context of extensive-rank matrix denoising \cite{barbier2024phase}. In this model similar phases were identified, with one being universal in the signal prior law and the other not, with the estimator ``synchronising'' with the hidden signal beyond the transition. We believe that this picture and the one found in the present paper are not just similar, but are actually both a manifestation of the same fundamental mechanism in matrix inference/learning.

From a technical point of view, our derivation is based on a Gaussian ansatz on the replicated post-activations of the hidden layer, which generalises Conjecture 3.1 of~\citet{cui2023bayes}, where it is specialised to the case of linearly many data ($n=\Theta(d)$). To obtain this generalisation, we write the kernel arising from the covariance of the aforementioned post-activations as an infinite series of scalar order parameters derived from the expansion of the activation function in the Hermite basis, following an approach recently devised in~\citet{aguirre2024RF} in the context of the random features model (see also \citet{hu2024asymptotics} and \citet{10.1214/20-AOS1990}). Another key ingredient of our analysis is a generalisation of an ansatz used in the replica method by \citet{sakata2013} for dictionary learning.

From the algorithmic perspective, 
we adapt to generic activation the GAMP-RIE (generalised approximate message-passing with rotational invariant estimator), introduced in \citet{maillard2024bayes} for the special case of quadratic activation. The resulting algorithm described in Appendix~\ref{app:glm}, which \emph{cannot} find the specialisation solution (where it exists) by construction, nevertheless matches the prediction performance associated with the universal branch of our theory for all $\alpha$. As a side investigation, we show empirically that finding the specialisation solution with popular algorithms is potentially hard for some target functions: the algorithms we tested either fail to find it and instead get stuck in a sub-optimal glassy phase (Metropolis-Hastings sampling for the case of binary prior), or may find it but in a training time increasing exponentially with $d$ (ADAM and Hamiltonian Monte Carlo for the case of Gaussian prior). For specific choices of the distribution of the read-out weights, the evidence of hardness is less conclusive and requires further investigation.
Given this observation, it would be interesting to settle whether GAMP-RIE has the best prediction performance achievable by a polynomial-time learner when $n=\Theta(d^2)$.

\section{Teacher-student setting}
We consider supervised learning with a shallow neural network in the classical teacher-student setup. The data-generating model, i.e., the teacher, is thus a two-layer neural network itself, with read-out weights $\bv^0\in\R^k$ and internal weights $\bW^0\in\R^{k\times d}$, drawn entrywise i.i.d. from $P_v^0$ and $P^0_W$, respectively; we assume $P^0_W$ to be centred and with unit variance. The whole set of parameters of the teacher is denoted $\btheta^0=(\bv^0,\bW^0)$.
The inputs are i.i.d. standard Gaussian vectors $\bx_\mu\in\R^{d}$ for $\mu=1,\dots, n$. The responses $y_\mu$ are possibly random outputs of a kernel $P^0_{\rm out}$:
\begin{align} \label{eq:teacher}
    y_\mu\sim P^0_{\rm out}\big(\cdot\mid \lambda^0_\mu\big), \ \ \lambda^0_\mu(\btheta^0):=\frac{\bv^{0\intercal}}{\sqrt{k}}\sigma\Big(\frac{\bW^0\bx_\mu}{\sqrt{d}}\Big).
\end{align}
The kernel can be stochastic or model a deterministic rule if taking $P^0_{\rm out}(y|\lambda)=\delta(y-\tau^0(\lambda))$ for some outer non-linearity $\tau^0$. The activation function $\sigma$ is applied entrywise to vectors and admits an expansion in Hermite polynomials with Hermite coefficients  $(\mu_\ell)_{\ell\geq0}$ (see Appendix~\ref{app:hermite}): $ \sigma(x) = \sum_{\ell \ge 0 } \frac{\mu_\ell}{\ell !}\He_\ell (x)$. In the main we assume it has vanishing 0th Hermite coefficient 
in order to simplify the presentation, i.e., that it is centred $\EE_{z\sim\mathcal{N}(0,1)}\sigma(z)=0$; in Appendix~\ref{app:non-centered} we relax this assumption.
The input/output pairs $\mathcal{D}=\{(\bx_\mu,y_\mu)\}_{\mu \leq n}$ forms the training set for a student network with matching architecture. 

The Bayesian student learns via the posterior distribution of the weights $\btheta=(\bv,\bW)$ given the training data, defined by
\begin{align*}
&dP(\btheta \mid\mathcal{D}):=\frac1{\mathcal{Z}}dP_v(\bv)dP_W(\bW)\prod_{\mu=1}^nP_{\rm out}\big(y_\mu\mid \lambda_\mu(\btheta)\big) 
\end{align*}
with post-activation $\lambda_\mu(\btheta):=k^{-1/2}\bv^{\intercal}\sigma(d^{-1/2}{\bW\bx_\mu})$ 
and $P_v,P_W$ are the priors assumed by the student, which are also fully factorised. From now on, we focus on the Bayes-optimal case $P_W=P_W^0,P_v=P_v^0,P_{\rm out}=P_{\rm out}^0$, but the approach can be extended to account for a mismatch.

We aim at evaluating the average generalisation error of the student. Let $(\bx_{\rm test}, y_{\rm test}\sim P_{\rm out}(\,\cdot\mid \lambda^0_{\rm test}))$ be a fresh sample drawn using the teacher independently from $\mathcal{D}$, where $\lambda_{\rm test}^0$ is defined as in Eq.~\eqref{eq:teacher} with $\bx_\mu$ replaced by $\bx_{\rm test}$. Given any prediction function $\tau$, the Bayes estimator for the test response reads $\hat{y}^{\rm \tau}(\bx_{\rm test},\calD)
    :=\langle \tau(\lambda_{\rm test}(\btheta)) \rangle$, where the expectation $\langle \,\cdot\, \rangle :=\EE[ \,\cdot \mid \mathcal{D}]$ is w.r.t. the posterior $dP(\btheta \mid\mathcal{D})$. Then, for a
performance measure $\mathcal{C}:\mathbb{R}\times \mathbb{R}\mapsto\mathbb{R}_{\ge 0}$ the Bayes generalisation error is
\begin{align}
\varepsilon^{\mathcal{C},\tau}:=\EE_{\btheta^0,\calD,\bx_{\rm test},y_{\rm test}}\mathcal{C}\big(y_{\rm test}, \langle \tau(\lambda_{\rm test}(\btheta)) \rangle\big).\label{eq:Bayes_error_def}
\end{align}
An important case is the square loss $\mathcal{C}(y,\hat y)=(y-\hat y)^2$ with the choice $\tau(\lambda)=\int dy\, y\, P_{\rm out}( y\mid \lambda)=:\EE[y\mid \lambda]$. 
The Bayes-optimal mean-square generalisation error follows:
\begin{align}\label{eq:gen_error_def}
    \varepsilon^{\rm opt} &:= \EE_{\btheta^0,\calD,\bx_{\rm test},y_{\rm test}}\big(y_{\rm test} - \big\langle\EE[y\mid \lambda_{\rm test}(\btheta)]\big\rangle\big)^2.
\end{align}
In the text we consider, as main example, linear read-out with Gaussian label noise,
\begin{equation}
\label{eq:Pout_Delta}
    P_{\rm out}(y\mid\lambda) =\frac{\exp(-\frac1{2\Delta}(y-\lambda)^2)}{\sqrt{2\pi\Delta}}.
\end{equation}
In this case, the generalisation error $\varepsilon^{\rm opt}$ takes a simpler form for numerical evaluation than \eqref{eq:gen_error_def}, thanks to the concentration of ``overlaps'' entering it, see Appendix~\ref{app:gen_err}.

In order to theoretically access $\varepsilon^{\mathcal{C},\tau},\varepsilon^{\rm opt}$ and other relevant quantities, one can tackle the computation of the average log-partition function, or ``free entropy'' in statistical mechanics vocabulary: $f_n:=\EE\ln\mathcal{Z}(\mathcal{D})/n$, where $\mathcal{Z}=\mathcal{Z}(\mathcal{D})$ is the normalisation of the posterior, and the expectation is w.r.t. the training data $\mathcal{D}$ and $\btheta^0$. The mutual information between teacher weights and the data is related to the free entropy $f_n$, see Appendix~\ref{app:mutual_info}. E.g., in the case of linear read-out with Gaussian label noise we have $I(\btheta^0;\mathcal{D})/(kd) = -\frac{\alpha}{\gamma} f_n - \frac{\alpha}{2\gamma} \ln(2\pi e \Delta)$. Considering the mutual information per parameter allows us to interpret $\alpha$ as a sort of signal-to-noise ratio, s.t. the mutual information defined in this way increases with it.

We consider the challenging extensive-width regime with quadratically many samples, i.e., a large size limit 
\begin{align}
 d,k,n\to+\infty \quad \text{with} \quad  \frac{k}{d}\to\gamma, \quad \frac{n}{d^2}\to\alpha  .\label{thermolim}
\end{align}
We denote this joint $d,n,k$ limit with these rates by $\widetilde\lim$.

\emph{Notations:} Bold is for vectors and matrices, $d$ is the input dimension, $k$ the width of the hidden layer, $n$ the size of the training set $\mathcal{D}$, with asymptotic ratios $k/d\to \gamma$ and $n/d^2\to \alpha$, 
$s$ will be the number of replicas in the replica method, $\bA^{\circ \ell}$ is the Hadamard power, i.e., $(\bA^{\circ \ell})_{ij} = A_{ij}^\ell$, $(\bv)$ is the diagonal matrix $\diag(\bv)$,  $(\mu_\ell)$ are the Hermite coefficients of the activation $ \sigma(x) = \sum_{\ell \ge 0} \frac{\mu_\ell}{\ell !}\He_\ell (x)$.

\section{Results: learning transition and Bayes generalisation error}\label{sec:result}

\paragraph{Learning transition}

Our first result is a tractable heuristic formula for the location of the learning transition, based on a free entropy comparison.
To state it, let us first  introduce
\begin{equation}
\begin{aligned}
    \label{eq:covariance_vars}
        &q_K(q_2,q_W) := \mu_1^2+ \frac{\mu_2^2}{2} q_2 + g(q_W)\\
        &r_2:= 1 + \gamma (\E v_1^0)^2\\
        &r_K := \mu_1^2+ \frac{\mu_2^2}{2}r_2 + g(1)
\end{aligned} 
    \end{equation}
    with $g(x) := \sum_{\ell = 3}^\infty \frac{\mu_{\ell}^2}{\ell !} x^\ell$ (see also \eqref{eq:g_func} for a more explicit expression of it), and the auxiliary potentials
    \begin{equation*}
    \begin{aligned}
        &\psi_{P_W}(\hat q_W) := \mathbb{E}_{w^0,\xi} \ln\mathbb{E}_w \,e^{-\frac{\hat{q}_W}{2}w^2 + \hat{q}_W w^0 w + \sqrt{\hat{q}_W} \xi w}\\
        &\psi_{P_{\rm out}}(q_K,r_K) := \int dy\, \mathbb{E}_{\xi,u^0}P_{\rm out}(y\mid \xi\sqrt{q_K}\\
        &\ +u^0\sqrt{r_K-q_K})
        \ln \mathbb{E}_{u}P_{\rm out}(y\mid \xi\sqrt{q_K}+u\sqrt{r_K-q_K})
    \end{aligned}
    \end{equation*}
    where $w^0,w \sim P_W$ and $\xi,u_0,u \sim \calN(0,1)$. Moreover, let
    \begin{align*}
    \iota(\hat q_2):=\frac{1}{8}+\frac{1}{2} \int \ln|x-y|d\mu_{\bY(\hat q_2)}(x)d\mu_{\bY(\hat q_2)}(y),
    \end{align*} 
    where $\mu_{\bY(\hat q_2)}$ is the asymptotic spectral density of the observation matrix in the denoising problem of the matrix $\bS^0:=\bW^{0\intercal}(\bv^0)\bW^0$ given $\bY(\hat q_2)=\sqrt{\hat q_2/kd}\,\bS^0+\bZ$, with $\bZ$ a standard GOE matrix (a symmetric matrix whose upper triangular part has i.i.d. entries from $\mathcal{N}(0,(1+\delta_{ij})/d)$).
\begin{result}[Learning transition]
\label{res:free_entropy}
For any $\alpha$ and $\gamma$, under the scaling limit \eqref{thermolim} we predict a learning phase transition located 
at
\begin{align}
    \alpha_{\rm sp}(\gamma):=\min\big\{\alpha : f_{\rm sp}(\alpha,\gamma)\ge f_{\rm uni}(\alpha,\gamma)\big\},
    \label{eq:alpha_c}
\end{align}
    where $f_{ {\rm uni} / {\rm sp}}$, the free entropies per datum associated with, respectively, the universal and specialisation solutions, are
    \begin{equation*}
        \begin{aligned}
           &f_{\rm uni}:=\underset{q_2,\hat{q}_2}{\rm extr}\Big\{ \psi_{P_{\rm{out}}}\!(q_K(q_2,0),r_K)\!+\!\frac{\hat q_2(r_2\!-\!q_2)}{4\alpha}\!-\!\frac{\iota(\hat q_2)}{\alpha}\Big\}\\
&f_{\rm sp} := \underset{q_2,\hat{q}_2,q_W,\hat{q}_W}{{\rm extr}} \Big\{ \frac{\gamma}{\alpha} \psi_{P_W}(\hat q_W) 
    \!+\!\psi_{P_{{\rm out}}}(q_K(q_2,q_W),r_K) \\
    &\,\,
    -\frac{\gamma}{2\alpha}q_W \hat{q}_W 
    + \frac{(r_2-q_2) \hat{q}_2}{4\alpha}
    - \frac{1}{4\alpha}\ln [ 1+\hat q_2   (1 - q_W^2) ]
    \Big\} .
        \end{aligned}
    \end{equation*}
    The extremisation operation ${\rm extr}\{\cdots\}$ selects the solution of $\nabla\{\cdots\}=\mathbf{0}$ which maximizes $\{\cdots\}$.
\end{result}

The extremisation needed to obtain $f_{\rm uni}$, $f_{\rm sp}$ yields the two systems of equations~\eqref{eq:Suni},~\eqref{FP_equations_generic_ch} that can be solved numerically by standard methods (see the provided code).

For quadratic activation, the transition occurs if the distribution of the teacher and student's weights is discrete. For more-than-quadratic activations, we predict the transition to occur even for Gaussian weights (see Fig.~\ref{fig:gen_error_gauss} and App.~\ref{sec:gaussian_prior}). In this article, we report both the cases where the weights are binary $\pm 1$ and Gaussian. Then, $\alpha < \alpha_{\rm sp}$ corresponds to the \emph{universal phase}, where $f_{\rm uni}$ obtained from the Gaussian weights theory approximates well the log-partition function of the model, independently on the choice of the prior over the inner weights. Instead, $\alpha > \alpha_{\rm sp}$ is the \emph{specialisation phase} where $f_{\rm sp}$ is a better approximation. We will discuss their differences.

\paragraph{Bayes generalisation error} 

Another main result is a heuristic formula for the generalisation error. 
By assuming that the joint law of $(\lambda(\btheta^a,\bx_{\rm test}))_{a\ge 0}=(\lambda^a)_{a\ge 0}$ for a common test input $\bx_{\rm test}  \notin \mathcal{D}$, where $(\btheta^a)_{a\ge 1}$ are conditionally i.i.d. samples from the posterior $dP(\,\cdot \mid\mathcal{D})$ and $\btheta^0$ is the teacher, is a centred Gaussian distribution, our framework predicts its covariance. Our approximation for the Bayes error in the limit $\widetilde \lim$ follows.
\begin{result}[Covariance of the post-activations and Bayes generalisation error]
\label{res:gen_error}
For $\alpha<\alpha_{\rm sp}(\gamma)$ let $q_K^*=q_K(q_2^*,0)$ where $(q_2^*,\hat q_2^*)$ are the extremizers of $f_{\rm uni}$ (yielding its maximum value). For $\alpha>\alpha_{\rm sp}(\gamma)$ let $q_K^*=q_K(q_2^*,q_W^*)$ where $(q_W^*,\hat q_W^*,q_2^*,\hat q_2^*)$ are the extremizers of $f_{\rm sp}$ (yielding again its maximum value). Assuming joint Gaussianity of the  post-activations $(\lambda^a)_{a\ge 0}$, in the limit $\widetilde \lim$ their mean is zero and their covariance is predicted to be $\EE \lambda^a \lambda^{b} = q_K^*+(r_K-q_K^*)\delta_{ab}$, see App.~\ref{app:gen_err}.

Assume $\mathcal{C}$ has series expansion $\mathcal{C}(y,\tau)=\sum_{i\ge 0} c_i(y)\tau^i$. The limiting Bayes generalisation error is approximated by
\begin{align}\label{eq:Bayes_error_result}
    \widetilde \lim\,\varepsilon^{\mathcal{C},\tau}\!\!=\EE_{(\lambda^a)}\EE_{y_{\rm test}|\lambda^0} \sum_{i\ge 0} c_i(y_{\rm test}(\lambda^0))\prod_{a=1}^i\tau(\lambda^a).
\end{align}
In particular, letting $\EE[\,\cdot\mid \lambda]=\int dy \,(\,\cdot\,)\, P_{\rm out}(y\mid \lambda)$, the limiting Bayes-optimal mean-square generalisation error is
  \begin{equation}
      \widetilde \lim\, \varepsilon^{\rm opt} =  \EE_{\lambda^0,\lambda}\big(\EE[y^2\mid \lambda^0] - \EE[y\mid \lambda^0]\EE[y\mid \lambda] \big).
  \label{eq:gen_err_result}
  \end{equation}
\end{result}
We will interpret the variables $q_2^*$, $q_W^*$ as ``overlaps'' between combinations of teacher and student's weights. This result assumed that $\mu_0=0$; see App.~\ref{app:non-centered} if this is not the case.

\section{Numerical experiments}

Results~\ref{res:free_entropy} and \ref{res:gen_error} together provide an effective theory for the generalisation capabilities of a Bayesian shallow network with generic activation. 
Our analysis
pinpoints the presence of two distinct phases: a universal one, where the prior on the inner weights is irrelevant (only its first two moments matter), and a specialisation one where the generalisation error becomes prior-dependent.

\begin{figure}[t!!]
\begin{center}
\centerline{\includegraphics[width=1\linewidth,trim={0 0 0 0.2cm},clip]{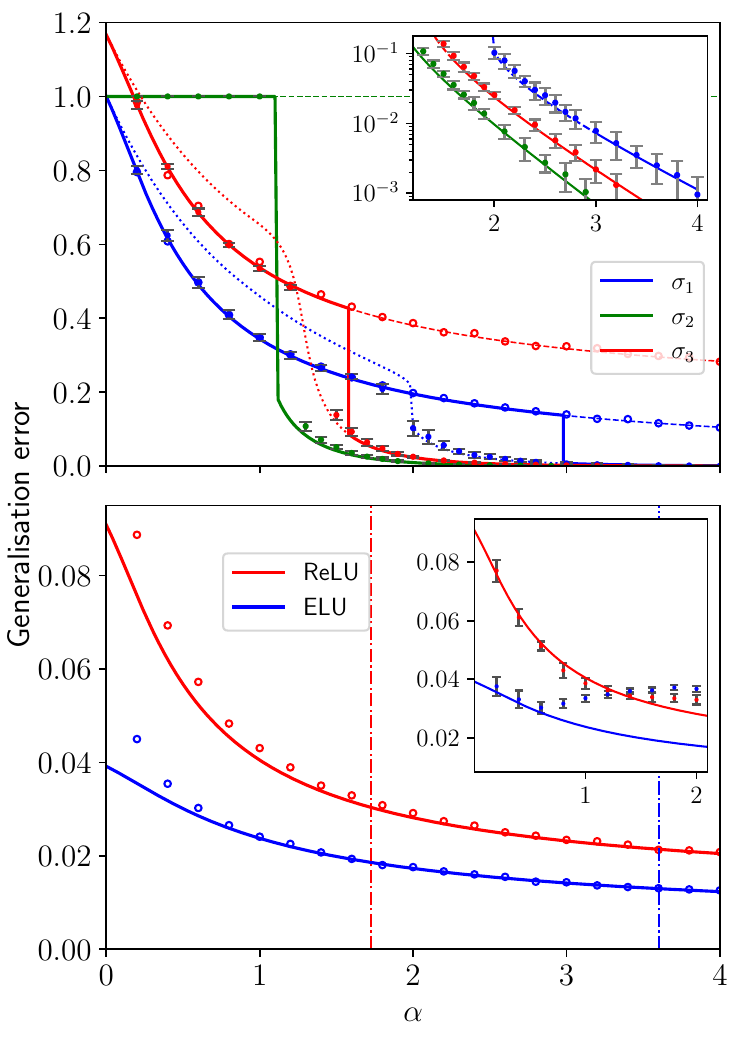}}
\vspace{-10pt}
    \caption{\textbf{Top:} Theoretical prediction (solid curves) of the Bayes-optimal mean-square generalisation error for \emph{binary inner weights} and polynomial activations: $\sigma_1(x) = \He_2(x)/\sqrt 2$, $\sigma_2(x) = \He_3(x)/\sqrt 6$, $\sigma_3(x) = \He_2(x)/\sqrt 2 + \He_3(x)/6$, with $\gamma = 0.5$, $d=150$, Gaussian label noise with $\Delta=1.25$, and fixed read-outs $\bv=\bv^0 = \mathbf{1}$. Dots are obtained by plugging the overlaps obtained from MCMC into Eq.~\eqref{eq:simple_gen_error_for_numerics} in App.~\ref{app:gen_err}, which neglects some finite size effects by assuming Eq.~\eqref{Q=QwQv} (which is validated numerically, see Fig.~\ref{fig:overlaps}). Circles are the error of GAMP-RIE \citep{maillard2024bayes} extended to generic activation, obtained by plugging estimator \eqref{eq:output_GAMP_RIE} in \eqref{eq:gen_error_def}.
    Points for GAMP-RIE and MCMC are averaged over 16 data instances. Error bars for MCMC are the standard deviation over instances (omitted for GAMP-RIE, but of the same order).
    The specialisation transitions (vertical lines) are identified comparing the free entropy of the two phases, see Eq.~\eqref{eq:alpha_c} and App.~\ref{app:mutual_info}. Dashed and dotted lines denote, respectively, universal and specialisation branches where they are metastable. The MCMC points follow the specialisation curve before the transition as they are obtained with informative initialisation, converging to the specialisation solution once it becomes accessible. The inset zooms on the specialisation phase. \textbf{Bottom:} All parameters as above, except $\Delta=0.1$. Generalisation error of the universal branch for popular activations, which possibly corresponds to the algorithmically tractable performance for binary prior. The dashed lines are the specialisation transition. The MCMC points (inset) are obtained using \eqref{eq:general_finitesize_gen_error_noapprox}, to account for lack of equilibration due to glassiness, which prevents using \eqref{eq:simple_gen_error_for_numerics}. Even in the possibly glassy region, the GAMP-RIE attains the universal branch performance.}
    \label{fig:gen_errors_univ_spec}
\end{center}
\vskip -0.3in
\end{figure}

Before explaining our theory, let us compare its predictions with simulations. In Fig.~\ref{fig:gen_errors_univ_spec}, we report the theoretical curves from Result~\ref{res:gen_error}, focusing on the optimal mean-square generalisation error, for networks with $\pm1$ inner weights and Gaussian output channel~\eqref{eq:Pout_Delta}, 
for different activation functions in the hidden layer. The numerical points are of two kinds: the dots, obtained from Monte Carlo Metropolis–Hastings sampling of the posterior distribution of the model's weights, and the circles, obtained from an extension of the GAMP-RIE of~\citet{maillard2024bayes} to account for generic activation (see App.~\ref{app:glm}). The universal phase, where the error of the student with binary inner weights matches the one of a student with $P_W= \calN(0,1)$, is superseded at $\alpha_{\rm sp}$ (obtained from Result~\ref{res:free_entropy}, Eq.~\eqref{eq:alpha_c}; see also App.~\ref{app:mutual_info}, Fig.~\ref{fig:MIcrossing} and Fig.~\ref{fig:gaussian} left) by a specialisation phase where the student's inner weights start aligning with the teacher's ones. This transition is different in nature w.r.t. the perfect recovery threshold identified in~\citet{maillard2024bayes}, which is the point where the student with Gaussian weights learns perfectly $\bW^{0\intercal}(\bv^0)\bW^0$ (but \emph{not} $(\bW^{0\intercal},\bv^0)$) and thus attains perfect generalisation in the case of purely quadratic activation and zero label noise ($\Delta\to 0$ in Eq.~\eqref{eq:Pout_Delta}). In Fig.~\ref{fig:gen_errors_univ_spec},
we split the case of polynomial activations (top panel) and the one of $\relu$, $\elu$ (defined in Table~\ref{tab:Hermite}) for illustration purposes: in the latter case, for low values of $\Delta$, MCMC with informative initialisation remains stuck without equilibrating, while for higher values of $\Delta$, $\alpha_{\rm sp}$ is too high to be sampled with our implementation.
The remarkable agreement between theoretical curves and experimental points in both phases supports the assumptions used in Sec.~\ref{sec:theory}.

\begin{figure}[t]
\begin{center}
\centerline{\includegraphics[width=0.9\linewidth]{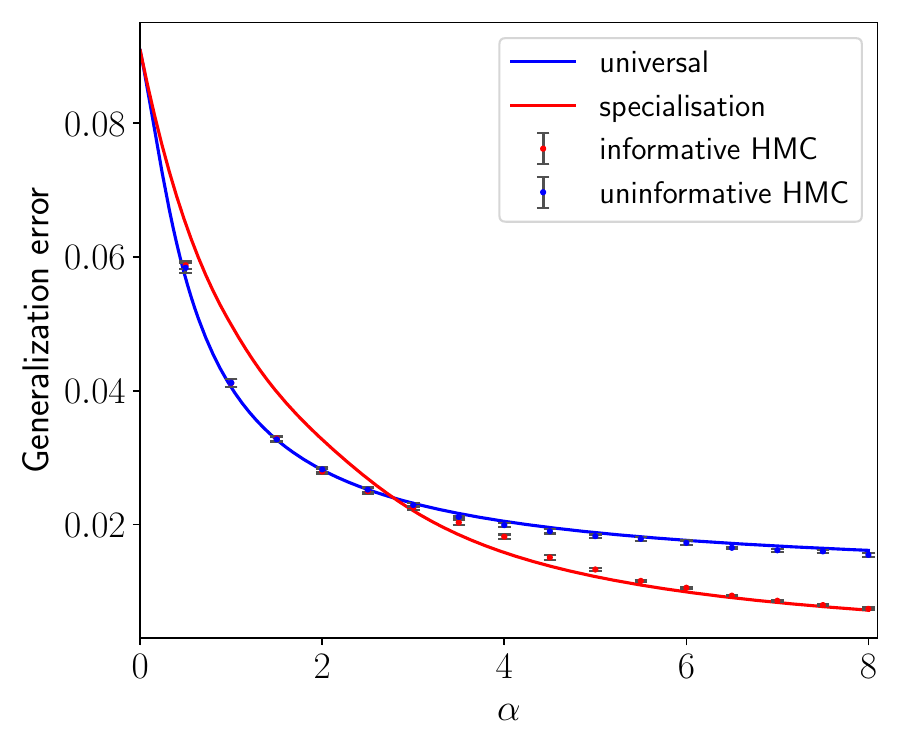}}
\vspace{-10pt}
    \caption{Theoretical prediction (solid curves: blue for the universal branch, red for the specialisation one) of the Bayes-optimal mean-square generalisation error for \emph{Gaussian inner weights} and ReLU activation, $d=150, \gamma=0.5, \Delta=0.1$, fixed read-outs $\bv=\bv^0 = \mathbf{1}$. Here the specialisation transition is at $\alpha_{\rm sp} \approx 5.54$. The numerical points are obtained with Hamiltonian Monte Carlo with informative/uninformative initialisation. Each point has been obtained by averaging over 9 instances of the training set. The generalisation error for a given training set is evaluated  by $\frac{1}{2}  \EE_{\bx_{\rm test} \sim \mathcal N(0, I_d)} (\lambda_{\rm test}(\btheta^a)-\lambda_{\rm test}(\btheta^0))^2 $, using a single sample $\btheta^a=(\bv,\bW^a)$ from the posterior; the average over $\bx_{\rm test}$ is computed empirically from $10^4$ i.i.d. test samples. We assume this quantity to be $(\varepsilon^{\rm Gibbs} - \Delta)/2 = \varepsilon^{\rm opt} - \Delta$, where the Gibbs error $\varepsilon^{\rm Gibbs}$ is defined in Eq.~\eqref{eq:Gibbs_error} in App.~\ref{app:gen_err}, and its relationship with the Bayes error is reported in Eq.~\eqref{eq:Gibbs_v_Bayes_error}. To use this formula, we are assuming: (i) concentration of the Gibbs error w.r.t. the posterior distribution, in order to evaluate it from a single sample per instance; (ii) validity of the Nishimori identities for the empirical distribution sampled by HMC, when sampling configurations corresponding to both the universal solution and the specialisation one; these assumptions are validated by the agreement with the theoretical curves.}
    \label{fig:gen_error_gauss}
\end{center}
\vskip -0.2in
\end{figure}

An interesting effect our theory predicts is that, for Gaussian inner weights, specialisation does occur, but only if the activation function contains Hermite polynomials of degree higher than two: for a quadratic activation only the universal phase is present (an observation that matches the results of~\citet{maillard2024bayes}), as the free entropy of the specialisation branch is always lower, and thus never selected by criterion~\eqref{eq:alpha_c}. On the contrary, with more-than-quadratic activations and high-enough $\alpha$, the Bayes-optimal student is able to synchronise even with a Gaussian teacher, by somehow realising that the higher order terms of its Hermite decomposition are not label noise but they are informative on the decision rule.  We report in Fig.~\ref{fig:gen_error_gauss} the case of $\relu$ activation and Gaussian prior, comparing our theory with Hamiltonian Monte Carlo (HMC) simulations: the agreement validates our approach in this setting too. We dedicate App.~\ref{sec:gaussian_prior} to comment more on the case of Gaussian prior.

Even when dominating the posterior measure, we observe in simulations that the specialisation solution can be algorithmically hard to reach. With a discrete distribution of read-outs (such as $P_v=\delta_1$ or Rademacher), simulations for binary inner weights exhibit it only when sampling with informative initialisation (i.e., the MCMC runs to sample $\btheta$ are initialised in the vicinity of the teacher's $\btheta^0$). Moreover, even in cases where algorithms (such as ADAM or HMC for Gaussian inner weights) are able to find the specialisation solution, they manage to do so only after a training time increasing exponentially with $d$, and for relatively small values of the label noise $\Delta$: Fig.~\ref{fig:gen_error_gauss} reports the case of HMC for Gaussian prior, ReLU activation and $\Delta = 0.1$, converging to the specialisation solution only if initialised informatively. In App.~\ref{sec:hardness} we also report cases for which both ADAM (with optimised hyperparameters) and HMC initialised uninformatively can approach the specialisation performance, but they seem to require an exponential time in $d$. For what concerns the case of continuous distribution of read-outs, e.g. $P_v=\calN(0,1)$, our numerical results are inconclusive on hardness, and deserve a larger scale numerical investigation.

The two identified phases are akin to those recently described in \citet{barbier2024phase} for matrix denoising. The model we consider is also a matrix model in $\bW$, with the amount of observations scaling as the number of matrix elements. When data are scarce, the student cannot break the numerous symmetries of the problem, resulting in an ``effective rotational invariance'' at the source of the prior universality, with posterior samples having a vanishing overlap with $\btheta^0$. On the other hand, when data are sufficiently abundant, $\alpha>\alpha_{\rm sp}$, there is a ``synchronisation'' of the student's samples with the teacher. From an algorithmic point of view, however, for certain target functions the student seems to be able to find these highly performing weight configurations only when it is strongly informed about the ground-truth weights, or after a training time exponential in $d$, both scenarios signalling a possible statistical-to-computational gap.

The phenomenology observed depends on the activation function selected. 
In particular, by expanding $\sigma$ in Hermite basis we realise that the way the first three terms enter information theoretical quantities is completely described by order 0, 1 and 2 tensors later defined in \eqref{eq:def_S}, that are combinations of the inner and read-out weights. In the regime of quadratically many data, order 0 and 1 tensors are recovered exactly by the student because of the overwhelming abundance of data compared to their dimension. The challenge is thus to learn the second order tensor. On the contrary, we claim that learning any higher order tensors can only happen when the student aligns its weights with $\btheta^0$: before this ``synchronisation'', they play the role of an effective noise. This is the mechanism behind the specialisation solution.
For odd activation $\sigma$, where $\mu_2=0$, the aforementioned order-2 tensor does not contribute any more to learning. Indeed, we observe numerically that the generalisation error sticks to a constant value for $\alpha<\alpha_{\rm sp}$, whereas at the phase transition it suddenly drops. This is because the learning of the order-2 tensor is skipped entirely, and the only chance to perform better is to learn all the other higher-order tensors through specialisation.

By extrapolating universality results to generic activations, we are able to use the GAMP-RIE of~\citet{maillard2024bayes}, publicly available at~\citet{maillard2024github}, to obtain a polynomial-time predictor for test data. Its generalisation error follows our universal theoretical curve even in the $\alpha$ regime where MCMC sampling experiences a computationally hard phase with worse performance, and in particular after $\alpha_{\rm sp}$ (see Fig.~\ref{fig:gen_errors_univ_spec}, circles). Extending this algorithm, initially proposed for quadratic activation only, to generic activation is possible thanks to the identification of an \emph{effective} GLM on which the learning problem can be mapped (while the mapping is exact when $\sigma(x)=x^2$ as exploited by \citet{maillard2024bayes}), see Appendix~\ref{app:glm}. The key observation is that our effective GLM representation holds not only from a theoretical perspective to describe the universal phase, but also algorithmically.

Finally, we emphasise that our theory is consistent with  \citet{cui2023bayes}, as our generalisation curves at $\alpha\to0$ match theirs at $\alpha_1:=n/d\to\infty$, which is when the student learns perfectly the combinations $\bv^{0\intercal}\bW^0/\sqrt{k}$ (but nothing more).

\section{Evaluation of the free entropy and generalisation error by the replica method\label{sec:theory}}
The goal is to compute the asymptotic free entropy by the replica method \cite{mezard1987spin}, a powerful heuristic from spin glasses that can be used in machine learning  \cite{engel2001statistical}. Define the ``replicated free entropy'' $f_{n,s}:=\ln \EE\mathcal{Z}^s(\mathcal{D})/(ns)$. The starting point to tackle the data average is $\widetilde\lim\,\EE \ln \calZ /n=\widetilde \lim \lim_{s\to 0^+}f_{n,s}= \lim_{s\to 0^+}\widetilde\lim f_{n,s}$, assuming the limits commute. Recall $\btheta^0$ are the teacher weights. Consider first $s\in \mathbb{N}^+$. Let 
\begin{align*}
    \{\lambda^a(\btheta^a)\}_{a=0,\ldots,s}:=\Big\{\frac{\bv^{a\intercal}}{\sqrt{k}}\sigma\Big(\frac{\bW^a\bx}{\sqrt{d}}\Big)\Big\}_{a=0,\ldots,s}
\end{align*}
be ``replicas'' of the post-activation. We have
\begin{equation*}
\begin{aligned}
\EE\mathcal{Z}^s(\mathcal{D})&=\int \prod_{a=0}^sdP_v(\bv^a)dP_W(\bW^a)\\
    &\quad\times\Big[\EE_\bx\int dy \prod_{a=0}^s P_{\rm out}(y\mid \lambda^a(\btheta^a))\Big]^n.
    \end{aligned}
\end{equation*}
The key is to now identify the law of $\{\lambda^a\}_{a=0,\ldots,s}$,
which are dependent random variables due to the common random Gaussian input $\bx$, conditionally on $\{\btheta^a:=(\bv^a,\bW^a)\}_a$. Our key hypothesis is that \emph{we assume $\{\lambda^a\}$ to be jointly Gaussian}, an ansatz we cannot prove due to the presence of the non-linearity but that we validate a posteriori thanks to the excellent match between our theory and the empirical generalisation curves, see Sec.~\ref{sec:result}. Similar Gaussian assumptions have been the crux of a whole line of recent works on the analysis of neural networks, and are now known under the name of ``Gaussian equivalence'' \citep{goldt2020modeling,hastie2022surprises,mei2022generalization,goldt2022gaussian,hu2022universality}. This can also sometimes be heuristically justified based on Breuer–Major Theorems \citep{nourdin2011quantitative,pacelli2023}. 

\begin{figure}[t!!]
    \centering
    \includegraphics[width=1\linewidth]{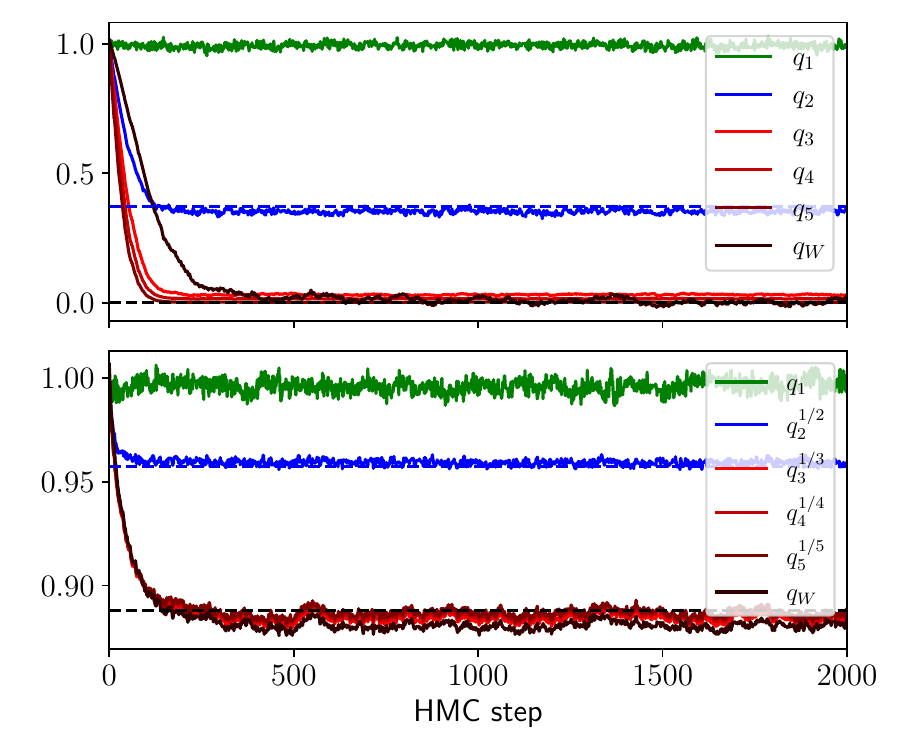}
    \vspace{-20pt}
    \caption{Hamiltonian Monte Carlo dynamics of the overlaps $q_W=Q_W^{01}$ and $q_\ell=Q_\ell^{01}$ ($\ell = 1,\dots,5)$ between student and teacher weights, with activation function $\sigma(x) = \text{He}_1(x) + \text{He}_2(x)/\sqrt{2} + \text{He}_3(x)/6$, $d=200$, $\gamma=0.5$, $\Delta=1.25$ and two different choices of sample rate: $\alpha=0.5$ (\textbf{Top}), $\alpha=5$ (\textbf{Bottom}). The inner weights $\bW^0$ of the teacher are Gaussian, while the read-outs $\bv^0$ binary. The dynamics is initialised informatively, i.e. on the teacher weights, and the read-outs kept fixed during training. The overlap $q_1$ is fluctuating close to 1 in both figures. \textbf{Top}: The overlaps $q_W$ and $q_\ell$ for $\ell  \ge 3$ at equilibrium converge to 0, while $q_2$ can be estimated by the universal theory (blue dashed line). \textbf{Bottom}: The overlaps $q_\ell$ for $\ell \ge 3$ are 
    trivially equal to $q_W^\ell$, also during the dynamics, in agreement with \eqref{Q=QwQv}. The theoretical values of the overlaps $q_W$ and $q_2$ are shown in black and blue dashed lines, respectively.
}
    \label{fig:overlaps}
\end{figure}

Recalling the Hermite expansion of $\sigma$, by using Mehler's formula, see App.~\ref{app:hermite}, the covariance $K^{ab}:=\EE\lambda^a\lambda^b$ reads
\begin{align}
K^{ab}&=\sum_{\ell=1}^\infty\frac{\mu^2_\ell}{\ell!}
\sum_{i,j=1}^k \frac{v^a_i(\Omega^{ab}_{ij})^\ell v^b_j}{k}=:\sum_{\ell=1}^\infty\frac{\mu^2_\ell}{\ell!} Q_\ell^{ab}
\label{eq:K}
\end{align}
where, given two replica indices $a,b$, we introduced the matrix overlap with indices $i,j=1,\ldots, k$ defined as
\begin{align}
\Omega^{ab}_{ij}:=\sum_{\alpha=1}^d\frac{W_{i\alpha}^aW^b_{j\alpha}}{d}.\label{omega}
\end{align}
The covariance matrix $\bK$ of $(\lambda^a)$ is a complicated object but, as we argue hereby, simplifications occur in the large dimension limit. In particular, the first two ``overlaps'' below will play a special role:
\begin{align}    Q_1^{ab}&=\sum_{\alpha=1}^d\sum_{i,j=1}^k\frac{v_i^aW_{i\alpha}^aW_{j\alpha}^bv_j^b}{kd}, \label{eq:def_Q1}\\
Q_2^{ab}&=\sum_{\alpha_1,\alpha_2=1}^d\sum_{i,j=1}^k \frac{v_i^a W_{i\alpha_1}^aW_{i\alpha_2}^aW_{j\alpha_1}^b W_{j\alpha_2}^b v_j^b}{kd^2}.
    \label{eq:def_Q2}
\end{align}
We claim that the higher-order overlaps $(Q_\ell^{ab})_{\ell\ge 3}$, a priori needed for the covariance $K^{ab}$, can be simplified drastically as functions of simpler order parameters:
\begin{align}
    Q_W^{ab}:=\frac{1}{kd}\Tr[\bW^a\bW^{b\intercal}],\quad Q_v^{ab}:=\frac{1}{k}\bv^{a\intercal}\bv^{b}. \label{eq:def_QW}
\end{align}
The reason is the following. In the covariance $K^{ab}$, Eq.~\eqref{eq:K}, the Hadamard powers of the overlap $\bOmega^{ab}$ appear inside quadratic forms with read-out vectors a priori correlated with it. For Hadamard powers $\ell \ge 3$, the off-diagonal part of the matrix $(\bOmega^{ab})^{\circ \ell}$ obtained from typical weight matrices sampled from the posterior, is of sufficiently small order to consider it diagonal when evaluating any quadratic form, including with vectors strongly aligned with its eigenvectors. In other words, the eigenvectors of $(\bOmega^{ab})^{\circ \ell}$ are sufficiently close to the standard basis for any quadratic form to be dominated by the diagonal contribution in the large system limit. The same happens, e.g., for a standard Wishart matrix: its eigenvectors and the ones of its square Hadamard power are delocalised, while for higher powers, the eigenvectors are strongly localised. Moreover, we assume the diagonal of $\bOmega^{ab}$ to concentrate onto a constant, thus equal to $Q_W^{ab}$. With these observations in mind, we get the following simplification at leading order:
\begin{align}\label{eq:Omega_ansatz}
    (\Omega_{ij}^{ab})^\ell\approx\delta_{ij}(Q_W^{ab})^\ell \quad \text{for}\quad \ell\geq 3.
\end{align}
Approximate equality here is up to a matrix with vanishing norm in the large size limit. This implies in particular that
\begin{align}
    Q^{ab}_\ell \approx (Q_W^{ab})^\ell Q^{ab}_v \quad \text{for}\quad \ell\geq 3.\label{Q=QwQv}
\end{align}
This assumption is verified numerically, see Fig.~\ref{fig:overlaps}, which shows that it even holds during sampling by Monte Carlo and not just at equilibrium.
For what follows, it is convenient to define the symmetric tensors $\bS_\ell^a$ with entries
\begin{equation}    S^a_{\ell;\alpha_1\ldots\alpha_\ell} := \frac{1}{\sqrt{k}}\sum_{i=1}^k v_i^a W^a_{i\alpha_1} \cdots W^a_{i\alpha_\ell} .
    \label{eq:def_S}
\end{equation}
Indeed, the generic $\ell$-th term of the series~\eqref{eq:K} can be written as the overlap $\bQ_\ell\in\mathbb{R}^{s+1\times s+1}$ of these tensors, for example 
\begin{align*}
    Q_1^{ab} &= \frac{1}{d} \bS_{1}^{a\intercal} \bS_{1}^b,\qquad Q_2^{ab} = \frac{1}{d^2} \Tr \,\bS_{2}^{a} \bS_{2}^b.
\end{align*}
Then, the average replicated partition function reads $\EE\mathcal{Z}^s = \int d\bQ_1d\bQ_2 d\bQ_Wd\bQ_v \exp(F_S+ nF_E)$
where $F_E, F_S$ are functions of the symmetric matrices $\bQ_1,\bQ_2,\bQ_W,\bQ_v\in \mathbb{R}^{s+1\times s+1}$. The so-called ``energetic potential'' is defined as
\begin{align}
    &e^{F_E}
    :=\int dyd\blambda\frac{e^{-\frac{1}{2}\blambda^\intercal\bK^{-1}\blambda}}{\sqrt{(2\pi)^{s+1}\det \bK}}\prod_{a=0}^s P_{\rm out}(y\mid \lambda^a). \label{eq:FE}
    \end{align}
    It takes this form following our Gaussian assumption on the replicated post-activations, conditional on the overlaps. The ``entropic potential'' taking into account the degeneracy of the overlap order parameters is instead given by
    \begin{align}
    &e^{F_S}:= 
    \int \prod_{a=0}^s d\bS_1^a d\bS_2^a \int \prod_{a=0}^s dP_v(\bv^a)dP_W(\bW^a)\nonumber\\
    &\times\prod_{a=0}^s\delta\Big(\bS^a_2-
    \frac{\bW^{a\intercal}(\bv^a)\bW^a}{\sqrt{k}}\Big)
    \delta\Big(\bS^a_1-\frac{\bv^{a\intercal}\bW^{a}}{\sqrt{k}}\Big)\nonumber\\
    &\times
    \prod_{a\leq b,0}^s\delta\Big(Q_W^{ab}-\frac{\Tr[\bW^a\bW^{b\intercal}]}{kd}\Big)\delta\Big(Q_v^{ab}-\frac{\bv^{a\intercal} \bv^{b}}{k}\Big) \nonumber\\    
    &\times
    \prod_{a\leq b,0}^s\Big[\delta\Big(Q_1^{ab}-\frac{\bS_{1}^{a\intercal} \bS_{1}^b}{d} \Big)\delta\Big(Q_2^{ab}-\frac{\Tr \,\bS_{2}^{a} \bS_{2}^b}{d^2} \Big)\Big].\label{eFS}
\end{align}
The energetic term is easily computed, see App.~\ref{app:energetic_potential}. For the entropic term, we interpret \eqref{eFS} as the (unnormalised) average of the last factor $\prod_{a\leq b}[\delta(Q_1^{ab}-\cdots)\delta(Q_2^{ab}-\cdots)]$ under the law of the tensors $(\bS_1^a,\bS_2^a)$ induced  by the replicated weights conditionally on $\bQ_W,\bQ_v\in\mathbb{R}^{s+1\times s+1}$:
\begin{align*}
    &P((\bS^a_1,\bS_2^a)_{a=0}^s\mid\bQ_W,\bQ_v):=
    V_W(\bQ_W)^{-kd}V_v(\bQ_v)^{-k}\nonumber\\
    &\times\int \prod_{a=0}^s dP_v(\bv^a)dP_W(\bW^a)\nonumber\\
    &\times \prod_{a=0}^s
    \delta\Big(\bS^a_2-
    \frac{\bW^{a\intercal}(\bv^a)\bW^a}{\sqrt{k}}\Big)
    \delta\Big(\bS^a_1-\frac{\bv^{a\intercal}\bW^{a}}{\sqrt{k}}\Big)\nonumber\\
    &\times\prod_{a\leq b,0}^s\delta(kdQ_W^{ab}-\Tr[\bW^a\bW^{b\intercal}])\delta(kQ_v^{ab}-\bv^{a\intercal}\bv^b)
\end{align*}with the normalisations
\begin{align*}
    &V_W(\bQ_W)^{kd}\nonumber:=\\
    &\int \prod_{a=0}^sdP_W(\bW^a)\prod_{a\leq b,0}^s\delta(kd\,Q_W^{ab}-\Tr[\bW^a\bW^{b\intercal}]),\\
    &V_v(\bQ_v)^k:=\int \prod_{a=0}^sdP_v(\bv^a)\prod_{a\leq b,0}^s\delta(kQ_v^{ab}-\bv^{a\intercal}\bv^b).
\end{align*}
Given that the number $n$ of data scales as $d^2$, and that $\bS_1^a$ are only $d$-dimensional, they can be reconstructed perfectly: we assume that at equilibrium the related overlaps $Q_1^{ab}$ are identically 1, or saturate to their maximum value. In other words, in the quadratic regime, the $\mu_1$ contribution in $\sigma(x) = \sum_{\ell \ge 0} \frac{\mu_\ell}{\ell !}\He_\ell (x)$ is perfectly learnable, while the higher order coefficients play a non-trivial role. In fact, once the deltas fixing $Q_1^{ab}$ are written in Fourier representation, this appears clear, since their exponents are of $O(k)$, whereas the leading terms are of $O(k^2)$, ultimately implying trivial saddle point equations for $Q_1^{ab}$, if tracked down. We thus neglect said delta functions over $Q_1^{ab}$ and set directly $\bQ_1 \to \bm{1}\bm{1}^{\intercal}$, the all-ones matrix.
In contrast, \citet{cui2023bayes} study the linear data regime $n\sim k$, where the $\mu_1$ term is the only potentially learnable one.

After these operations we get at leading exponential order
\begin{align}
    &\EE\mathcal{Z}^n(\mathcal{D})= \int \prod_{a\leq b,0}^s dQ_2^{ab}dQ_W^{ab}dQ_v^{ab}\nonumber\\
    &\times  e^{nF_E(\bQ_1\to \bm{1}\bm{1}^{\intercal},\bQ_2,\bQ_W,\bQ_v) + kd\ln V_W(\bQ_W)}\label{eq:Zn_last_common}\\
    &\times\int dP((\bS_2^a)\mid\bQ_W,\bQ_v)\prod_{a\leq b,0}^s
    \delta(d^2Q_2^{ab}-\Tr[\bS^a_2\bS^{b\intercal}_2]),\nonumber
\end{align}
where $P((\bS_2^a)\mid\bQ_W,\bQ_v)$ is the conditional marginal law of $(\bS_2^a)$. We neglected the sub-leading term $\exp(k\ln V_v(\bQ_v))$ which cannot affect the final free entropy at leading order.
From here on, our ansatz on $P((\bS_2^a)\mid\bQ_W,\bQ_v)$ will determine which of the two phases is described. 

In the \emph{universal} phase, which occurs for low $\alpha$, scarcity of data prevents the student from learning separately the teacher's weights $(\bv^0,\bW^0)$, that are instead recovered only via the combinations $(\bS_1^0=k^{-1/2}\bv^{0\intercal}\bW^{0},\bS_2^0=k^{-1/2}\bW^{0\intercal}(\bv^0)\bW^0)$. Here, the generalisation error is independent of the choice of the prior $P_W$. 
A second approach, inspired by \citet{sakata2013} (see also \citet{kabashima2016phase}), accurately predicts the Bayes-optimal performance of the model for high $\alpha$ and a large class of target functions: here the student is able to overlap non-trivially with the actual teacher's $(\bv^0,\bW^0)$. We call this the \emph{specialisation} phase. 

\paragraph{Universal phase\label{subsec:universal}}
For low $\alpha$, the student is sensitive only to the combinations $\bS_\ell$, defined in \eqref{eq:def_S}. In the large $d$ limit an effective rotational invariance of the matrix $\bS_2$ holds. As argued also in \citet{barbier2024phase} this makes it impossible to have $O(1)$ overlaps of the columns of the student's $\bW$ with those of the teacher, resulting in a trivial overlap $\bQ_W$. Instead, each column of $\bW$ has a non-trivial overlap profile of $O(1/\sqrt{k})$ with all columns of $\bW^0$, a property captured by spherical integrals \cite{itzykson1980planar,matytsin1994large,guionnet2002large}. Therefore, a meaningful ansatz for $dP((\bS_2^a)\mid\bQ_W,\bQ_v)$ to plug in \eqref{eq:Zn_last_common} is 
\begin{equation}
    dP((\bS_2^a)\mid\bQ_W,\bQ_v) =\prod_{a=0}^ndP(\bS_2^a),\quad Q^{ab}_W=\delta_{ab},
    \label{eq:S2_universal}
\end{equation}
where $dP(\bS_2^a)$ is the probability distribution of the random matrix $k^{-1/2}\tilde \bW^{a\intercal}(\bv^a)\tilde\bW^a$, with each $\tilde \bW^a$ being made of i.i.d. standard Gaussian entries due to universality, 
and i.i.d. $v^a_i\sim P_v$. We stress that Gaussian universality is only on the choice of the prior over the inner weights $\bW$, as the distribution of $\bv$ enters explicitly the law of $\bS_2$ \cite{maillard2024bayes}. From \eqref{Q=QwQv} one can see that ansatz \eqref{eq:S2_universal} removes the dependence on $\bQ_v$ from the partition function (we are assuming $Q_v^{aa}=1$). 

\paragraph{Specialisation phase\label{subsec:specialisation}}
For high enough $\alpha$, a Bayes-optimal student can learn something about the teacher's weights. In this regime, for the same reason we took $\bQ_1\to \bm{1}\bm{1}^\intercal$ to write \eqref{eq:Zn_last_common} ($\bQ_1$ is a statistics trivially learnable in the quadratic data regime $\alpha >0$), we can assume that $\bQ_v\to \bm{1}\bm{1}^\intercal$ as well: if the student is able to learn non-trivially the $dk = \Theta(d^2)$ inner weights $\bW^0$, then it must be that the data contains enough information to reconstruct perfectly the few $k=\Theta(d)$ parameters of the read-out layer $\bv^0$. In this phase, the ansatz we propose to plug in \eqref{eq:Zn_last_common} is
\begin{align}
    &dP((\bS_2^a)\mid\bQ_W,\bQ_v) = \Big(\prod_{a=0}^s d \bS^a_{2}\prod_{\alpha=1}^d  \delta(S^a_{2;\alpha \alpha}  \!-\! \sqrt{k}\E v)\Big)\nonumber\\    &\qquad\times\prod_{\alpha_1<\alpha_2}^d  
    \frac{e^{-\frac{1}{2}\sum_{a,b=0}^s S^{a}_{2;\alpha_1\alpha_2}(\bQ_W^{\circ 2})^{-1}_{ab}S^{b}_{2;\alpha_1\alpha_2}}}{\sqrt{(2\pi)^{s+1}\det(\bQ_W^{\circ 2})}},\label{eq:S2_specialisation}
\end{align}
where $\E v$ is the mean of the read-out prior $P_v$. In words, first, the diagonal elements of $\bS_2^a$ are $d$ random variables whose $O(1)$ fluctuations cannot affect the free entropy in the asymptotic regime we are considering, being too few compared to $n=\Theta(d^2)$. Hence, we assume they concentrate to their mean. Concerning the $d(d-1)/2$ off-diagonal elements of the matrices $(\bS_2^a)_a$, they are zero-mean variables whose distribution at given $\bQ_W$, $\bQ_v$ is assumed to be factorised over the input indices. 
It is not hard to show that the true measure $dP((\bS_2^a)\mid\bQ_W,\bQ_v)$ in Eq.~\eqref{eq:Zn_last_common} is such that $\widetilde{\lim}\,\E[ \Tr \bS^a_{2} \bS^{b\intercal}_{2} \mid \bQ_W, \bQ_v]/d^2 = (Q_W^{ab})^2 Q_v^{ab}=(Q_W^{ab})^2$, 
which is non-trivial due to $\bQ_W$ when the student aligns its hidden layer with the teacher. The Gaussian ansatz~\eqref{eq:S2_specialisation} is the simplest one that matches this property.

The full derivation of our results under the ans\"atze~\eqref{eq:S2_universal}, \eqref{eq:S2_specialisation} combined with a replica symmetric assumption, i.e., a form $Q^{ab} = r \delta_{ab} + q (1-\delta_{ab})$ for all overlaps and for $K^{ab}$, is found in App.~\ref{app:replicas} and yields the free entropies in Result~\ref{res:free_entropy}. Replica symmetry is rigorously known to be correct in general settings of Bayes-optimal learning, see \citet{barbier2022strong} and \citet{barbier2019adaptive}.

\section{Conclusion and perspectives}

In this work we provided an effective description of the optimal generalisation capability of a fully-trained two-layer neural network of extensive width with generic activation when the sample size scales with the number of trainable parameters. The analysis in this setting has resisted for a long time to attempts based on mean-field approaches used, e.g., to study committee machines \cite{barkai1992broken,engel1992storage,schwarze1992generalization,schwarze1993generalization,mato1992generalization,monasson1995weight,aubin2018committee,baldassi2019properties}. We unveil two phases, each requiring a specific ansatz in replica theory: a universal phase where the model performance is independent of the law of its internal weights and the teacher network is not recovered, and a specialisation phase where the student's inner weights can align with the teacher's.

A natural extension is to consider non Bayes-optimal models, e.g., trained 
through empirical risk minimisation to learn a teacher with mismatched architecture. 
The formalism we provide here can be extended to these cases, by keeping track of additional order parameters. 
The extension to deeper architectures is also possible, in the vein of \citet{cui2023bayes} and \citet{pacelli2023} who analysed the overparametrised proportional regime.  Extensions to account for structured inputs is another direction: data with a covariance~\citep{monasson1992,loureiro2021real}, mixture models~\citep{delgiudice1989,loureiro2021gmm}, hidden manifolds~\citep{goldt2020modeling}, object manifolds and simplexes~\citep{chung2018manifold,rotondo2020beyond}, etc.

Phase transitions in supervised learning are known in the statistical mechanics literature at least since~\citet{gyorgyi1990first}, when theoretical understanding was limited to linear models. An interesting research direction is the possible connection with Grokking, a sudden drop in generalisation error occurring during the training of neural nets close to interpolation (see~\citet{rubin2024grokking} for an interpretation in terms of thermodynamic first-order phase transitions).

A more systematic analysis on the computational hardness of the problem (as carried out for multi-index models in~\citet{troiani2025fundamental}) is an important step towards a full characterisation of the class of functions that are fundamentally hard to learn. A striking observation from our preliminary analysis in App.~\ref{sec:hardness} is that target functions with random continuous read-out weights are easier to learn than with discrete distributions, yet we cannot rule out that they also require an exponential time to be learned. 

As a final note, we observe small but non-negligible deviations of experiments from the theory for some target functions close to transitions (see for instance Fig.~\ref{fig:gen_error_gauss} around $\alpha=4$). If not due to finite size effects, we aim at correcting these small discrepancies in future works.

\section*{Software and data}

A GitHub repository to reproduce the results can be found at \href{https://github.com/Minh-Toan/extensive-width-NN}{https://github.com/Minh-Toan/extensive-width-NN}

\section*{Acknowledgements}
J.B., F.C., M.-T.N., M.P. were funded by the European Union (ERC, CHORAL, project number 101039794). Views and opinions expressed are however those of the authors only and do not necessarily reflect those of the European Union or the European Research Council. Neither the European Union nor the granting authority can be held responsible for them.

M.P. thanks Vittorio Erba and Pietro Rotondo for interesting discussions and suggestions.

\bibliography{main.bib}
\bibliographystyle{arxiv2025}

\newpage
\appendix
\onecolumn

\section{Hermite basis and Mehler's formula\label{app:hermite}}
Recall the Hermite expansion of the activation:
\begin{equation}
    \sigma(x) = \sum_{\ell = 0}^{\infty} \frac{\mu_\ell}{\ell !}\He_\ell (x).
    \label{eq:sigma_hermite}
\end{equation}
We are expressing it on the basis of the probabilist's Hermite polynomials, generated through
\begin{equation}
    \He_\ell(z) = \frac{\diff^\ell}{{\diff t}^\ell} \exp\big(t z - t^2/2 \big)\Big|_{t=0}.
    \label{eq:hermite_G}
\end{equation}
The Hermite basis has the property of being orthogonal with respect to the standard Gaussian measure, which is the distribution of the input data:
\begin{equation}
    \int Dz\, \He_k(z) \He_\ell(z) = \ell!\,\delta_{k\ell} ,
\end{equation}
where $Dz := \diff z \exp(-z^2/2)/\sqrt{2\pi} $. By orthogonality, the coefficients of the expansions can be obtained as
\begin{equation}
    \mu_\ell = \int Dz \He_\ell(z)\sigma(z).
\end{equation}
Moreover,
\begin{equation}
    \label{eq:avg_sigma}
    \E[\sigma(z)^2] = \int D z\, \sigma(z)^2 = \sum_{\ell=0}^{\infty} \frac{\mu_\ell^2}{\ell!}.
\end{equation}
These coefficients for some popular choices of $\sigma$ are reported in Table~\ref{tab:Hermite} for reference.
\begin{table}[b]
    \caption{First Hermite coefficients of the activation functions reported in Fig.~\ref{fig:gen_errors_univ_spec}. $\theta$ is the Heaviside step function.}
    \label{tab:Hermite}
    \vskip 0.15in
    \begin{center}
    \begin{small}
    \begin{tabular}{l|ccccccc}
    \toprule
        $\sigma(z)$ & $\mu_0$ & $\mu_1$ & $\mu_2$ & $\mu_3$ & $\mu_4$ & $\cdots$ & $\E_z[\sigma(z)^2]$  \\
        \midrule
        $\relu(z) = z \theta(z)$ & $1/\sqrt{2 \pi }$ & $1/2$ & $1/\sqrt{2 \pi }$  & 0 & $-1/\sqrt{2 \pi }$& $\cdots$ & 1/2 \\
        $\elu(z) = z\theta(z) + (e^{z} - 1)\theta(-z) $ & 0.16052 & 0.76158 & 0.26158 & -0.13736 & -0.13736 & $\cdots$ & 0.64494 \\
        \bottomrule
    \end{tabular}
    \end{small}
    \end{center}
    \vskip -0.1in
\end{table}
The Hermite basis can be generalised to an orthogonal basis with respect to the Gaussian measure with generic variance:
\begin{equation}
    \He_\ell^{[r]}(z) = \frac{\diff^\ell}{\diff t^\ell} \left. \exp(tz -  t^2 r/2)\right|_{t=0},
\end{equation}
so that, with $D_r z := \diff z \exp(-z^2/2r)/\sqrt{2\pi r} $, we have
\begin{equation}
    \int D_r z\, \He_k^{[r]}(z) \He_\ell^{[r]}(z) =  \ell! \,r^\ell\delta_{k\ell}.
\end{equation}

From Mehler's formula
\begin{equation}
    \frac{1}{2\pi\sqrt{r^2-q^2}} \exp\!\left[-\frac{1}{2} (u,v) \begin{pmatrix}
        r & q \\ q & r
    \end{pmatrix}^{-1} \begin{pmatrix}
        u\\v
    \end{pmatrix} \right] = \frac{e^{-\frac{u^2}{2r}}}{\sqrt{2\pi r}} \frac{e^{-\frac{v^2}{2r}}}{\sqrt{2\pi r}} \sum_{\ell = 0}^{+\infty} \frac{q^{\ell}}{\ell! r^{2\ell}} \He_{\ell}^{[r]}(u) \He_{\ell}^{[r]}(v),
    \label{eq:mehler}
\end{equation}
and by orthogonality of the Hermite basis, \eqref{eq:K} readily follows by noticing that the variables 
$(h_i^a = (\bW^a \bx)_i/\sqrt{d})_{i,a}
$
at given $(\bW^a)$ are Gaussian with covariances $\Omega^{ab}_{ij}$, Eq. \eqref{omega}, so that
\begin{equation}
    \EE [\sigma(h_{i}^a)\sigma(h_{j}^b)] = \sum_{\ell=0}^{\infty} \frac{(\mu_\ell^{[r]})^2}{\ell!r^{2\ell}} (\Omega_{ij}^{ab})^\ell,\qquad \mu_\ell^{[r]} = \int D_rz\, \He^{[r]}_\ell(z)\sigma(z).
\end{equation}
Moreover, as $r=\Omega^{aa}_{ii}$ converges for $d$ large to the variance of the prior of $\bW^0$ by Bayes-optimality, whenever $\Omega^{aa}_{ii} \to 1$ we can specialise this formula to the simpler case $r=1$ we reported in the main text.

\section{Nishimori identities}\label{app:nishiID}
The Nishimori identities are a very general set of symmetries arising in inference in the Bayes-optimal setting as a consequence of Bayes' rule. To introduce them, consider a test function $f$ of the teacher weights, collectively denoted by $\btheta^0$, of $s-1$ replicas of the student's weights $(\btheta^a)_{2\leq a\leq s}$ drawn conditionally i.i.d. from the posterior, and possibly also of the training set $\mathcal{D}$: $f(\btheta^0,\btheta^2,\dots,\btheta^s ;\mathcal{D})$. Then
\begin{align}
    \mathbb{E}_{{\boldsymbol{\theta}}^0,\mathcal{D}}\langle f(\btheta^0,\btheta^2,\dots,\btheta^s ;\mathcal{D})\rangle=
    \mathbb{E}_{{\boldsymbol{\theta}}^0,\mathcal{D}}\langle f(\btheta^1,\btheta^2,\dots,\btheta^s ;\mathcal{D})\rangle,
\end{align}where we have replaced the teacher's weights with another replica from the student. The proof is elementary, see e.g. \cite{barbier2019glm}.

The Nishimori identities have some consequences also on our replica symmetric ansatz for the free entropy. In particular, they constrain the values of some order parameters. For instance
\begin{align}
    m_2= \widetilde{\lim}
    \frac{1}{d^2}\EE_{\mathcal{D},\btheta^0}\langle\Tr[\bS_2^a\bS_2^0]\rangle=
    \widetilde{\lim}\frac{1}{d^2}\EE_{\mathcal{D}}\langle\Tr[\bS_2^a\bS_2^b]\rangle= q_2,\quad \text{for }a\neq b
\end{align}assuming concentration of such order parameters takes place, which can be proven in great generality in Bayes-optimal learning \cite{barbier2021overlap,barbier2022strong}. Another example is
\begin{align}
    r_2=\widetilde{\lim}\frac{1}{d^2}\EE_{\mathcal{D}}\langle\Tr[(\bS_2^a)^2]\rangle=\widetilde{\lim}\frac{1}{d^2}\EE_{\btheta^0}\Tr[(\bS_2^0)^2]=\rho_2=1+\gamma(\EE v^0)^2.
\end{align}When the value of some order parameters is determined by the Nishimori identities, as for $r_2,\rho_2$, then the respective Fourier conjugates $\hat r_2,\hat \rho_2$ vanish (meaning that the desired constraints were already enforced without the need of additional delta functions). This is because in the entropic count of how many configurations make $r_2,\rho_2$ take those values in the posterior measure, these constraints are automatically imposed by the measure.

\section{Alternative representation for the mean-square generalisation error\label{app:gen_err}}

In this section we report the details on how to obtain Result~\ref{res:gen_error} and how to write the generalisation error defined in \eqref{eq:gen_error_def} in a form more convenient for numerical sampling. From its definition, 
the Bayes-optimal generalisation error can be recast as
\begin{align}
    \varepsilon^{\rm opt}= \EE_{\btheta^0,\bx_{\rm test} } \EE[y^2_{\rm test}\mid \lambda^0]
    -2\EE_{\btheta^0,\mathcal{D},\bx_{\rm test}} \EE[y_{\rm test}\mid \lambda^0]\langle\EE[y\mid \lambda]\rangle
    +  \EE_{\btheta^0,\mathcal{D},\bx_{\rm test}}\langle \EE[y\mid \lambda]\rangle^2,\label{eyStartingPoint}
\end{align}
where $\EE[y\mid \lambda]=\int dy\, y\, P_{\rm out}(y\mid \lambda)$, and  $\lambda^0$, $\lambda$ are the random variables (random due to the test input $\bx_{\rm test}$, drawn independently of the training data $\mathcal{D}$, and their respective weights $\btheta^0,\btheta$)
\begin{align}
    \lambda^0=\lambda(\btheta^0,\bx_{\rm test}) =\frac{\bv^{0\intercal}}{\sqrt{k}}\sigma\Big(\frac{\bW^0\bx_{\rm test}}{\sqrt{d}}\Big),\qquad
    \lambda=\lambda^1=\lambda(\btheta,\bx_{\rm test}) =\frac{\bv^{\intercal}}{\sqrt{k}}\sigma\Big(\frac{\bW\bx_{\rm test}}{\sqrt{d}}\Big).
\end{align}
Recall that the bracket $\langle\,\cdot\,\rangle$ is the average w.r.t. to the posterior and acts on $\btheta^1=\btheta,\btheta^2,\ldots$ which are replicas, i.e., conditionally i.i.d. samples from $dP(\,\cdot\mid \mathcal{D})$. Notice that the last term on the r.h.s.\ of \eqref{eyStartingPoint}, that can be rewritten as
\begin{align*}
    \EE_{\btheta^0,\mathcal{D},\bx_{\rm test}}\langle \EE[y\mid \lambda]\rangle^2=
    \EE_{\btheta^0,\mathcal{D},\bx_{\rm test}}\langle   \EE[y\mid \lambda^1]\EE[y\mid \lambda^{2}]\rangle,
\end{align*}with superscripts being replica indices, i.e., $\lambda^a:=\lambda(\btheta^a,\bx_{\rm test})$.

In order to show Result~\ref{res:gen_error} for a generic $P_{\rm out}$ we assume the joint Gaussianity of the variables $(\lambda^0,\lambda^1,\lambda^2,\ldots)$, with covariance given by $K^{ab}$ with $a,b=0,1,2,\ldots$. Indeed, in the limit $\widetilde\lim$, our theory considers $(\lambda^a)_{a\ge 0}$ as jointly Gaussian under the randomness of a common input, here $\bx_{\rm test}$, conditionally on the weights $(\btheta^a)$. Their covariance depends on the weights $(\btheta^a)$ through various overlap order parameters introduced in the main. But in the large limit $\widetilde\lim$ these overlaps are assumed to concentrate under the quenched posterior average $\EE_{\btheta^0,\mathcal{D}}\langle\,\cdot\,\rangle$ towards non-random asymptotic values predicted to be \eqref{eq:K_RS}, with the overlaps entering $K^{ab}$ given by the solution of $(\rm S_{uni})$ or $(\rm S_{sp})$ (depending on the phase) with maximum free entropy. This hypothesis is then confirmed by the excellent agreement between our theoretical predictions based on this assumption and the experimental results. This implies directly the result~\eqref{eq:Bayes_error_result} from definition \eqref{eq:Bayes_error_def}. For the special case of optimal mean-square generalisation error it yields
\begin{align}\label{eq:general_gen_error_Gaussian_eq}
    \widetilde\lim \,\varepsilon^{\rm opt}=\EE_{\lambda^0}\EE[y^2_{\rm test}\mid \lambda^0]-2
    \EE_{\lambda^0,\lambda^1}\EE[y_{\rm test}\mid \lambda^0]\EE[y\mid \lambda^1]
    +\EE_{\lambda^1,\lambda^2}\EE[y\mid \lambda^1]\EE[y\mid \lambda^2]
\end{align}
where, in the replica symmetric ansatz, 
\begin{align}
    \EE[(\lambda^0)^2]=K^{00},\quad \EE[\lambda^0\lambda^1]= \EE[\lambda^0\lambda^2]=K^{01},\quad \EE[\lambda^1\lambda^2]=K^{12},\quad 
    \EE[(\lambda^1)^2]=\EE[(\lambda^2)^2]=K^{11}.
\end{align} For the dependence of the elements of $\bK$ on the overlaps under this ansatz we defer the reader to \eqref{eq:m_K-q_K-def}, \eqref{eq:rho_K-r_K-def}. In the Bayes-optimal setting, using the Nishimori identities (see App.~\ref{app:nishiID}), one can show that $K^{01}=K^{12}$ and $K^{00}=K^{11}$. Because of these identifications, we would additionally have
\begin{align}
    \EE_{\lambda^0,\lambda^1}\EE[y_{\rm test}\mid \lambda^0]\EE[y\mid \lambda^1]
    =\EE_{\lambda^1,\lambda^2}\EE[y\mid \lambda^1]\EE[y\mid \lambda^2].
\end{align}Plugging the above in \eqref{eq:general_gen_error_Gaussian_eq} yields \eqref{eq:gen_err_result}.

Let us now prove a formula for the optimal mean-square generalisation error written in terms of the overlaps that will be simpler to evaluate numerically, which holds for the special case of linear read-out with Gaussian label noise $P^0_{\rm out}(y\mid \lambda)=P_{\rm out}(y\mid \lambda)=\exp(-\frac1{2\Delta}(y-\lambda)^2)/\sqrt{2\pi \Delta}$. The following derivation is exact and does not require any Gaussianity assumption on the random variables $(\lambda^a)$. For the linear Gaussian channel the means verify $\EE[y\mid\lambda]=\lambda$ and $\EE[y^2\mid\lambda]=\lambda^2+\Delta$. Plugged in \eqref{eyStartingPoint} this yields
\begin{align}
    \varepsilon^{\rm opt}-\Delta=\EE_{\btheta^0,\bx_{\rm test} } \lambda^2_{\rm test}
    -2\EE_{\btheta^0,\mathcal{D},\bx_{\rm test}}  \lambda^0 \langle\lambda\rangle
    +  \EE_{\btheta^0,\mathcal{D},\bx_{\rm test}}\langle  \lambda^1\lambda^2\rangle,
\end{align}whence we clearly see that the generalisation error depends only on the covariance of $\lambda_{\rm test}(\btheta^0)=\lambda^0(\btheta^0),\lambda^1(\btheta^1),\lambda^2(\btheta^2)$ under the randomness of the shared input $\bx_{\rm test}$ at fixed weights, regardless of the validity of the Gaussian equivalence principle we assume in the replica computation. This covariance was already computed in \eqref{eq:K}; we recall it here for the reader's convenience
\begin{align}
    K(\btheta^a,\btheta^b) :=\EE\lambda^a\lambda^b=\sum_{\ell=1}^\infty\frac{\mu_\ell^2}{\ell!}\frac{1}{k}\sum_{i,j=1}^k v_i^a(\Omega^{ab}_{ij})^\ell v^b_j=
    \sum_{\ell=1}^\infty\frac{\mu_\ell^2}{\ell!}Q_\ell^{ab},
\end{align}
where $\Omega^{ab}_{ij}:=d^{-1}\sum_{\alpha=1}^d W_{i\alpha}^a W_{j\alpha}^b$, and $Q_\ell^{ab}$ as introduced in \eqref{eq:K} for $a,b=0,1,2$. We stress that $K(\btheta^a,\btheta^b)$ is not the limiting covariance $K^{ab}$ whose elements are in \eqref{eq:m_K-q_K-def}, \eqref{eq:rho_K-r_K-def}, but rather the finite size one. $K(\btheta^a,\btheta^b)$ provides us with an efficient way to compute the generalisation error numerically, that is through the formula
\begin{align}
    \varepsilon^{\rm opt}-\Delta&=\EE_{\btheta^0}K(\btheta^0,\btheta^0)-2\EE_{\btheta^0,\mathcal{D}}\langle K(\btheta^0,\btheta^1)\rangle+
    \EE_{\btheta^0,\mathcal{D}}\langle K(\btheta^1,\btheta^2)\rangle=\sum_{\ell=1}^\infty\frac{\mu_\ell^2}{\ell!}\EE_{\btheta^0,\mathcal{D}} \langle Q_{\ell}^{00}- 2Q_{\ell}^{01}+ Q^{12}_{\ell} \rangle.
    \label{eq:general_finitesize_gen_error_noapprox}
\end{align}
In the above, the posterior measure $\langle\,\cdot\,\rangle$ is taken care of by Monte Carlo sampling (when it equilibrates). In addition, one can verify numerically that inside an arbitrary quadratic form one can replace in the large system limit the matrix
\begin{align}
\label{eq:omega_l>3}
    (\Omega^{ab}_{ij})^\ell\approx (Q_W^{ab})^\ell\delta_{ij}=\Big(\frac{1}{kd}\Tr [\bW^a\bW^{b\intercal}]\Big)^\ell\delta_{ij}\quad\text{for }\ \ell\geq 3,
\end{align}and this for any two conditionally i.i.d. posterior samples $\bW^a,\bW^b$, both in the universal phase (where $Q_W^{ab} =\delta_{ab}$) and in the specialisation phase (where $Q_W^{ab}$ is non-trivial). Putting all these ingredients together we get
\begin{align}
    \varepsilon^{\rm opt}-\Delta
    &=\EE_{\btheta^0,\mathcal{D}} \Big\langle\mu_1^2(Q_1^{00}-2Q^{01}_1 +Q^{12}_1)+\frac{\mu_2^2}{2}(Q_2^{00}-2Q^{01}_2+Q^{12}_2)+
    g(Q_W^{00})- 2g(Q^{01}_W)+g(Q^{12}_W) \Big\rangle.
    \label{eq:general_finitesize_gene_error}
\end{align}
In the Bayes-optimal setting one can use again the Nishimori identities that imply $\EE_{\btheta^0,\mathcal{D}} \langle Q^{12}_{1}\rangle=\EE_{\btheta^0,\mathcal{D}} \langle Q^{01}_{1}\rangle$, and analogously $\EE_{\btheta^0,\mathcal{D}} \langle Q^{12}_{2}\rangle=\EE_{\btheta^0,\mathcal{D}} \langle Q^{01}_{2}\rangle$ and $\EE_{\btheta^0,\mathcal{D}} \langle g(Q^{12}_{W})\rangle=\EE_{\btheta^0,\mathcal{D}} \langle g(Q^{01}_{W})\rangle$. Inserting these identities in \eqref{eq:general_finitesize_gene_error} one gets
\begin{align}
    \varepsilon^{\rm opt}-\Delta&=\EE_{\btheta^0,\mathcal{D}} \Big\langle\mu_1^2(Q_1^{00}-Q^{01}_1 )+\frac{\mu_2^2}{2}(Q_2^{00}-Q^{01}_2)+
    g(Q_W^{00})- g(Q^{01}_W)\Big\rangle.
    \label{eq:simple_gen_error_for_numerics}
\end{align}

This formula makes no assumption (other than~\eqref{eq:omega_l>3}, which we checked numerically for the first higher-order overlaps, see Fig.~\ref{fig:overlaps}) on the distribution of the $\lambda$'s. That it depends only on the covariance is simply a consequence of the quadratic nature of the generalisation error we consider.

\begin{remark}
    Note that the derivation up to \eqref{eq:general_finitesize_gen_error_noapprox} did not assume Bayes-optimality (while \eqref{eq:simple_gen_error_for_numerics} does). Therefore, one can consider it in cases where the true posterior average $\langle \,\cdot\,\rangle$ is replaced by one not verifying the Nishimori identities. This is the formula we use to compute the generalisation error of Monte Carlo-based estimators in Fig.~\ref{fig:gen_errors_univ_spec}, bottom. This is indeed needed to compute the generalisation in the glassy regime, where MCMC cannot equilibrate.
\end{remark}

\begin{remark}
    It is easy to check that, if the posterior distribution verifies the Nishimori identities, the so-called Gibbs error
    \begin{equation}
        \varepsilon^{\rm Gibbs} := \EE_{\bm{\theta}^0,\calD,\bx_{\rm test},y_{\rm test}}
        \big\langle \big(y_{\rm test} - \EE[y\mid \lambda_{\rm test}(\btheta)]\big)^2 \big\rangle
        \label{eq:Gibbs_error}
    \end{equation}
    satisfies, in the case of Gaussian label noise,
    \begin{equation}
        \varepsilon^{\rm Gibbs} - \Delta = 2(\varepsilon^{\rm opt} - \Delta)\,.
        \label{eq:Gibbs_v_Bayes_error}
    \end{equation}
    Indeed, proceeding as before, one can show that
    \begin{equation}
        \varepsilon^{\rm Gibbs} - \Delta = \sum_{\ell=1}^\infty\frac{\mu_\ell^2}{\ell!}\EE_{\btheta^0,\mathcal{D}} \langle Q_{\ell}^{00}- 2Q_{\ell}^{01}+ Q^{11}_{\ell}\rangle .
    \end{equation}
    By the Nishimori identities, $\EE_{\btheta^0,\mathcal{D}} \langle Q^{11}_\ell\rangle=\EE_{\btheta^0,\mathcal{D}} \langle Q^{00}_\ell \rangle$, so that
    \begin{equation}
        \varepsilon^{\rm Gibbs} - \Delta = 2 \sum_{\ell=1}^\infty\frac{\mu_\ell^2}{\ell!}\EE_{\btheta^0,\mathcal{D}} \langle Q_{\ell}^{00}- Q_{\ell}^{01}\rangle ,
    \end{equation}
    whereas, from Eq.~\eqref{eq:general_finitesize_gen_error_noapprox},
        \begin{equation}
        \varepsilon^{\rm opt} - \Delta = \sum_{\ell=1}^\infty\frac{\mu_\ell^2}{\ell!}\EE_{\btheta^0,\mathcal{D}} \langle Q_{\ell}^{00}- Q_{\ell}^{01}\rangle .
    \end{equation}
\end{remark}

\section{Linking free entropy and mutual information\label{app:mutual_info}}

It is possible to relate the mutual information (MI) of the inference task to the free entropy $f_n=\E\ln \mathcal{Z}$ introduced in the main. Indeed, recalling that the teacher parameters are denoted $\bm{\theta}^0 = (\bW^0,\bv^0)$, we can write the MI as
\begin{equation}
    \frac{I(\bm{\theta}^0;\mathcal{D})}{kd} = \frac{\mathcal{H}(\mathcal{D})}{kd} - \frac{\mathcal{H}(\mathcal{D}\mid\bm{\theta}^0)}{kd},
\end{equation}
where $\mathcal{H}(Y\mid X)$ is the conditional Shannon entropy of $Y$ given $X$. It is straightforward to show that the free entropy is
\begin{equation}
  -\frac{\alpha}{\gamma}f_n = \frac{\mathcal{H}(\{ y_\mu\}_{\mu \leq n}|\{ \bx_\mu\}_{\mu \leq n})}{kd} = \frac{\mathcal{H}(\mathcal{D})}{kd} - \frac{\mathcal{H}(\{ \bx_\mu\}_{\mu \leq n})}{kd},
\end{equation}
by the chain rule for the entropy. On the other hand $\mathcal{H}(\mathcal{D}\mid\btheta^0)=\mathcal{H}(\{y_\mu\}\mid\btheta^0,\{\bx_\mu\})+\mathcal{H}(\{\bx_\mu\})$, i.e.,
\begin{equation}
    \frac{\mathcal{H}(\mathcal{D}\mid \btheta^0)}{kd} \approx -\frac{\alpha}{\gamma} \EE_{\lambda}  \int dy P_{\text{out}}(y|\lambda) \ln P_{\text{out}}(y|\lambda) + \frac{\mathcal{H}(\{ \bx_\mu\}_{\mu \leq n})}{kd} ,
\end{equation}
where $\lambda\sim \calN(0,r_K)$, with $r_K$ given by \eqref{eq:covariance_vars} (assuming here that $\mu_0=0$, see App.~\ref{app:non-centered} if the activation $\sigma$ is non-centred), and the equality holds asymptotically in $\widetilde\lim$.
This allows us to express the MI as
\begin{equation}
\label{eq:MI_generic_channel}
    \frac{I(\btheta^0;\mathcal{D})}{kd} = -\frac{\alpha}{\gamma} f_n + \frac{\alpha}{\gamma} \EE_{\lambda}  \int dy P_{\text{out}}(y|\lambda) \ln P_{\text{out}}(y|\lambda).
\end{equation} 
Specialising the equation to the Gaussian channel, one obtains
\begin{equation}
\label{eq:MI_gaussian_channel}
    \frac{I(\btheta^0;\mathcal{D})}{kd} = -\frac{\alpha}{\gamma} f_n - \frac{\alpha}{2\gamma} \ln(2\pi e \Delta).
\end{equation} 
Note that the choice of normalising by $kd$ is not accidental. Indeed, the number of parameters is $kd+k\approx kd$. Hence with this choice one can interpret the parameter $\alpha$ as an effective signal-to-noise ratio. 

\begin{figure}[t]
\vskip 0.2in
\begin{center}
\centerline{\includegraphics[width=.5\columnwidth]{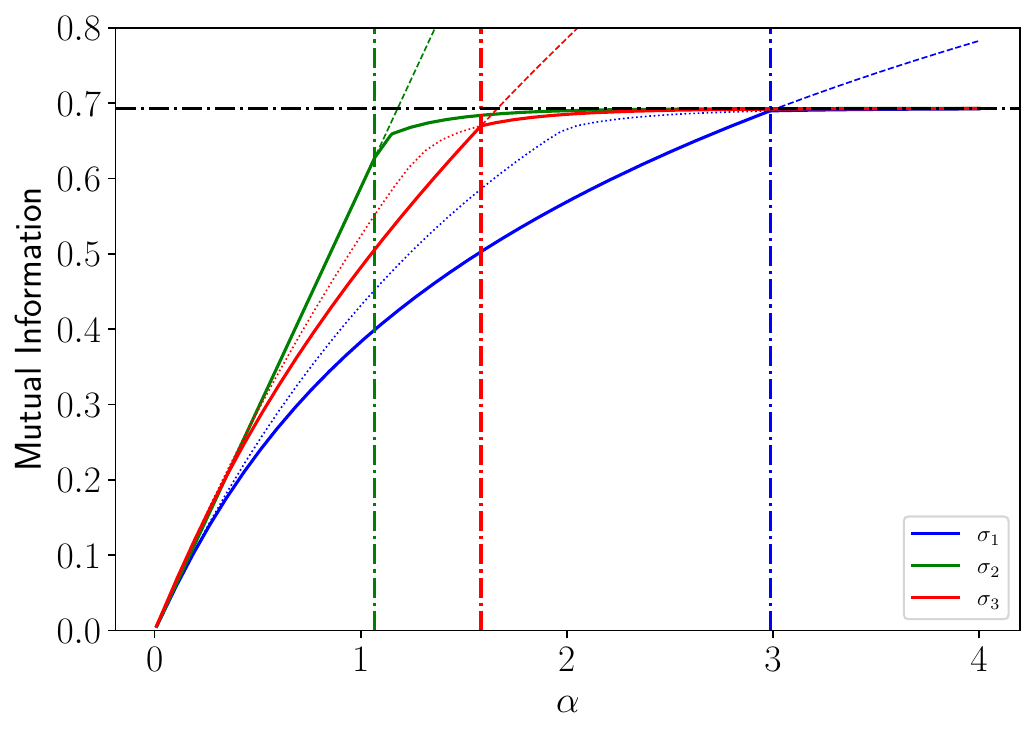}}
\caption{Mutual information in the universal (dashed) and specialisation (dotted) phase, for the activation functions in Fig.~\ref{fig:gen_errors_univ_spec} (top panel), $\gamma=0.5$ and Gaussian label noise with variance $\Delta=1.25$. $\alpha_{\rm sp}$ (dash-dot line) is the point where they cross (same as criterion~\eqref{eq:alpha_c} in the main, due to the relationship between free entropy and mutual information). The continuous line represents the stable (equilibrium) branch. The horizontal black dash-dot line at $\ln 2$ corresponds to the upper bound on the mutual information per parameter for Rademacher inner weights, as proven in \citet{barbier2024phase}, towards which the mutual information converges when $\alpha\to \infty$.}
 \label{fig:MIcrossing}
\end{center}
\vskip -0.2in
\end{figure}

Thanks to the relation between free entropy and mutual information, and using the theory devised in the main, we are able to approximate the mutual information in the universal and specialisation phases, identifying the critical value of $\alpha$ where the transition between the two occurs (see Eq.~\eqref{eq:alpha_c}). In Fig.~\ref{fig:MIcrossing} we report the curves we used to obtain $\alpha_{\rm sp}$ in Fig.~\ref{fig:gen_errors_univ_spec}, top panel. 

\begin{remark}
 The arguments of \citet{barbier2024phase} to show the existence of an upper bound on the mutual information per variable in the case of discrete variables and the associated inevitable breaking of prior universality beyond a certain threshold in matrix denoising apply to the present model too. It implies, as in the aforementioned paper, that the mutual information per variable cannot go beyond $\ln 2$ for Rademacher inner weights. Our theory is consistent with this fact as emphasised by the vertical line in Fig.~\ref{fig:MIcrossing}.
\end{remark}

\section{Details of the replica calculation\label{app:replicas}}
We report here the details of the replica calculation we sketched in the main text, both for the universal and the specialisation phases. The common starting point is \eqref{eq:Zn_last_common}. The energetic potential $F_E$ in \eqref{eq:FE} has always the same form in the two approaches, while the entropic terms will depend on the phase. We shall thus treat them separately.

\subsection{Energetic potential}\label{app:energetic_potential}
The replicated energetic term under our Gaussian assumption on the joint law of the post-activations replicas is reported here for the reader's convenience:
\begin{equation}
    F_E = \ln \int dy\int d\blambda\frac{e^{-\frac{1}{2}\blambda^\intercal\bK^{-1}\blambda}}{\sqrt{(2\pi)^{s+1}\det \bK}} \prod_{a=0}^sP_{\rm out}(y\mid\lambda^a).
\end{equation}
After applying our ansatz \eqref{eq:Omega_ansatz} and using that $Q_1^{ab}=1$ in the quadratic-data regime, the covariance matrix $\bK$ in replica space defined in \eqref{eq:K} reads
\begin{align}
\label{eq:K_RS}
    \bK &= \mu_1^2+ \frac{\mu_2^2}{2}\bQ_2 + \bQ_v \circ g(\bQ_W),
\end{align}
where the function
\begin{equation}
\label{eq:g_func}
    g(x) = \sum_{\ell = 3}^\infty \frac{\mu_{\ell}^2}{\ell !} x^\ell = 
     \E_{(y,z)|x} [\sigma(y)\sigma(z)] - \mu_0^2-\mu_1^2 x - \frac{\mu_2^2}{2}x^2, \qquad (y,z) \sim \calN\left((0,0),\begin{pmatrix}1&x\\x&1\end{pmatrix} \right),
\end{equation}
is applied component-wise to the matrix elements of $\bQ_W$, and $\circ$ is the Hadamard product. $F_E$ is already expressed as a low-dimensional integral, but within the replica symmetric (RS) ansatz it simplifies considerably. The RS ansatz amounts to assume that the saddle point solutions are dominated by order parameters of the form (below $\bm{1}_s$ and $\I_s$ are the all-ones vector and identity matrix of size $s$)
\begin{equation}
\label{eq:RS_W}
     \bQ_W = \begin{pmatrix}
        \rho_W & m_W \bm{1}_s^\intercal\\
        m_W\bm{1}_s & (r_W - q_W) \I_s + q_W \bm{1}_s \bm{1}_s^\intercal 
    \end{pmatrix} \iff
    \hat{\bQ}_W = \begin{pmatrix}
        \hat{\rho}_W & -\hat{m}_W \bm{1}_s^\intercal\\
        -\hat{m}_W\bm{1}_s & (\hat{r}_W + \hat{q}_W) \I_s - \hat{q}_W \bm{1}_s \bm{1}_s^\intercal 
    \end{pmatrix},   
\end{equation}
\begin{equation}
\label{eq:RS_Q2}
     \bQ_2 = \begin{pmatrix}
        \rho_2 & m_2 \bm{1}_s^\intercal\\
        m_2\bm{1}_s & (r_2 - q_2) \I_s + q_2 \bm{1}_s \bm{1}_s^\intercal 
    \end{pmatrix} \iff
    \hat{\bQ}_2 = \begin{pmatrix}
        \hat{\rho}_2 & -\hat{m}_2 \bm{1}_s^\intercal\\
        -\hat{m}_2\bm{1}_s & (\hat{r}_2 + \hat{q}_2) \I_s - \hat{q}_2 \bm{1}_s \bm{1}_s^\intercal 
    \end{pmatrix}, 
    \end{equation}
    \begin{equation}
    \bQ_v = \begin{pmatrix}
        \rho_v & m_v \bm{1}_s^\intercal\\
        m_v\bm{1}_s & (r_v - q_v) \I_s + q_v \bm{1}_s \bm{1}_s^\intercal 
    \end{pmatrix} \iff
        \hat{\bQ}_v = \begin{pmatrix}
        \hat{\rho}_v & -\hat{m}_v \bm{1}_s^\intercal\\
        -\hat{m}_v\bm{1}_s & (\hat{r}_v + \hat{q}_v) \I_s - \hat{q}_v \bm{1}_s \bm{1}_s^\intercal 
    \end{pmatrix}, 
\end{equation}where we reported the ansatz also for the Fourier conjugates for future convenience, though not needed for the energetic potential. The RS ansatz, which is equivalent to an assumption of concentration of the order parameters, is known to be asymptotically exact in the large system limit when analysing Bayes-optimal inference and learning, as in the present paper, see \cite{nishimori2001statistical,barbier2021overlap,barbier2022strong}. Under the RS ansatz $\bK$ acquires a similar form:
\begin{align}
    \bK=\begin{pmatrix}
        \rho_K & m_K \bm{1}_s^\intercal\\
        m_K\bm{1}_s & (r_K - q_K) \I_s + q_K \bm{1}_s \bm{1}_s^\intercal 
    \end{pmatrix}
\end{align}with 
\begin{align}\label{eq:m_K-q_K-def}
    &m_K=\mu_1^2+ \frac{\mu_2^2}{2}m_2 + m_v  g(m_W),\quad &q_K=\mu_1^2+ \frac{\mu_2^2}{2}q_2 + q_v  g(q_W), \\
    \label{eq:rho_K-r_K-def}
    &\rho_K=\mu_1^2+ \frac{\mu_2^2}{2}\rho_2 + \rho_v g(\rho_W),\quad &r_K=\mu_1^2+ \frac{\mu_2^2}{2}r_2 + r_v g(r_W).
\end{align}
In the RS ansatz it is thus possible to give a convenient low-dimensional representation of the multivariate Gaussian integral of $F_E$ in terms of white Gaussians:
\begin{align}
    &\lambda^a=\xi\sqrt{q_K}+u^a\sqrt{r_K-q_K}\quad \text{for }a=1,\dots,s,\qquad \lambda^0=\xi\sqrt{\frac{m_K^2}{q_K}}+u^0\sqrt{\rho_K-\frac{m_K^2}{q_K}}
\end{align}where $\xi,(u^a)_{a=0}^s$ are all i.i.d. standard Gaussian variables. Then
\begin{align}
    F_E=\ln\int dy \,\EE_{\xi,u^0}P_{\rm out}\Big(y\mid \xi\sqrt{\frac{m_K^2}{q_K}}+u^0\sqrt{\rho_K-\frac{m_K^2}{q_K}}\Big)\prod_{a=1}^s\EE_{u^a}P_{\rm out}(y\mid \xi\sqrt{q_K}+u^a\sqrt{r_K-q_K}).
\end{align}
The last product over the replica index $a$ contains all identical factors thanks to the RS ansatz, therefore, by expanding for $s\to 0$ we get
\begin{align}\label{eq:energetic_pot_RS_general}
    F_E=s\int dy\, \EE_{\xi,u^0}P_{\rm out}\Big(y\mid \xi\sqrt{\frac{m_K^2}{q_K}}+u^0\sqrt{\rho_K-\frac{m_K^2}{q_K}}\Big)\ln \EE_{u}P_{\rm out}(y\mid \xi\sqrt{q_K}+u\sqrt{r_K-q_K})+O(s^2).
\end{align}
For the Gaussian channel $P_{\rm out}(y\mid\lambda)=\exp(-\frac1{2\Delta}(y-\lambda)^2)/\sqrt{2\pi\Delta}$ the above gives
\begin{align}\label{eq:F_E_GaussChann}
    F_E=-\frac{s}{2}\ln\big[2\pi(\Delta+r_K-q_K)\big]-\frac{s}{2}\frac{\Delta+\rho_K-2m_K+q_K}{\Delta+r_K-q_K}+O(s^2).
\end{align}
In the Bayes-optimal setting the Nishimori identities enforce 
\begin{align}\label{eq:nishi_r_2}
 &r_2=\rho_2 = \lim_{d\to\infty}\frac{1}{d^2}\EE\Tr[(\bS_2^0)^2]= 1 + \gamma (\E v^0)^2,\\ &r_v=\rho_v=r_W=\rho_W=1, \\ 
 \label{eq:nishi_q2}
 &m_2=q_2, \quad m_v=q_v, \quad m_W=q_W,
\end{align}
which implies also that
\begin{align}
r_K=\rho_K=\mu_1^2+\frac12r_2\mu_2^2 +g(1) ,\,\quad 
 m_K=q_K.
\end{align}
Therefore the above simplifies to
\begin{align}\label{eq:energetic_pot_RS_BayesOpt}
    F_E&=s\int dy\, \EE_{\xi,u^0}P_{\rm out}(y\mid \xi\sqrt{q_K}+u^0\sqrt{r_K-q_K}\Big)\ln \EE_{u}P_{\rm out}(y\mid \xi\sqrt{q_K}+u\sqrt{r_K-q_K})+O(s^2)\\
    &=:s \, \psi_{P_{\rm{out}}}(q_K(q_2,q_W,q_v);r_K)+O(s^2).\label{eq:defPsiPout}
\end{align}
Notice that the energetic contribution to the free entropy has the same form as in the generalised linear model \cite{barbier2019glm}. For our running example of Gaussian output channel 
the function $\psi_{P_{\rm out}}$ reduces to
\begin{align}\label{eq:F_E_BayesOpt_GaussChann}
     \psi_{P_{\rm{out}}}(q_K(q_2,q_W,q_v);r_K)=-\frac{1}{2}\ln\big[2\pi(\Delta+r_K-q_K)\big]-\frac{1}{2}.
\end{align}
In what follows we shall restrict ourselves only to the replica symmetric ansatz, in the Bayes-optimal setting. Therefore, identifications as the ones in \eqref{eq:nishi_r_2}-\eqref{eq:nishi_q2} are assumed.

\subsection{Free entropy and mutual information for the universal phase}
Let us take the computation from \eqref{eq:S2_universal}. When the number of data $n$ is sent to $+\infty$, the integral over the order parameters in \eqref{eq:Zn_last_common} is dominated by the saddle points w.r.t. $\bQ_2,\bQ_W,\bQ_v$. As anticipated $Q_W^{ab}=\delta_{ab}$, i.e., $m_W=q_W=0$, and consequently only $Q_v^{aa}=1$ appears in the expression due to \eqref{eq:m_K-q_K-def} and $g(0)=0$. The only nontrivial saddle point is thus performed over the order parameter $\bQ_2=(Q_2^{ab})_{0\leq a\leq b\leq s}$. Therefore, in the thermodynamic limit the leading order contribution to the replicated free entropy reads
\begin{align}\label{eq:replicated_freent_universal_1}
    f_{n,s}=\text{extr}\Big\{
    \frac{1}{s}F_E(\bQ_2)+\frac{kd}{ns}\ln V_W(\I_{s+1})+\frac{1}{ns}\ln\int \prod_{a=0}^sdP(\bS^a_2)\prod_{a\leq b,0}^s\delta\big(d^2Q_2^{ab}-\Tr[\bS^a_2\bS_2^{b\intercal}]\big)
    \Big\}+o(1),
\end{align}where we have abused the notation $F_E(\bQ_2):= F_E(\bQ_1=\mathbf{1}\mathbf{1}^\intercal,\bQ_2,\I_{s+1},\I_{s+1})$. Extremisation is over $\bQ_2$ only. 

The only high-dimensional part remaining is that of the variables $(\bS_2^a)_{0\leq a\leq s}$. Using the Fourier representation of the delta function\footnote{In this manuscript, we often represent the delta function using its Fourier representation $\delta(x-c)=\frac1{2\pi }\int_{i\mathbb{R}} d\hat x \exp(\hat x(x-c))$. Formally the integration is over the imaginary axis $i\mathbb{R}$. The complex-valued Fourier conjugates $\hat x$ associated with order parameters will enter effective actions and the final integrals will be performed by saddle-point through contour deformation in $\mathbb{C}$. In inference problems, saddle-point integration will always pick real-valued parameters for all the integrated quantities, including Fourier conjugates. Therefore, we will never specify that integrals over Fourier parameters are over $i\mathbb{R}$. Moreover, trivial multiplicative constants such as the $1/2\pi$ appearing here play no role in the final equations, and will therefore be dropped without notice.}, the last term in \eqref{eq:replicated_freent_universal_1} rewrites as
\begin{align}
    J_{n,s}(\bQ_2):=\frac{1}{ns}\ln\int\prod_{a\leq b,0}^s d\hat Q_2^{ab}\exp\Big[-\frac{d^2}{4}\sum_{a,b=0}^sQ_2^{ab}\hat Q_2^{ab}\Big]\int\prod_{a=0}^s dP(\bS_2^a)\exp\Big[
    \frac{d}{4}\sum_{a,b=0}^s\hat{Q}^{ab}_2\Tr\Big(\frac{\bS_2^a}{\sqrt{d}}\frac{\bS_{2}^{b\intercal}}{\sqrt{d}}\Big)
    \Big]
\end{align}up to vanishing corrections. Notice that we have re-normalised $\bS_2^a$ by $\sqrt{d}$ in order to work with matrices with $O(1)$ eigenvalues. Using the Bayes-optimality of the setting, we can perform an additional simplifying RS ansatz on the saddle point optimisation: 
\begin{align}
&Q_2^{aa}=r_2, \quad 0\leq a\leq s, \quad \text{and} \quad Q_2^{ab}=q_2, \quad a\neq b,\\
&\hat Q_2^{aa}=-\hat r_2,\quad 0\leq a\leq s, \quad \text{and} \quad \hat Q_2^{ab}=\hat q_2, \quad a\neq b.
\end{align}
Therefore, $J_{n,s}$ at leading order in $n$ appears as
\begin{align}
    J_{n,s}(q_2,r_2)=\text{extr}\Big\{&\frac{1}{ns}\ln\int\prod_{a=0}^s dP(\bS_2^a)\exp\Big(
    -\frac{d (\hat{r}_2+\hat q_2)}{4}\sum_{a=0}^s\Tr\Big[\Big(\frac{\bS_2^a}{\sqrt{d}}\Big)^2\Big]+\frac{d\hat{q}_2}{4}\Tr\Big[\Big(\sum_{a=0}^s\frac{\bS_2^a}{\sqrt{d}}\Big)^2\Big]
    \Big)\nonumber\\
    &+
    \frac{d^2}{4ns}\big((s+1)\hat r_2 r_2-s(s+1)\hat q_2 q_2\big)
    \Big\},
\end{align}where extremisation is w.r.t.\ $\hat r_2$ and $\hat q_2$. From the above it is clear that when $s\to0$, it must be the case that $\hat r_2$ vanish (otherwise a divergence appears). This happens because the Nishimori identities in the Bayes-optimal setting are indeed sufficient to fix the values of $Q_2^{aa}=r_2$ without the need of Fourier conjugates. In order to take the $0$ replica limit, one then decouples replicas with a Hubbard-Stratonovich transformation which introduces an expectation over a standard GOE matrix $\bZ\in\mathbb{R}^{d\times d}$ with $O(1)$ eigenvalues (i.e., a symmetric matrix whose upper triangular part has i.i.d. entries from $\mathcal{N}(0,(1+\delta_{ij})/d)$) through the identity
\begin{align*}
    \E_{\bZ} \, e^{\frac d 2 \Tr[\bM \bZ]} = e^{\frac d 4 \Tr[\bM^2]}
\end{align*}
for any symmetric matrix $\bM \in \mathbb{R}^{d\times d}$. After these standard steps the replica limit of $J_{n,s}$ reads
\begin{align}
    J_{n,0}(q_2)=\text{extr}\Big\{&\frac{1}{n}\EE_{\bar\bS_2^0,\bZ}\ln\int dP(\bar{\bS}_2)\exp\Big(
    -\frac{d \hat q_2}{4}\Tr\bar{\bS}_2^2+
    \frac{d\sqrt{\hat{q}_2}}{2} \Tr\big[\bar{\bS}_2\big(\sqrt{\hat{q}_2}\bar{\bS}^0_2+\bZ\big)\big]
    \Big)-\frac{1}{4\alpha}\hat q_2 q_2
    \Big\}
\end{align}where  $\bar{\bS}_2:=\frac{1}{\sqrt{kd}}\sum_{i=1}^k v_i\bW_i\bW_i^\intercal=\frac{\sqrt{\gamma}}{k}\sum_{i=1}^k v_i\bW_i\bW_i^\intercal$ and similarly for $\bar{\bS}_2^0$. The high-dimensional integral that remains is the free entropy per datum of a Bayes-optimal matrix denoising problem:
\begin{align}\label{eq:}
    \bY(\hat q_2)=\sqrt{\hat q_2}\,\bar{\bS}^0_2+\bZ,
\end{align}with a rotationally invariant prior on $\bar{\bS}^0_2$. Therefore, we can directly import the results from \cite{matrix_inference_Barbier,maillard2024bayes}:
\begin{align}
    J_{n,0}(q_2)=\frac{1}{\alpha}\text{extr}\Big\{\frac{\hat q_2 (r_2-q_2)}{4}-\iota(\hat q_2)
    \Big\}
\end{align}
where we remind the reader that $r_2 = 1 + \gamma (\E v^0)^2$, and
\begin{align}
    \iota(\hat q_2):=\lim_{d\to\infty}\frac{1}{d^2}I(\bar\bS_2^0;\bY(\hat q_2))=\frac{1}{8}+\frac{1}{2}\int \ln|x-y|\,d\mu_{\bY(\hat q_2)}(x)d\mu_{\bY(\hat q_2)}(y).
\end{align} Here $I(\bar\bS_2^0;\bY(\hat q_2))$ is the MI related to the channel \eqref{eq:}, $\mu_{\bY(\hat q_2)}$ is the asymptotic spectral law of the observation matrix $\bY(\hat q_2)$. Extremisation is w.r.t.\ $\hat{q}_2$ only.

The other quantity to simplify is the entropic contribution $\ln V_W(\I_{s+1})$. It is not difficult to verify that when the matrix overlap in the argument is the identity this contribution is vanishing. The intuitive reason is that in that case the $\delta$'s in the integral defining $V_W$ are imposing constraints that are already approximately verified by samples from the prior $P_W$ with high probability. Therefore, integration w.r.t.\ the prior of these constraints is virtually still measuring the whole probability space, yielding $V_W(\I_{s+1})=1$ at leading exponential order.

Furthermore, by \eqref{eq:K_RS}, in this phase we have
\begin{align}
    q_K=q_K(q_2,0,0)=\mu_1^2+\frac{\mu_2^2}{2}q_2,\qquad r_K= \mu_1^2+\frac{\mu_2^2}{2}( 1 + \gamma (\E v^0)^2) + g(1).
\end{align}

Hence, the final replica symmetric potential in the universal phase reads
\begin{align}\label{eq:free_ent_univ_general}
    f_{\rm uni}=\text{extr}\Big\{\psi_{P_{\rm{out}}}(q_K(q_2,0,0);r_K)+\frac{\hat q_2(r_2-q_2)}{4\alpha}-\frac{1}{\alpha}\iota(\hat q_2)\Big\},
\end{align}
while the mutual information (see App.~\ref{app:mutual_info}) is
\begin{align}\label{eq:MI_universal_phase}
    I_{\rm uni}=-\frac{\alpha}{\gamma}f_{\rm uni}+\frac{\alpha}{\gamma} \EE_{\lambda}  \int dy P_{\text{out}}(y|\lambda) \ln P_{\text{out}}(y|\lambda).
\end{align}
In our running example of Gaussian channel $P_{\rm out}$ with noise intensity $\Delta$, the above reads
\begin{align}\label{eq:free_nrg_universal_phase}
    f_{\rm uni}=\text{extr}\Big\{-\frac{1}{2}\ln(2 \pi)-\frac{1}{2}\ln(\Delta +r_K-\mu_1^2-\frac{\mu_2^2}{2}q_2)-\frac{1}{2} +\frac{\hat q_2(r_2-q_2)}{4\alpha}-\frac{1}{\alpha}\iota(\hat q_2)\Big\}
\end{align}
and the mutual information is
\begin{equation}
    I_{\rm uni} = \text{extr}\Big\{\frac{\alpha}{2\gamma}\ln(\Delta +r_K-\mu_1^2-\frac{\mu_2^2}{2}q_2) -\frac{\hat q_2(r_2-q_2)}{4 \gamma}+\frac{1}{\gamma}\iota(\hat q_2)\Big\} - \frac{\alpha}{2 \gamma} \ln \Delta.
\end{equation}
For a generic output channel the system of saddle point equations read
\begin{align}\label{eq:Suni}
({\rm S_{uni}})
\begin{sqcases}
    &q_2=r_2-\frac{1}{\hat q_2}(1-\frac{4\pi^2}{3}\int \mu^3_{\bY(\hat q_2)}(y)dy),\\ 
    &\hat q_2=4\alpha\,\partial_{q_2}\psi_{P_{\rm out}}(q_K(q_2,0,0);r_K).
    \end{sqcases}
\end{align}
Only the second equation is channel-dependent. For a Gaussian output channel $P_{\rm out}(y\mid \lambda)=\exp(-\frac1{2\Delta}(y-\lambda)^2)/\sqrt{2\pi \Delta}$ we have
\begin{align}
({\rm S_{uni}})
\begin{sqcases}
   & q_2=r_2-\frac{1}{\hat q_2}(1-\frac{4\pi^2}{3}\int \mu^3_{\bY(\hat q_2)}(y)dy),\\
    &\hat q_2=
   \alpha\mu_2^2/[\Delta+r_K-\mu_1^2-q_2\mu_2^2/2].
     \end{sqcases} 
\end{align}

\subsection{Free entropy and mutual information for the specialisation phase}
Following from the ansatz \eqref{eq:S2_specialisation}, the replicated partition function for the specialisation phase
reads (again, equality here is at leading exponential order and we already took $\bQ_1$ and $\bQ_v$ as all-ones matrices, and used a Fourier representation for the delta function fixing $\bQ_2$ in \eqref{eq:Zn_last_common})
\begin{align}
    \EE\mathcal{Z}^s (\mathcal{D})&=\int \prod_{a\leq b,0}^s dQ_2^{ab}d\hat{Q}_2^{ab}dQ_W^{ab}\, e^{n F_E(\bQ_1\to \bm{1}\bm{1}^\intercal,\bQ_2,\bQ_W,\bQ_v\to \bm{1}\bm{1}^\intercal)+kd\ln V_W(\bQ_W) +\frac{d^2}{4}\Tr\hat \bQ_2\bQ_2^\intercal}\nonumber\\
    &\qquad\times\Big[\int \prod_{a=0}^s dS^a_2\,\frac{1}{\sqrt{(2\pi )^{s+1}\det( \bQ_W^{\circ2})}}
    e^{-\frac{1}{2}\sum_{a,b=0}^s S^{a}_2(\bQ_W^{\circ 2})^{-1}_{ab} S^{b}_2-\frac{1}{2}\sum_{a,b=0}^s\hat Q_2^{ab}S^a_2S^b_2}
    \Big]^{d(d-1)/2}\nonumber\\
    &\qquad\qquad\qquad\times \int\Big(\prod_{a=0}^s\prod_{\alpha=1}^d dS^a_{2;\alpha\alpha} \delta(S^a_{2;\alpha\alpha}-\sqrt{k}(\EE v))\Big)e^{-\frac{1}{4}\sum_{a,b=0}^s\hat Q_2^{ab}\sum_{\alpha=1}^dS_{2;\alpha\alpha}^aS_{2;\alpha\alpha}^b}.
\end{align}
Integration over the diagonal elements $(S_{2;\alpha\alpha}^a)_{\alpha}$ can be done straightforwardly, yielding
\begin{align}
    \EE\mathcal{Z}^s (\mathcal{D})&=\int \prod_{a\leq b,0}^s dQ_2^{ab}d\hat{Q}_2^{ab}dQ_W^{ab} \,e^{n F_E(\bQ_1\to \bm{1}\bm{1}^\intercal,\bQ_2,\bQ_W,\bQ_v\to \bm{1}\bm{1}^\intercal)+kd\ln V_W(\bQ_W) +\frac{d^2}{4}\Tr\hat \bQ_2^\intercal(\bQ_2-\gamma\mathbf{1}\mathbf{1}^\intercal(\EE v)^2)}\nonumber\\
    &\qquad\times\Big[\int \prod_{a=0}^s dS^a_2\,\frac{1}{\sqrt{(2\pi )^{s+1}\det( \bQ_W^{\circ2})}}
    e^{-\frac{1}{2}\sum_{a,b=0}^s S^{a}_2(\bQ_W^{\circ 2})^{-1}_{ab} S^{b}_2-\frac{1}{2}\sum_{a,b=0}^s S^a_2\hat Q_2^{ab}S^b_2}
    \Big]^{d(d-1)/2}.
\end{align}
Using the change of variable $\bS_2\to (\bQ_W^{\circ 2})^{1/2}\bS_2$ (where $(\,\cdot\,)^{1/2}$ is a matrix square root), the remaining Gaussian integral over the off-diagonal elements of $\bS_2$ can be performed exactly, leading to
\begin{align}
\label{eq:exp_Z^n}
    \EE\mathcal{Z}^s(\mathcal{D})&=\int \prod_{a\leq b,0}^s dQ_2^{ab}d\hat{Q}_2^{ab}dQ_W^{ab}  \,  e^{nF_E(\bQ_2,\bQ_W)+kd\ln V_W(\bQ_W)+\frac{d^2}{4}\Tr\hat{\bQ}_2^\intercal(\bQ_2 {-\gamma \bm{1}\bm{1}^\intercal (\E v)^2})-\frac{d(d-1)}{4}\ln\det[\I_{s+1}+\hat\bQ_2 \bQ_W^{\circ2}]}
\end{align}
where, being in the Bayes-optimal setting we can assume $\bQ_1\to \bm{1}\bm{1}^\intercal$, $\bQ_v\to \bm{1}\bm{1}^\intercal$, and therefore $F_E(\bQ_2,\bQ_W):=F_E(\bQ_1\to \bm{1}\bm{1}^\intercal,\bQ_2,\bQ_W,\bQ_v\to \bm{1}\bm{1}^\intercal)$. 
In order to proceed and perform the $s\to 0^+$ limit, we use the RS ansatz for $\bQ_2$ and $\bQ_W$ introduced in \eqref{eq:RS_W} and \eqref{eq:RS_Q2}, combined with the Nishimori identities
, combined with the Nishimori identities
\begin{align}
    \begin{aligned}
        &m_W=q_W, \,r_W=\rho_W=1,\,r_2=\rho_2=1+\gamma(\EE v)^2,\\
        &\hat m_W=\hat q_W, \,\hat r_W=\hat \rho_W=0,\,\hat r_2=\hat \rho_2=0.
    \end{aligned}
\end{align}
We can start by evaluating the normalisation factor $V_W(\bQ_W)$ by representing the delta function fixing $\bQ_W$ in Fourier space and introducing its conjugate variable $\hat{\bQ}_W$ (both are symmetric matrices), so that 
\begin{equation}
\begin{aligned}
    V_W(\bQ_W)^{kd} = & \int d \hat{\bQ}_W e^{\frac{kd}{2} \Tr\bQ_W \hat{\bQ}_W}\Big[ \int \prod_{a=0}^s dP_W(w^a) \exp\Big(-\frac{1}{2}\sum_{a,b=0}^s w^a w^b \hat{Q}_W^{ab} \Big) \Big]^{kd} \\
    = & \int d \hat{\bQ}_W e^{\frac{kd}{2} \Tr\bQ_W \hat{\bQ}_W}\Big( \mathbb{E}_{w^0,\xi_w}\Big( \E_w \Big[e^{-\frac{\Hat{q}_W}{2}w^2 + \Hat{q}_W w^0 w + \sqrt{\Hat{q}_W} \xi_w w}\Big]\Big)^s\Big)^{kd},
\end{aligned}
\end{equation}
where in the second line we used the Hubbard-Stratonovich transformation, introduced $\xi_w \sim \mathcal{N}(0,1)$ and $w,w^0\sim P_W$.
Note that we reduced the matrix integral into a one-dimensional integral over a single element of the weight matrix $\bW$ using the factorisation of its prior law. 

With the RS ansatz \eqref{eq:RS_W}, \eqref{eq:RS_Q2} and the Nishimori identities computing traces is straightforward:
\begin{equation}
\begin{aligned}
    \Tr\bQ_W \hat{\bQ}_W =
    -s(s+1)\hat q_W q_W, \qquad \Tr\bQ_2 \hat{\bQ}_2 = 
    -s(s+1)\hat q_2 q_2.
\end{aligned}
\end{equation}
Finally, the determinant term in the exponent of the integrand of  \eqref{eq:exp_Z^n} reads
\begin{align}
\ln\det[\I_{s+1}+\hat\bQ_2\bQ_W^{\circ2}] = 
s\ln [1+\hat{q}_2 (1-q_W^2)]-s\hat{q}_2
+O(s^2).
\end{align}
All the terms that appear in the exponent of \eqref{eq:exp_Z^n} are now explicit.
In order to proceed with the calculations one should switch the limit in $s \to 0^+$ and $n,k,d \to \infty$ and compute the integrals with the saddle point approximation. These are standard procedures in a replica calculation, we thus report the result (which, as we recall, holds in the Bayes-optimal setting):
\begin{equation}
\begin{aligned}
\label{freeEntSpecialisation-Bayes-Opt}
    f_{\rm sp} = {\text{extr}} \Big\{& \frac{\gamma}{\alpha} \psi_{P_W}(\hat q_W) 
    +\psi_{P_{\text{out}}}(q_K(q_2,q_W,1);r_K) 
    -\frac{\gamma}{2\alpha}q_W \Hat{q}_W + \frac{(r_2-q_2) \Hat{q}_2}{4\alpha}
    - \frac{1}{4\alpha}\ln [ 1+\hat q_2   (1 - q_W^2) ]
    \Big\} ,
\end{aligned}    
\end{equation}where $\psi_{P_{\rm out}}$ is given by~\eqref{eq:defPsiPout}, extremisation is w.r.t.\ $\hat q_W,\hat q_2, q_W,q_2$, and
\begin{equation}
\label{eq:psi_p_w}
    \psi_{P_W}(\hat q_W) := \mathbb{E}_{w^0,\xi_w} \ln\E_w \Big[e^{-\frac{\Hat{q}_W}{2}w^2 + \Hat{q}_W w^0 w + \sqrt{\Hat{q}_W} \xi_w w}\Big].
\end{equation}
The mutual information follows from~\eqref{eq:MI_generic_channel}:
\begin{equation}
\label{eq:MI_specialisation_phase}
    I_{\rm sp} = -\frac{\alpha}{\gamma} f_{\rm sp} + \frac{\alpha}{\gamma} \EE_{\lambda}  \int dy P_{\text{out}}(y|\lambda) \ln P_{\text{out}}(y|\lambda).
\end{equation} 
With the shortcut notation
\begin{equation}
    \thav{\,\cdot\,}_{\Hat{q}_W}=\thav{\,\cdot\,}_{\Hat{q}_W}(\xi_w,w_0)  :=\frac{\int dP_W(x) (\,\cdot\,)e^{-\frac{\Hat{q}_W}{2}x^2 + (\Hat{q}_W w^0 + \sqrt{\Hat{q}_W} \xi_w)x}}
    {\int dP_W(y) e^{-\frac{\Hat{q}_W}{2}y^2 + (\Hat{q}_W w^0 + \sqrt{\Hat{q}_W} \xi_w)y}},
\end{equation}
the resulting saddle point equations therefore read
\begin{equation}
\label{FP_equations_generic_ch}
(\rm{S_{sp}})
\begin{sqcases}
& q_W = \EE_{w^0,\xi_w} [ w^0 \thav{x}_{\Hat{q}_W} ], \\
& \hat q_W =  {\hat{q}_2 q_W}/[\gamma+\gamma\hat{q}_2(1-q_W^2)] + 2\frac{\alpha}{\gamma} \partial_{q_W} \psi_{P_{\text{out}}}(q_K(q_2,q_W,1);r_K) , \\
& q_2= r_2 - (1-q_W^2)/[1 + \hat q_2(1-q_W^2)] , \\
& \hat q_2 = 4\alpha \,\partial_{q_2} \psi_{P_{\text{out}}}(q_K(q_2,q_W,1);r_K) .
\end{sqcases}   
\end{equation}

All the above formulae are easily specialised for the Gaussian output channel case using \eqref{eq:F_E_BayesOpt_GaussChann}. We report here, for the reader's convenience, the saddle point equations in such setting (recalling that $g$ is defined in \eqref{eq:g_func}):
\begin{equation}
(\rm{S_{sp}})
\begin{sqcases}
\label{FP_equations_gaussian_ch}
& q_W = \EE_{w^0,\xi_w} [ w^0 \thav{x}_{\Hat{q}_W} ], \\
& \hat q_W =  \hat{q}_2 q_W/[{\gamma+\gamma\hat{q}_2(1-q_W^2)}] + \frac{\alpha}{\gamma} {g'(q_W)}/[\Delta + \frac{\mu_2^2}{2}(r_2-q_2) + g(1) - g(q_W)] , \\
& q_2= r_2 - (1-q_W^2)/[1 + \hat q_2(1-q_W^2)] , \\
& \hat q_2 =  {\alpha\mu_2^2}/[\Delta + \frac{\mu_2^2}{2}(r_2-q_2) + g(1) - g(q_W)] .
\end{sqcases}    
\end{equation}

If one assumes that the the overlaps appearing in \eqref{eq:simple_gen_error_for_numerics} are self-averaging around the values that solve the saddle point equations, that is $Q^{00}_1,Q_1^{01}\to1$ (as assumed in this scaling), $Q_2^{00}\to r_2, Q_2^{01}\to q_2$, and $Q_W^{00}\to1,Q_W^{01}\to q_W$, then the limiting Bayes-optimal generalisation error for the Gaussian channel case appears as
\begin{equation}
\begin{aligned}
    \varepsilon^{\rm opt}-\Delta = \frac{\mu_2^2}{2}(r_2-q_2) + \big(g(1) - g(q_W)\big).
\end{aligned}    \label{eq:Gaussian_output_channel_generror}
\end{equation}

\section{Non-centred activations\label{app:non-centered}}
In this section we consider the generic case in which the activation function in \eqref{eq:sigma_hermite} is non-centred, i.e., $\mu_0\neq 0$. This reflects on the law of the post-activations, which will still be Gaussian, centred at
\begin{align}
    \EE\lambda^a=\frac{\mu_0}{\sqrt
    k}\sum_{i=1}^kv^a_i=:\mu_0
    \Lambda^a,
\end{align}and with the covariance given by \eqref{eq:K} (we are assuming $ Q_{W}^{aa}=1$; if not, $ Q_{W}^{aa}=r$, the formula can be generalised as explained in App.~\ref{app:hermite}). In the above, we have introduced the new mean parameter $\Lambda^a$. Notice that, if the $\bv^0$'s have a $\bar v=O(1)$ mean, then $\Lambda^a$ scales as $\sqrt{k}$ due to our choice of normalisation.

Concerning the energetic potential, it will now appear as
\begin{equation}
    F_E=F_E(\bK,\bLambda) = \ln \int dy\int d\blambda\frac{e^{-\frac{1}{2}\blambda^\intercal\bK^{-1}\blambda}}{\sqrt{(2\pi)^{s+1}\det \bK}} \prod_{a=0}^s P_{\rm{out}}(y\mid \lambda^a+\mu_0 \Lambda^a),
\end{equation}
while in the entropic part we have the additional constraint $\Lambda^a=\sum_iv_i^a/\sqrt{k}$:
\begin{align}
    F_S&(\bQ_v,\bQ_1,\bQ_2,\bQ_W,\bLambda): = 
    \ln 
    \int \prod_{a=0}^s d\bS_1^a d\bS_2^a \int \prod_{a=0}^s dP_v(\bv^a)dP_W(\bW^a)
    \nonumber\\
    &\times\prod_{a=0}^s\delta\Big(\bS^a_2-
    \frac{\bW^{a\intercal}(\bv^a)\bW^a}{\sqrt{k}}\Big)
    \delta\Big(\bS^a_1-\frac{\bv^{a\intercal}\bW^{a}}{\sqrt{k}}\Big)
    \prod_{a\leq b,0}^s\delta\Big(Q_W^{ab}-\frac{1}{kd}\Tr[\bW^a\bW^{b\intercal}]\Big)\delta\Big(Q_v^{ab}-\frac{\bv^{a\intercal} \bv^{b}}{k}\Big)
    \nonumber\\
    & \times\prod_{a\leq b,0}^s\Big[\delta\Big(Q_1^{ab}-\frac{1}{d} \sum_{\alpha=1}^d S_{1;\alpha}^{a} S_{1;\alpha}^b\Big)  \delta\Big(Q_2^{ab}-\frac{1}{d^2} \sum_{\alpha_1,\alpha_2 = 1}^d S_{2;\alpha_1\alpha_2}^{a} S_{2;\alpha_1\alpha_2}^b\Big)\Big]\prod_{a=1}^s\delta\Big(\Lambda^a-\frac{1}{\sqrt{k}}\sum_{i=1}^k v_i^a
    \Big).\label{eq:eFS_appendix}
\end{align}
As already mentioned for centred activations, the Dirac $\delta$'s on $\bQ_1$ and $\bQ_v$ never really contribute to the thermodynamics, as they involve a number of variables of $\Theta(d)=\Theta(k)$, whereas the free energy scales at $\Theta(n)=\Theta(d^2)=\Theta(k^2)$. This is even more so for the few variables $\Lambda^{a\ge 1}$, which are only $\Theta(s)$ in number. Hence, $F_S$ defined in~\eqref{eq:eFS_appendix} collapses on~\eqref{eFS} at leading order. A similar fact was already pointed out in~\citet{gardner1988}. Since $\bLambda$ only appear in the energetic potential $F_E$, their value can be determined by saddle point for $n$ large, as for the other order parameters. In other words, the student can always determine $\Lambda^{a\ge 1}$ from a maximum likelihood estimation at given teacher. 
Therefore, in what follows we can carry out the computation (and the replica trick) for a fixed realisation of $\Lambda^0$.
After saddle point integration, we have
\begin{equation}
    \EE [\mathcal{Z}^s(\mathcal{D},\Lambda^0)\mid \Lambda^0] = 
    \exp\underset{\substack{\Lambda^{a\ge 1},\bQ_1,\bQ_2,\\ \bQ_W,\bQ_v}}{\rm{extr}}\Big(F_S(\bQ_2,\bQ_W)+ nF_E(\bK,\bLambda)
    \Big).
\end{equation}
The treatment for $F_S$ is the same as the one discussed above, while $F_E$ becomes
\begin{align*}
    e^{F_E}=\int dy \,\EE_{\xi,u^0}P_{\rm out}\Big(y\mid \mu_0 \Lambda^0 + \xi\sqrt{\frac{m_K^2}{q_K}}+u^0\sqrt{\rho_K-\frac{m_K^2}{q_K}}\Big)\prod_{a=1}^s\EE_{u^a}P_{\rm out}(y\mid \mu_0 \Lambda + \xi\sqrt{q_K}+u^a\sqrt{r_K-q_K}),
\end{align*}
where we have assumed replica symmetry also in the $\Lambda^{a\ge 1} =:\Lambda$. Therefore, the simplification of the potential $F_E$ proceeds as in the centred activation case, yielding at leading order in the replicas
\begin{align*}
    \frac{F_E(r_K,q_K,\Lambda,\Lambda^0)}{s}\!=\!\int dy\, \EE_{\xi,u^0}P_{\rm out}\Big(y\mid \mu_0 \Lambda^0 + \xi\sqrt{q_K}+u^0\sqrt{r_K-q_K}\Big)\ln \EE_{u}P_{\rm out}(y\mid\mu_0 \Lambda + \xi\sqrt{q_K}+u\sqrt{r_K-q_K})
\end{align*}
in the Bayes-optimal setting. From this equation it is clear that the optimal student's estimate for $\Lambda$ is precisely $\Lambda = \Lambda^0$: indeed, $F_E$ is written in the form of a cross-entropy parametrised by $\Lambda$, and it attains its maximum at this value.

In the case when $P_{\rm out}(y\mid \lambda)=f(y-\lambda)$ then one can verify that the contributions due to the means, containing $\mu_0$, cancel each other. This is verified in our running example where $P_{\rm out}$ is the Gaussian channel:
\begin{equation}
    \frac{F_E(r_K,q_K,\Lambda,\Lambda^0)}{s} = 
    -\frac{1}{2}\ln\big[2\pi(\Delta+r_K-q_K)\big]-\frac{1}{2} - \frac{\mu_0^2}{2}\frac{(\Lambda - \Lambda^0)^2}{\Delta+r_K-q_K},
\end{equation}
which is identical to~\eqref{eq:F_E_BayesOpt_GaussChann} when $\Lambda=\Lambda^0$. We notice that the above arguments hold both with quenched and learnable read-out weights $\bv$.

\section{Equivalence to effective generalised linear models in the universal phase, and extension of GAMP-RIE to arbitrary activation} \label{app:glm}
The saddle point equations for the overlaps in the universal phase can also be derived from the effective equivalence of our model to a generalised linear model (GLM). For simplicity, let us consider $P_{\rm out}(y\mid\lambda)=\exp(-\frac1{2\Delta}(y-\lambda)^2)/\sqrt{2\pi\Delta}$ (these assumptions can be relaxed):
\begin{align}
y_{\mu} \mid (\btheta^0,\bx_\mu) \overset{\rm{d}}{=} \frac{\bv^{0 \intercal}}{\sqrt k} \sigma \left( \frac{\bW^0 \bx_\mu}{\sqrt d} \right) + \sqrt{\Delta} \,z_\mu, \quad \mu =1\dots,n,
\end{align}
where $z_\mu$ are i.i.d. standard Gaussian random variables. Expanding $\sigma$ in the Hermite polynomial basis we have
\begin{align}\label{eq:hexpan}
y_\mu \mid (\btheta^0,\bx_\mu) \overset{\rm{d}}{=} \mu_0\frac{\bv^{0\intercal} \bm{1}_k}{\sqrt{k}}+ \mu_1 \frac{\bv^{0\intercal} \bW^0 \bx_\mu}{\sqrt{kd}} + \frac{\mu_2}{2} \frac{\bv^{0\intercal}}{\sqrt k} \He_2 \left( \frac{\bW^0 \bx_\mu}{\sqrt d} \right) + \dots + \sqrt{\Delta} z_\mu 
\end{align}
where $\dots$ represents the terms beyond second order. 
Without loss of generality, for this choice of output channel we can set $\mu_0 = 0$ as discussed in App.~\ref{app:non-centered}. In the universal phase, the higher order terms in $\dots$ cannot be learned given quadratically many samples and, as a result, play the role of effective noise, which we assume independent of the first three terms. Given that, this noise is asymptotically Gaussian thanks to the central limit theorem (it is a projection of a centred function applied entry-wise to a vector with i.i.d.\ entries), its variance is $g(1)$ (see Eq.~\eqref{eq:g_func}). We thus obtain the effective equivalent model 
\begin{align}
y_\mu \mid (\btheta^0,\bx_\mu) \overset{\rm{d}}{=}  \mu_1 \frac{\bv^{0\intercal} \bW^0 \bx_\mu}{\sqrt{kd}} + \frac{\mu_2}{2} \frac{\bv^{0\intercal}}{\sqrt k} \He_2 \!\left( \frac{\bW^0 \bx_\mu}{\sqrt d} \right) + \sqrt{\Delta + g(1) } \, z_\mu ,
\end{align}
where $\overset{\rm d}{{}={}}$ mean equality in law. The first term in this expression can be learned with vanishing error given quadratically many samples (Remark \ref{rem:linear}), thus can be ignored. This further simplifies the model to
\begin{align}
\bar y_\mu :=  y_\mu - \mu_1 \frac{\bv^{0\intercal} \bW^0 \bx_\mu}{\sqrt{kd}}\overset{\rm d}{{}={}} \frac{\mu_2}{2} \frac{\bv^{0\intercal}}{\sqrt k} \He_2 \!\left( \frac{\bW^0 \bx_\mu}{\sqrt d} \right) + \sqrt{\Delta + g(1) } \, z_\mu,
\end{align}
where $\bar y_\mu$ is $ y_\mu$ with the (asymptotically) perfectly learned linear term removed, and the last equality in distribution is again conditional on $(\btheta^0,\bx_\mu)$. From the formula
\begin{align}
\frac{\bv^{0\intercal}}{\sqrt{k}} \He_2 \!\left( \frac{\bW^0 \bx_\mu}{\sqrt d} \right) = \Tr \frac{\bW^{0\intercal} (\bv^0) \bW^0}{d\sqrt{k}} \bx_\mu \bx_\mu^\intercal - \frac{\bv^{0\intercal} \bm{1}_k}{\sqrt{k}}\approx \frac{1}{\sqrt{k}d} \Tr[( \bx_\mu \bx_\mu^\intercal - \I_d)\bW^{0\intercal} (\bv^0) \bW^0 ],
\end{align}
where $\approx$ is exploiting the concentration $\Tr \bW^{0\intercal} (\bv^0) \bW^0 /(d\sqrt{k}) \to \bv^{0\intercal} \bm{1}_k/\sqrt{k}$,
and the Gaussian equivalence property that $\bM_\mu:=(\bx_\mu \bx_\mu^\intercal - \I_d)/\sqrt{d}$ behaves like a GOE sensing matrix, i.e., a symmetric matrix whose upper triangular part has i.i.d. entries from $\mathcal{N}(0,(1+\delta_{ij})/d)$ \cite{maillard2024bayes}, the model can be seen as a GLM with signal $\bar\bS^0_2 := \bW^{0\intercal} (\bv^0) \bW^0/\sqrt{kd}$:
\begin{align}\label{eq:Maillard_starting_point}
    y^{\rm GLM}_\mu=\frac{\mu_2}{2}\Tr [\bM_\mu \bar\bS^0_2] +\sqrt{\Delta+g(1)}\,z_\mu.
\end{align}
Starting from this equation, the arguments of App.~\ref{app:replicas} and \citet{maillard2024bayes}, based on known results on the GLM \cite{barbier2019glm} and matrix denoising \cite{barbier2022statistical,maillard2022perturbative,matrix_inference_Barbier}, allow us to obtain the free entropy of the GLM. The result is consistent with the one obtained in App.~\ref{app:replicas} with the replica method. 

We have thus identified an effective GLM representation of the learning model, which is valid in the universal phase. On the contrary, in the specialisation phase we cannot consider the $\dots$ terms in Eq.~\eqref{eq:hexpan} as noise uncorrelated with the first ones, as the model is aligning with the actual teacher's weights, such that it learns all the successive terms at once.

\begin{algorithm}[t]
   \caption{GAMP-RIE for training shallow neural networks with arbitrary activation}
   \label{alg:gamp}
\begin{algorithmic}\label{algo}
    \STATE {\bfseries Input:} Fresh data point $\bx_{\text{test}}$ with unknown associated response $y_{\text{test}}$, dataset $\mathcal{D}=\{(\bx_\mu, y_\mu)\}_{\mu=1}^n$. 
    \STATE {\bfseries Output:} Estimator $\hat y_{\text{test}}$ of $y_{\text{test}}$.
    \STATE Estimate $y^{(0)} := \mu_0 \bv^{0\intercal} \bm{1}/\sqrt{k}$ as 
    \begin{equation*}
        \hat y^{(0)} = \frac{1}{n}\sum_{\mu} y_\mu ;
    \end{equation*}
    \STATE  Estimate $ \langle \bv^\intercal \bW \rangle$ using Monte Carlo sampling;
    \STATE Estimate the $\mu_1$ term in the Hermite expansion (\ref{eq:hexpan}) as
    \begin{align*}
        \hat y_\mu^{(1)} &= \mu_1 \frac{ \langle \bv^\intercal \bW \rangle \bx_\mu }{\sqrt{kd}} ;
    \end{align*}
    \STATE  Compute 
    \begin{align*}
        \tilde y_\mu &=  \frac{y_\mu - \hat y_\mu^{(0)} - \hat y_\mu^{(1)}}{\mu_2/2} ; \qquad \tilde \Delta = \frac{\Delta + g(1)}{\mu_2^2/4} ;
    \end{align*}
    \STATE  Input $\{(\bx_\mu, \tilde y_\mu)\}_{\mu=1}^n$ and $\tilde \Delta$ into Algorithm 1 in \citet{maillard2024bayes} to estimate $\langle \bW^\intercal (\bv) \bW \rangle$;
    \STATE  Output 
    \begin{align} \label{eq:output_GAMP_RIE}
        \hat y_{\text{test}} = \hat y^{(0)} + \mu_1 \frac{ \langle \bv^\intercal \bW \rangle \bx_{\text{test}}}{\sqrt{kd}} + \frac{\mu_2}{2} \frac{1}{d\sqrt k} \Tr[ (\bx_{\text{test}} \bx_{\text{test}}^\intercal - \I ) \langle \bW^\intercal (\bv) \bW \rangle ].
    \end{align}
\end{algorithmic}
\end{algorithm}

We now assume that this 
mapping holds at the algorithmic level, namely, that we can process the data algorithmically as if they were coming from the identified GLM, and thus try to infer the signal $\bar\bS_2^0 = \bW^{0\intercal} (\bv^0) \bW^0/\sqrt {kd}$ and construct a predictor from it. Based on this idea, we propose the Algorithm~\ref{alg:gamp} below that can indeed reach the performance predicted by the universal branch of our theory. 

\begin{remark}\label{rem:linear}
In the linear data regime, where $n/d$ converges to a fixed constant $\alpha_1$, only the first term in (\ref{eq:hexpan}) can be learned while the rest behaves like noise. By the same argument as above, the model is equivalent to 
\begin{align}
     y_\mu = \mu_1 \frac{\bv^{0\intercal} \bW^0 \bx_\mu}{\sqrt{kd}} + \sqrt{\Delta + \nu - \mu_0^2 - \mu_1^2} \, z_\mu,
\end{align}
where $\nu = \EE_{z\sim\calN(0,1)}[\sigma^2(z)]$.
This is again a GLM with signal $\bS_1^0 = \bW^{0\intercal} \bv^0/\sqrt k$ and Gaussian sensing vectors $\bx_{\mu}$. Define $q_1$ as the limit of $\bS_1^{a\intercal}\bS_1^b /d$ where $\bS_1^a, \bS_1^b$ are drawn independently from the posterior. With $k \rightarrow \infty$, the signal converges in law to a standard Gaussian vector. Using known results on GLMs with Gaussian signal, we obtain the following saddle point equations for $q_1$:
\begin{align*}
q_1 & = \frac{\hat q_1}{\hat q_1 + 1}, \qquad
\hat q_1 = \frac{\alpha_1}{1 + \Delta_1 - q_1},\quad \text{where} \quad \Delta_1 = \frac{\Delta + \nu - \mu_0^2 - \mu_1^2}{\mu_1^2}.
\end{align*}
In the quadratic data regime, as $\alpha_1=n/d$ goes to infinity, the overlap $q_1$ converges to $1$ and the first term in (\ref{eq:hexpan}) is learned with vanishing error.
\end{remark}

\begin{remark}
The same argument can be easily generalised for general $P_{\text{out}}$, leading to the following equivalent GLM in the universal phase of quadratic data regime:
\begin{align}
    y_\mu^{\rm GLM} \sim \tilde P_{\text{out}}(\cdot\mid \Tr [\bM_\mu \bar\bS^0_2] ), \quad \text{where} \quad  \tilde P_{\text{out}}(y|x) := \mathbb E_{z \sim \mathcal N(0,1)} P_{\text{out}}\Big(y \mid  \frac{\mu_2}{2} x + z\sqrt{g(1)} \Big),
\end{align}
and $\bM_\mu$ are independent GOE sensing matrices.
\end{remark}

\begin{remark}
One can show that the system of equations $({\rm S_{uni}})$ in the main or in \eqref{eq:Suni} can be mapped onto the fixed point of the state evolution equations (92), (94) of the GAMP-RIE in \citet{maillard2024bayes} up to changes of variables. This confirms that when \eqref{eq:Suni} has a unique solution, which is the case in all our tests, the GAMP-RIE asymptotically matches our universal solution. The deviations of the GAMP-RIE points at small $\alpha$ in Fig.~\ref{fig:gen_errors_univ_spec}, bottom part, are thus due to finite size effects. Assuming the validity of the aforementioned effective GLM, a potential improvement for discrete weights could come from a generalisation of GAMP which, in the denoising step, would correctly exploit the discrete prior over inner weights rather than using the RIE (which is prior independent). However, the results of \citet{barbier2024phase} suggest that optimally denoising matrices with discrete entries is hard, and the RIE is the best efficient procedure to do so. Consequently, we tend to believe that improving GAMP-RIE in the case of discrete weights is out of reach without strong side information about the teacher, or exploiting non-polynomial-time algorithms (see Appendix~\ref{sec:hardness}).
\end{remark}

\section{Gaussian prior over the inner weights\label{sec:gaussian_prior}}

\begin{figure}[t]
    \centering
    \includegraphics[height=5.5cm]{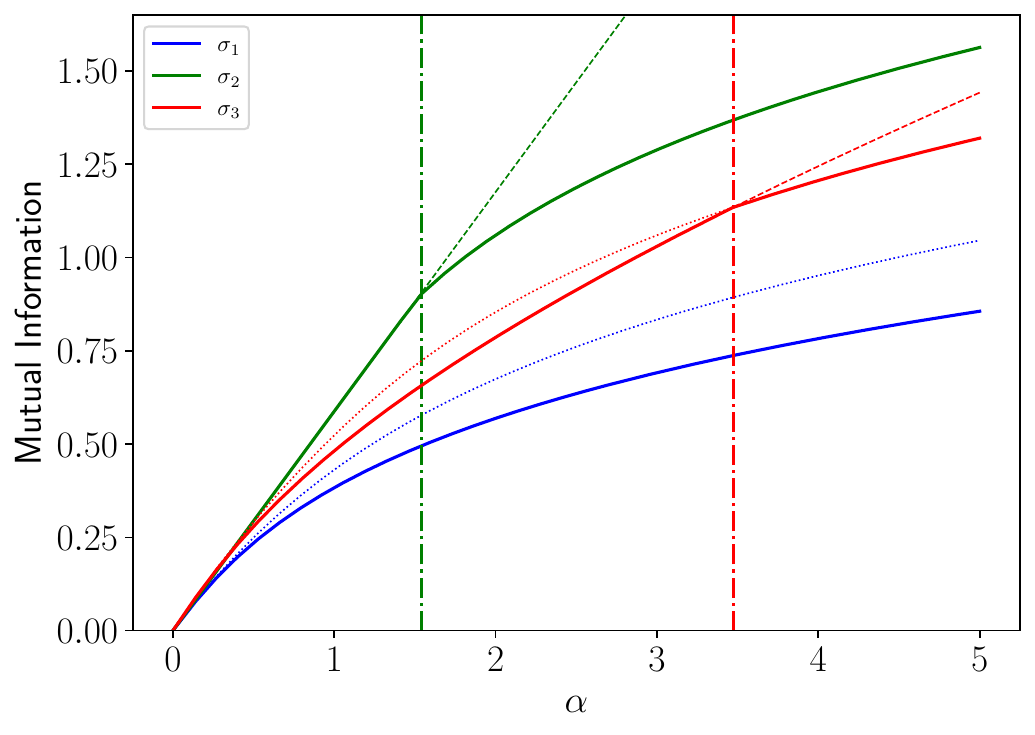}
    \includegraphics[height=5.5cm]{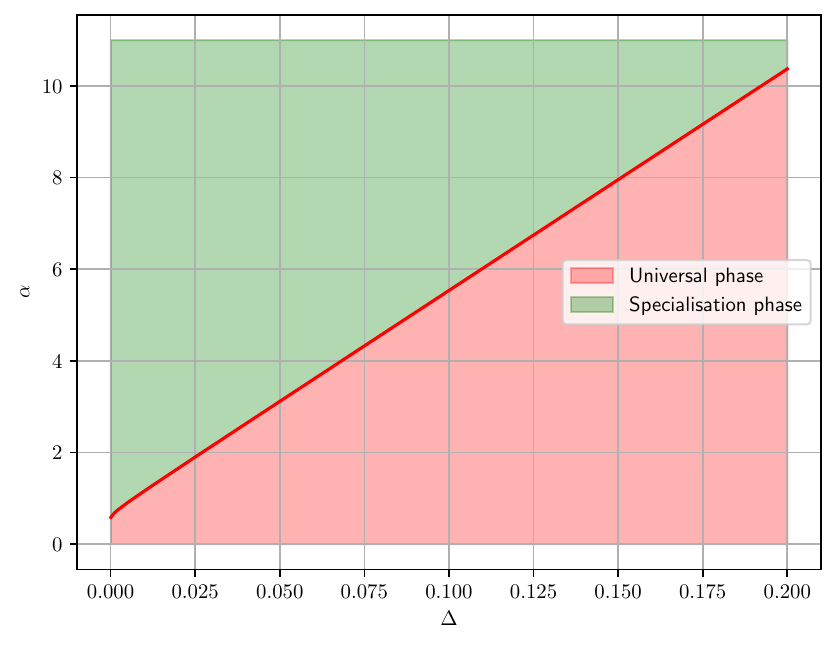}
    \caption{\textbf{Left}: Mutual information for Gaussian prior over the inner weights of the universal (dashed) and specialisation (dotted) solutions, for the activation functions in Fig.~\ref{fig:gen_errors_univ_spec} (top panel), $\gamma=0.5$ and Gaussian label noise with variance $\Delta=1.25$. $\alpha_{\rm sp}$ is the point where they cross (dash-dot line). \textbf{Right}: Phase diagram in the $(\Delta,\alpha)$ plane for Gaussian prior over the inner weights, ReLU activation and parameter $\gamma = 0.5$, with fixed read-outs $\bv =\bv^0 = \mathbf{1}$. The red line is the curve $\alpha_{\rm sp}$, defined in Eq.~\eqref{eq:alpha_c}.}
    \label{fig:gaussian}
\end{figure}

In most of this paper we focused on the case of inner weights with Rademacher prior, for which we showed the existence of a stable specialisation phase arising for $\alpha> \alpha_{\rm sp}$ and generic activation function. Another case of major practical interest is the one of inner weights with Gaussian prior, which we reported in Fig.~\ref{fig:gen_error_gauss} and we discuss more in this section. The theory we devised in the text is general in the choice of the prior, so that to obtain predictions for this case we only need to decline the function $\psi_{P_W}$ defined in Section~\ref{sec:result} (see also Eq.~\eqref{eq:psi_p_w}). For binary prior, the straightforward calculation gives
\begin{equation}
    \psi_{P_W = \Rad}(\hat q_W) 
    = -\frac{\Hat{q}_W}{2} +  \mathbb{E}_{\xi_w\sim\mathcal{N}(0,1)} \ln \cosh( \sqrt{\Hat{q}_W} \xi_w +\Hat{q}_W ) ,
\end{equation}
whereas, for Gaussian prior, we get
\begin{equation}
    \psi_{P_W = \calN(0,1)}(\hat q_W) 
    = \frac{\hat{q}_W}{2}-\frac{1}{2} \ln(1+\hat{q}_W).
\end{equation}
The free entropies of the universal (independent from $P_W$) and specialisation (dependent on $P_W$) solutions can be evaluated in both cases, as explained in the main text and in the previous appendices. For the polynomial activation functions considered in Fig.~\ref{fig:gen_errors_univ_spec}, the mutual informations obtained with binary prior over the inner weights are reported in Fig.~\ref{fig:MIcrossing}, while we report here the analogous curves obtained with Gaussian prior (Fig.~\ref{fig:gaussian}, left panel).

The two priors, while both showing a non-trivial $\alpha_{\rm sp}$ where specialisation arises, exhibit rather different behaviours in the related MIs and overlaps as functions of $\alpha$. In fact, in the case of Rademacher prior, the MI must saturate to the entropy of the prior itself (see the argument in \citet{barbier2024phase}), which is also reflected in the quick saturation of the overlap $q_W$ to 1. Specifically, declining the saddle point equations \eqref{FP_equations_gaussian_ch} to Rademacher prior it is easy to see that 
\begin{align}
    q_W=\EE\tanh(\sqrt{\hat q_W}\xi_w+\hat q_W)
\end{align}with $\hat q_W$ containing a global factor $\alpha$, and increasing with it. The presence of the hyperbolic tangent is what yields the characteristic exponential saturation to $1$ of $q_W$ when $\alpha$ grows, and thus the exponential decrease of the generalisation error in the specialisation phase, as can be seen from Fig.~\ref{fig:gen_errors_univ_spec}, top panel, inset.
For Gaussian prior instead,
\begin{align}
    q_W=\frac{\hat q_W}{1+\hat q_W}
\end{align}and when $\alpha$ approaches $+\infty$, since the dependency of $\hat q_W$ on it is always algebraic, one expects also $q_W$ to converge to $1$ with algebraic speed. This is also reflected in the MI, that for Gaussian prior is not supposed to saturate to a given value contrary to the discrete prior case.

A novelty with respect to the problem of matrix denoising is that the specialisation phase, akin to the factorisation phase pinpointed in \citet{barbier2024phase}, may occur also for Gaussian prior (in agreement with our numerical experiments), as the curves predicted by the universal and specialisation theories can cross. We observe that this happens when the activation function possesses at least a Hermite coefficient beyond the 2nd in its expansion, see the blue curve for $\sigma_1(x) = \He_2(x)/\sqrt{2}$ in Fig.~\ref{fig:gaussian}, left: the MIs of the two solutions never cross in this case (similarly to what happens in matrix denoising with Gaussian prior \cite{barbier2024phase}). Those terms are indeed the ones responsible for better generalisation: since the function $g$ contains only Hermite coefficients from the third on, having a non-vanishing overlap $q_W$ is the only chance to decrease the contribution $g(1)-g(q_W)$ in $\varepsilon^{\rm opt}$. 

This is in particular true for ReLU activation function, for which our theory predicts a phase digram in the $(\Delta,\alpha)$ plane reported in Fig.~\ref{fig:gaussian}, right panel.

\section{Algorithmic complexity of finding the specialisation solution\label{sec:hardness}}

\begin{figure}[p]
\begin{center}
\centerline{
\includegraphics[width=.49\linewidth,trim={0 0 0 0},clip]{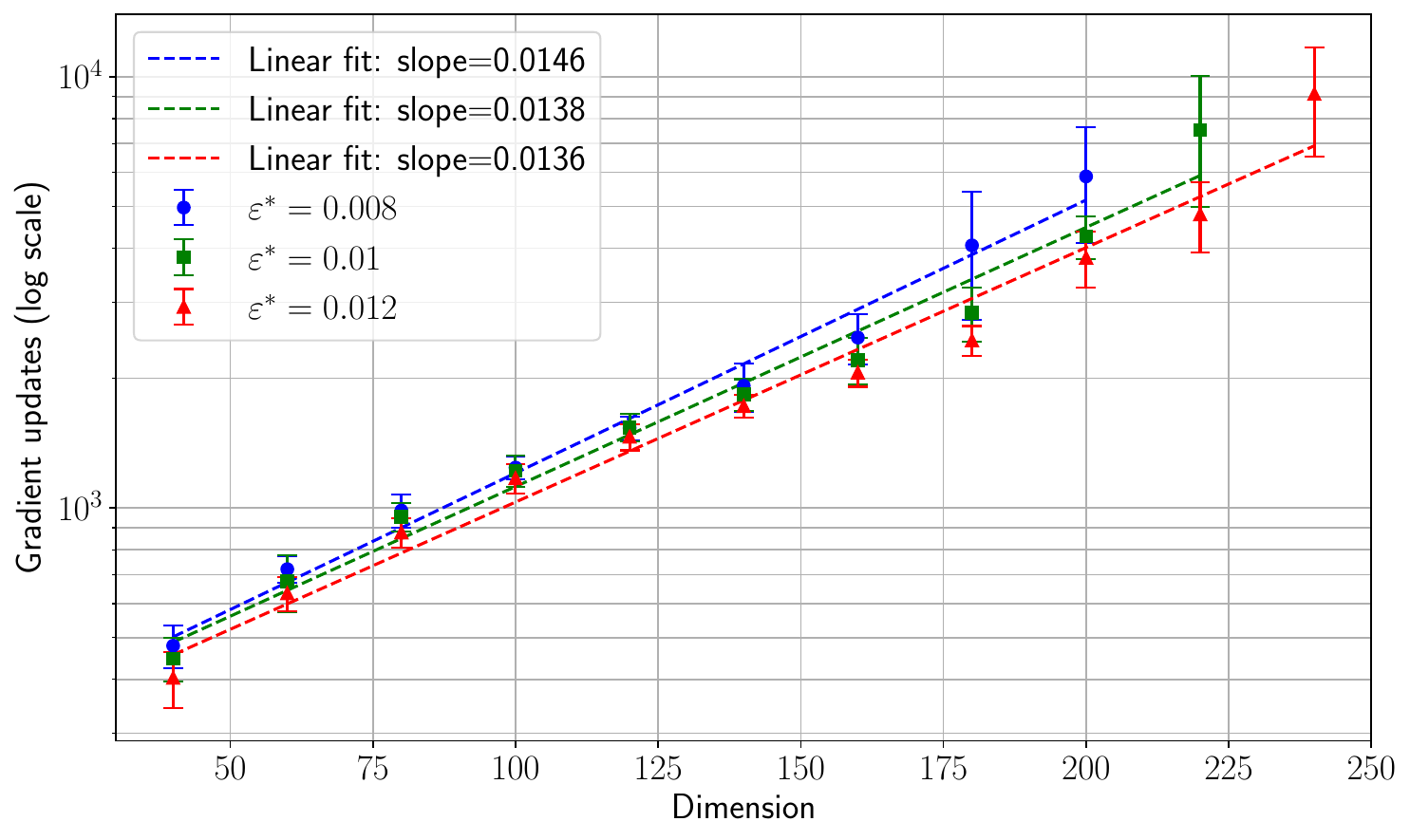}
\includegraphics[width=.49\linewidth,trim={0 0 0 0},clip]{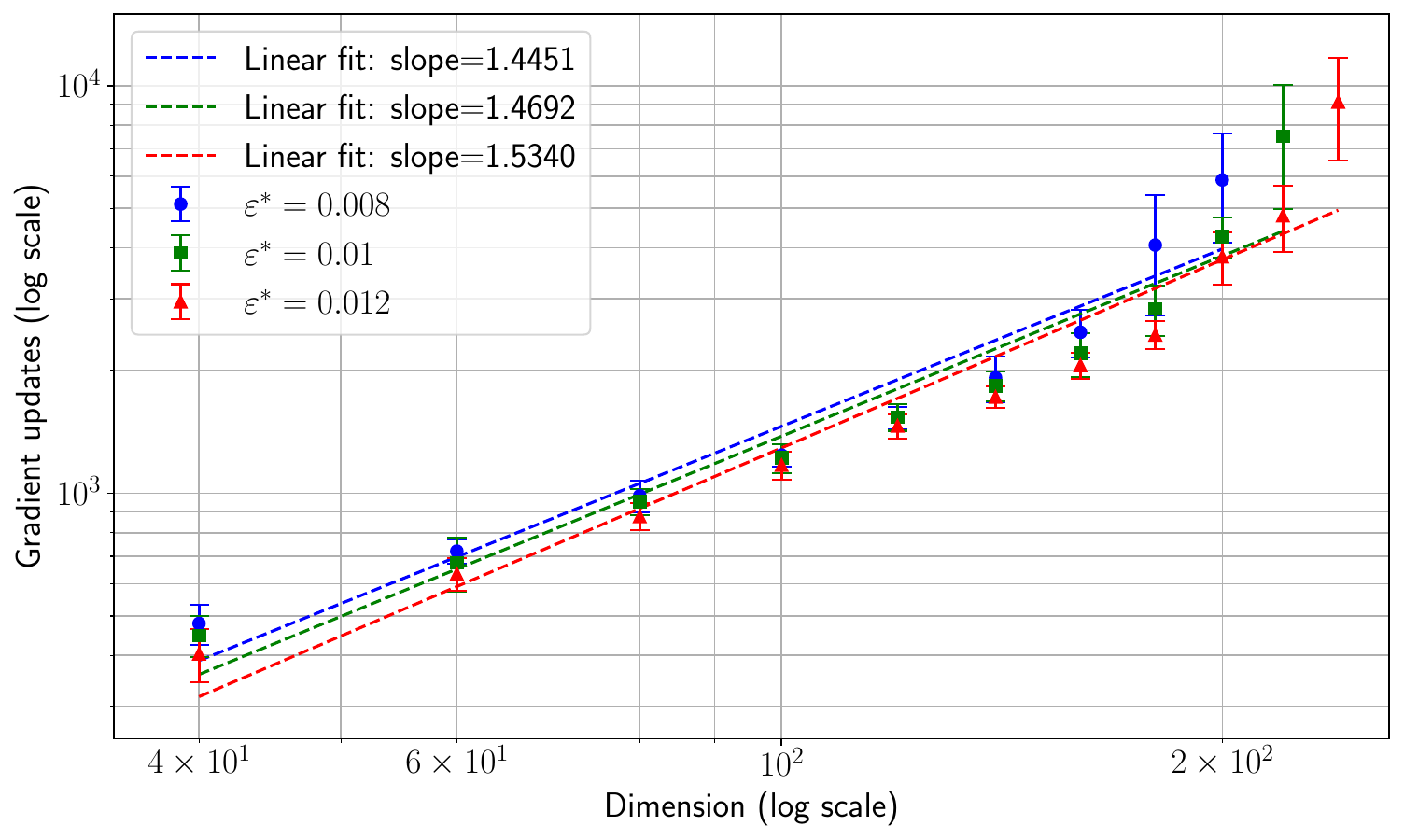}
}
\vspace{1pt} 
\centerline{
\includegraphics[width=.49\linewidth,trim={0 0 0 0},clip]{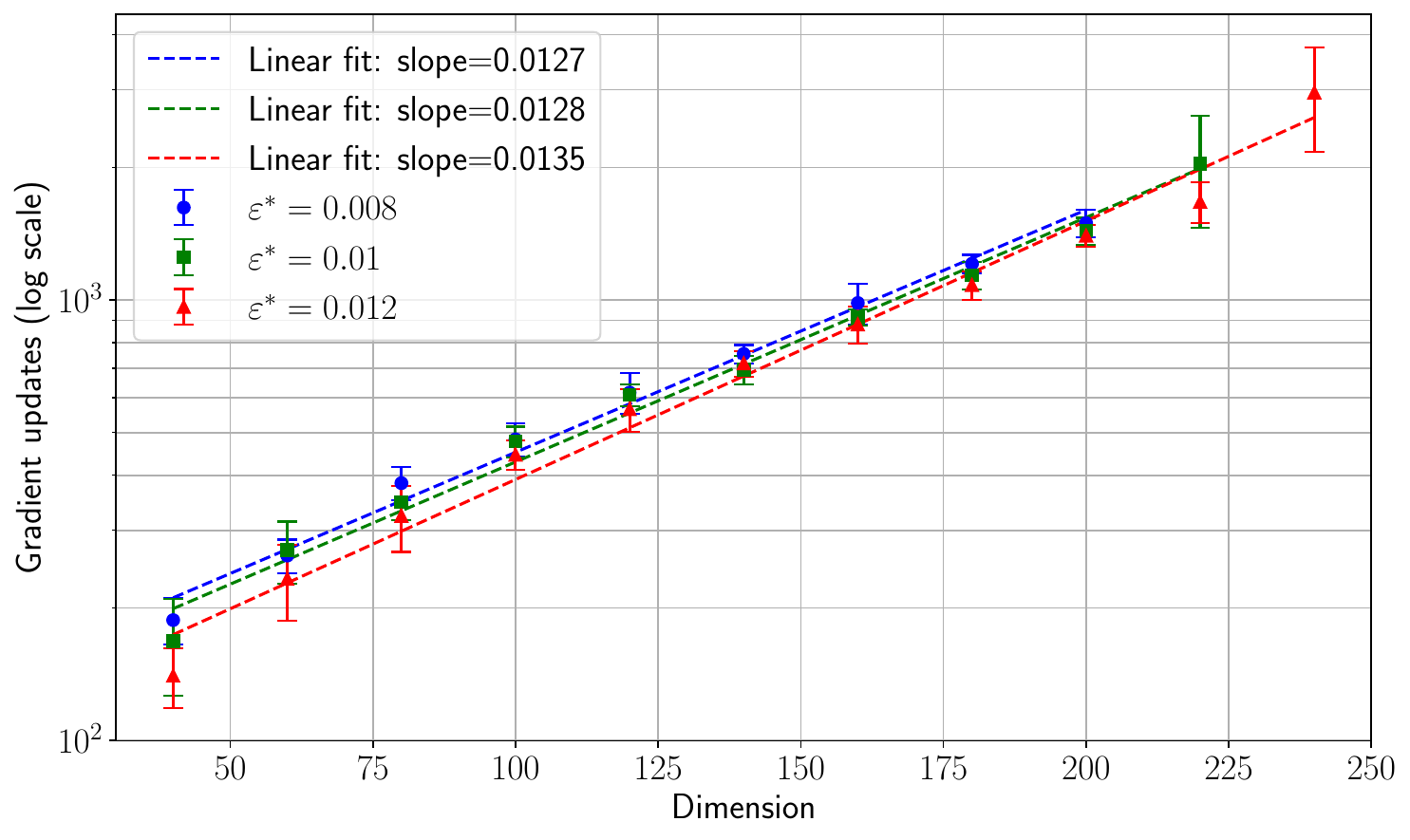}
\includegraphics[width=.49\linewidth,trim={0 0 0 0},clip]{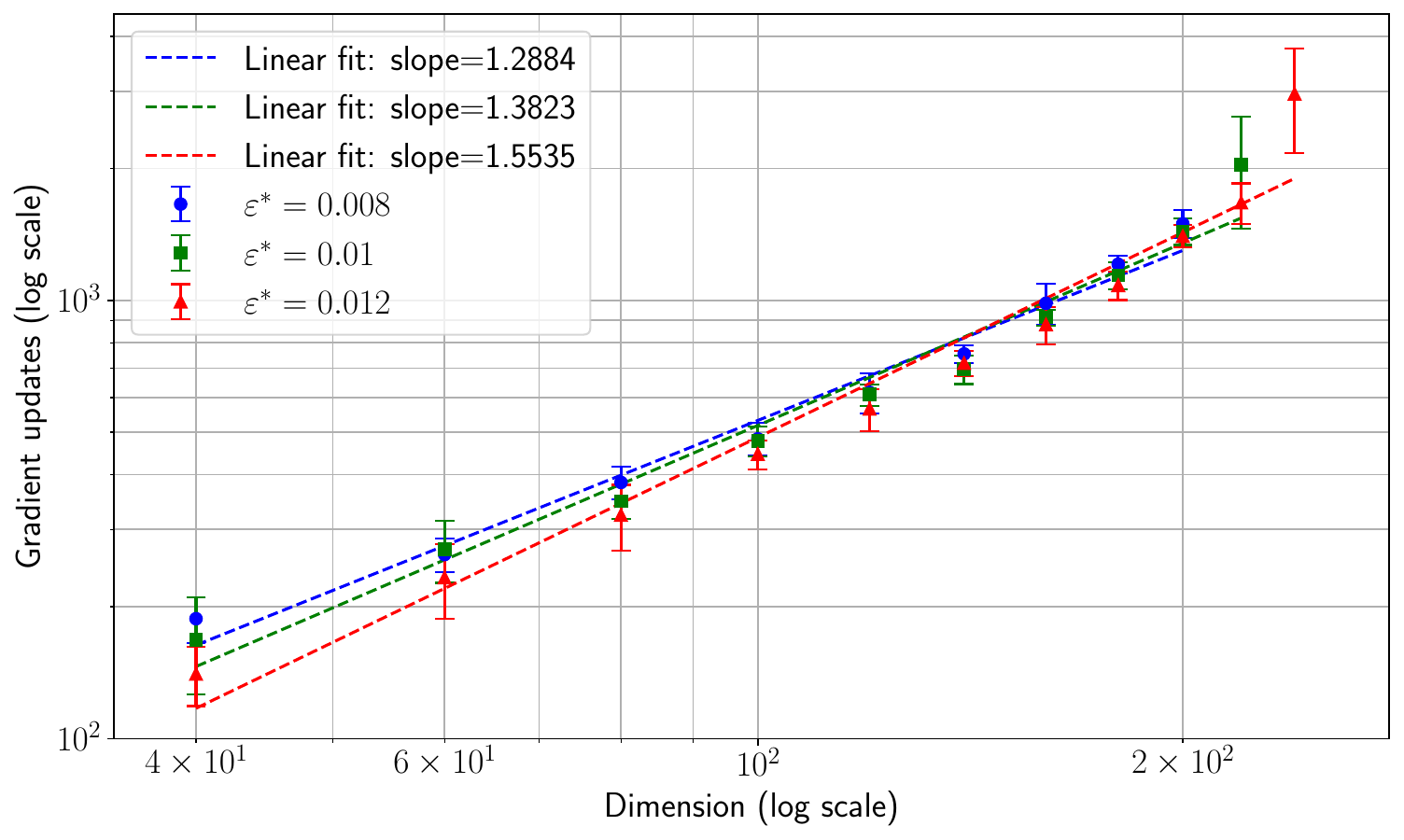}
}
\vspace{1pt} 
\centerline{
\includegraphics[width=.49\linewidth,trim={0 0 0 0},clip]{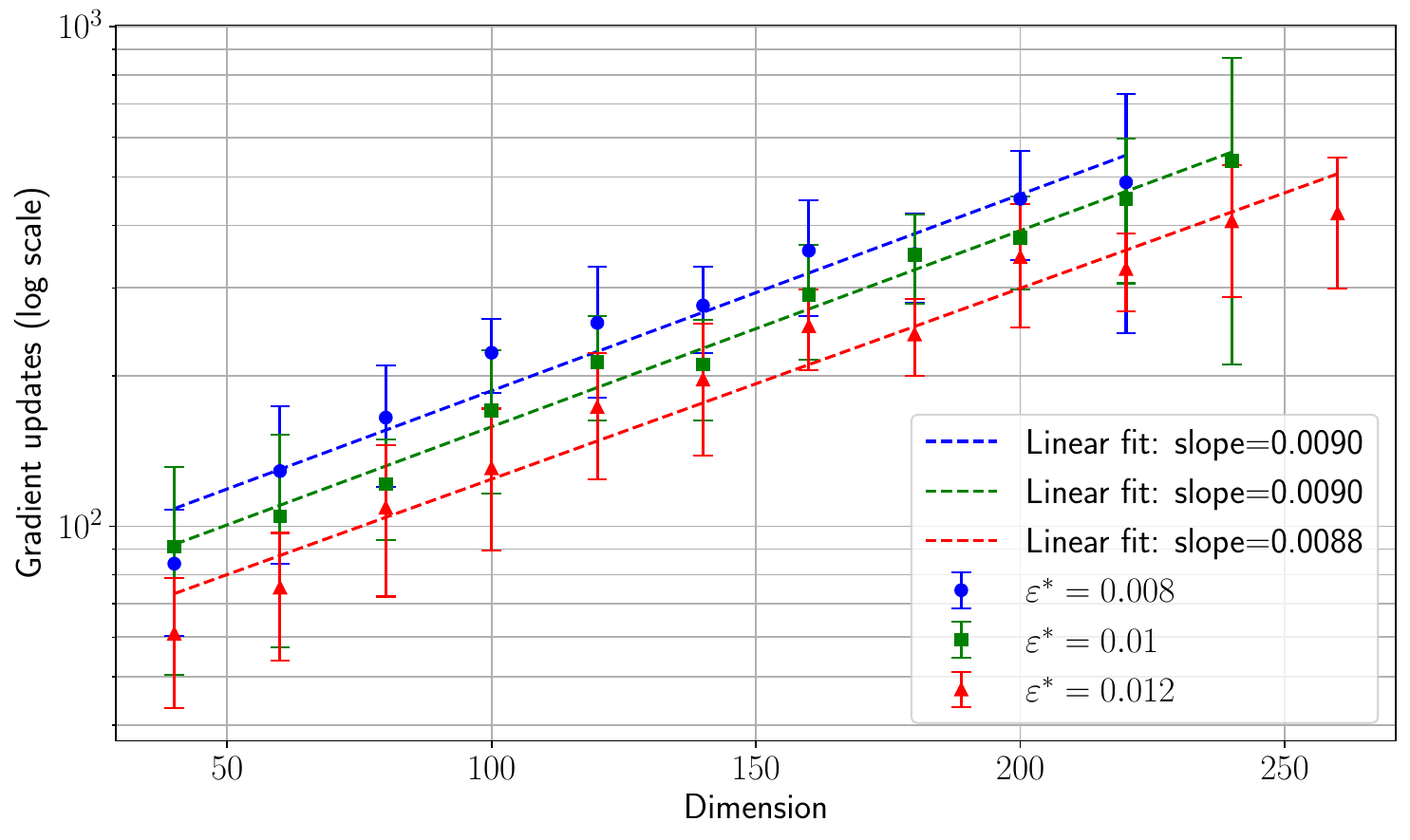}
\includegraphics[width=.49\linewidth,trim={0 0 0 0},clip]{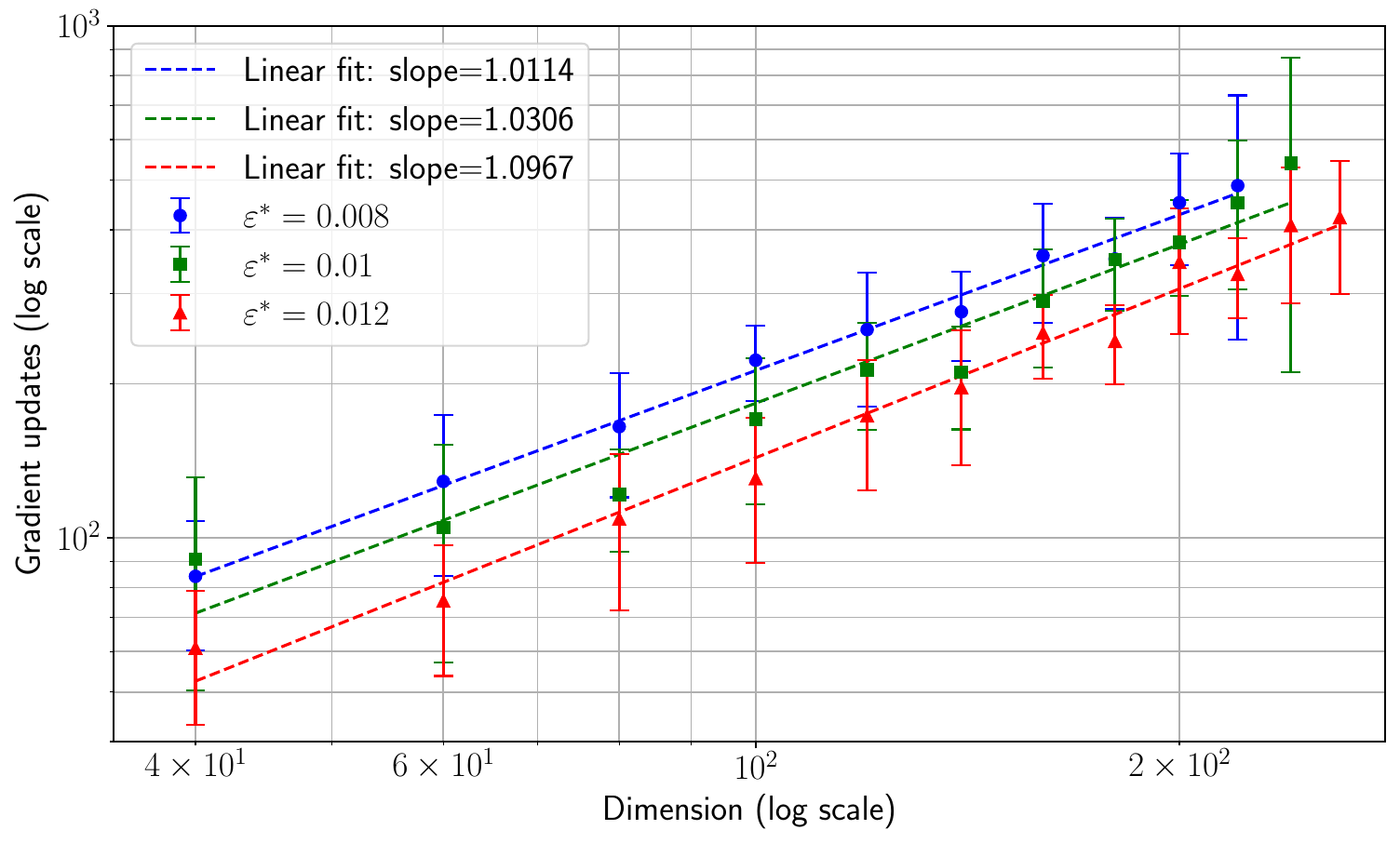}
}
\vspace{-10pt}
    \caption{
    Semilog (\textbf{Left}) and log-log (\textbf{Right}) plots of the number of gradient updates needed to achieve a test loss below the threshold $\varepsilon^*< \varepsilon^{\rm uni}$. Student network trained with ADAM with optimised batch size for each point. The dataset was generated from a teacher network with ReLU activation and parameters $\Delta = 10^{-4}$, $\gamma=0.5$ and $\alpha=5.0$ for which $\varepsilon^{\rm opt}-\Delta=1.115 \times 10^{-5}$.
     Points are obtained averaging over 10 teacher instances with error bars representing the standard deviation.
    Each row corresponds to a different distribution of the read-outs, kept fixed during training. \textbf{Top}: constant read-outs, for which the error of the universal branch is $\varepsilon^{\rm uni}-\Delta=1.217\times 10^{-2}$. \textbf{Center}: Rademacher read-outs, for which $\varepsilon^{\rm uni}-\Delta=1.218\times 10^{-2}$.
    \textbf{Bottom}: Gaussian read-outs, for which $\varepsilon^{\rm uni}-\Delta=1.210\times 10^{-2}$. The quality of the fits can be read from Table~\ref{tab:adam}.
}
    \label{fig:hardness_adam}
\end{center}
\vskip -0.3in
\end{figure}

\begin{table}[p]
    \centering
    \begin{tabular}{lc|c|c|c|c|c|c|}
            &        & \multicolumn{3}{c|}{$\chi^2$ exponential fit} & \multicolumn{3}{c|}{$\chi^2$ power law fit}\\
        \hline
        Read-outs    &$\epsilon^*=$& $0.008$ & $0.010$ & $0.012$ & $0.008$ & $0.010$ & $0.012$ \\
        \hline
         Constant&&$\bm{5.57}$ & $\bm{9.00}$ & $\bm{21.1}$ &$32.3$&$26.5$&$61.1$ \\
         Rademacher&& $\bm{4.51}$ & $\bm{6.84}$ & $\bm{12.7}$ &$12.0$&$17.4$ &$16.0$\\
         Uniform $[-\sqrt{3},\sqrt{3}]$&& $\bm{5.08}$ & $\bm{1.44}$ & ${4.21}$ &$8.26$ &$8.57$ &$\bm{3.82}$\\
         Gaussian&&$2.66$&$\bm{0.76}$&$3.02$ & $\bm{0.55}$&$2.31$ &$\bm{1.36}$\\
    \end{tabular}
    \caption{$\chi^2$ test for exponential and power-law fits for the time needed by ADAM to reach the thresholds $\varepsilon^*$, for various priors on the read-outs. Fits are displayed in \figurename~\ref{fig:hardness_adam}. Smaller values of $\chi^2$ (in bold, for given threshold and read-outs) indicate a better compatibility with the hypothesis.}
    \label{tab:adam}
\end{table}

\begin{figure}[t]
    \centering
    \includegraphics[width=0.48\linewidth]{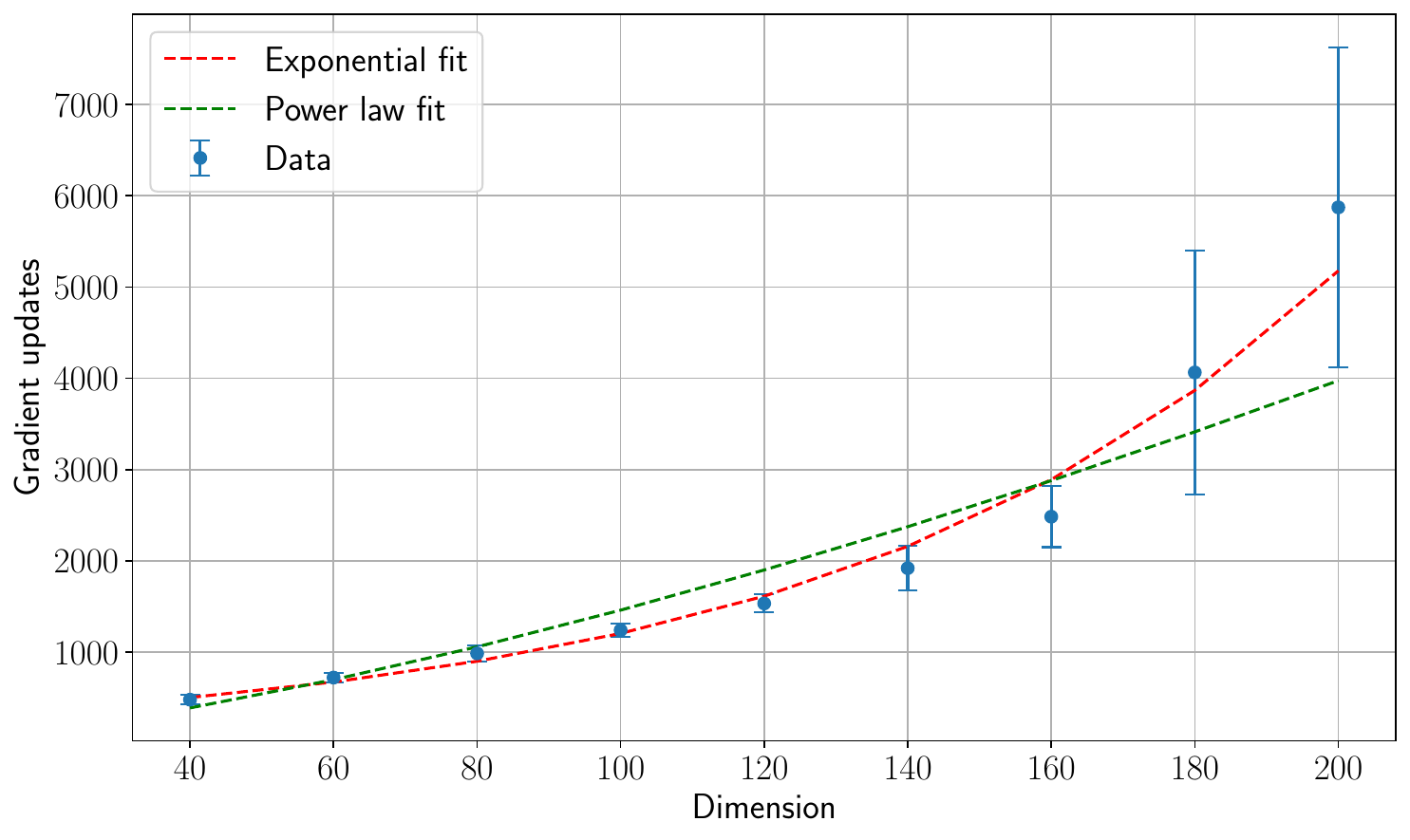}
    \includegraphics[width=0.48\linewidth]{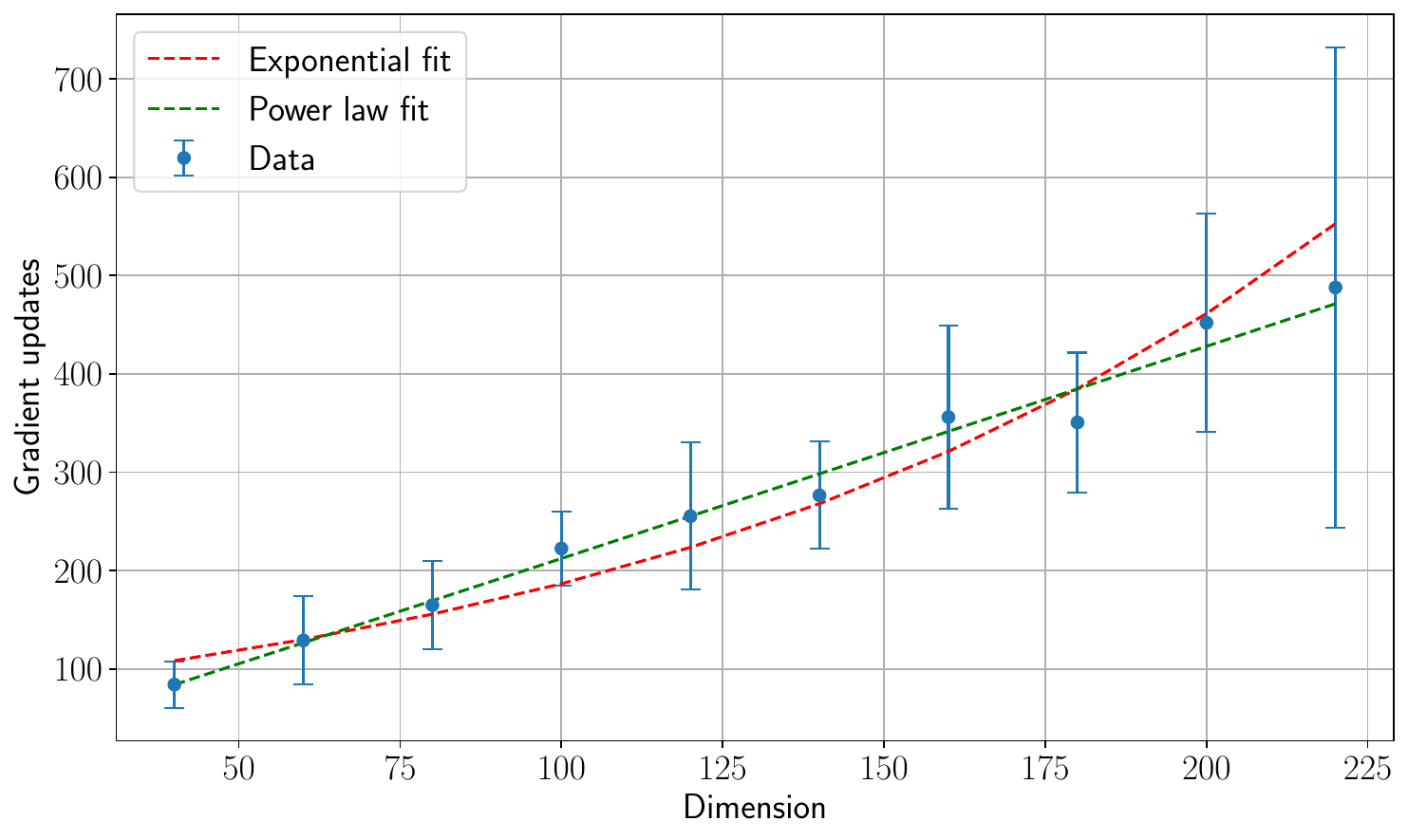}
    \caption{Same as in Fig.~\ref{fig:hardness_adam}, but in linear scale for better visualisation, for constant read-outs (\textbf{Left}) and Gaussian read-outs (\textbf{Right}), with threshold $\varepsilon^*=0.008$.}
    \label{fig:hardness_adam_linear}
\end{figure}

\begin{figure}[t]
    \centering
    \includegraphics[width=0.34\linewidth,trim={0 0 0 0},clip]{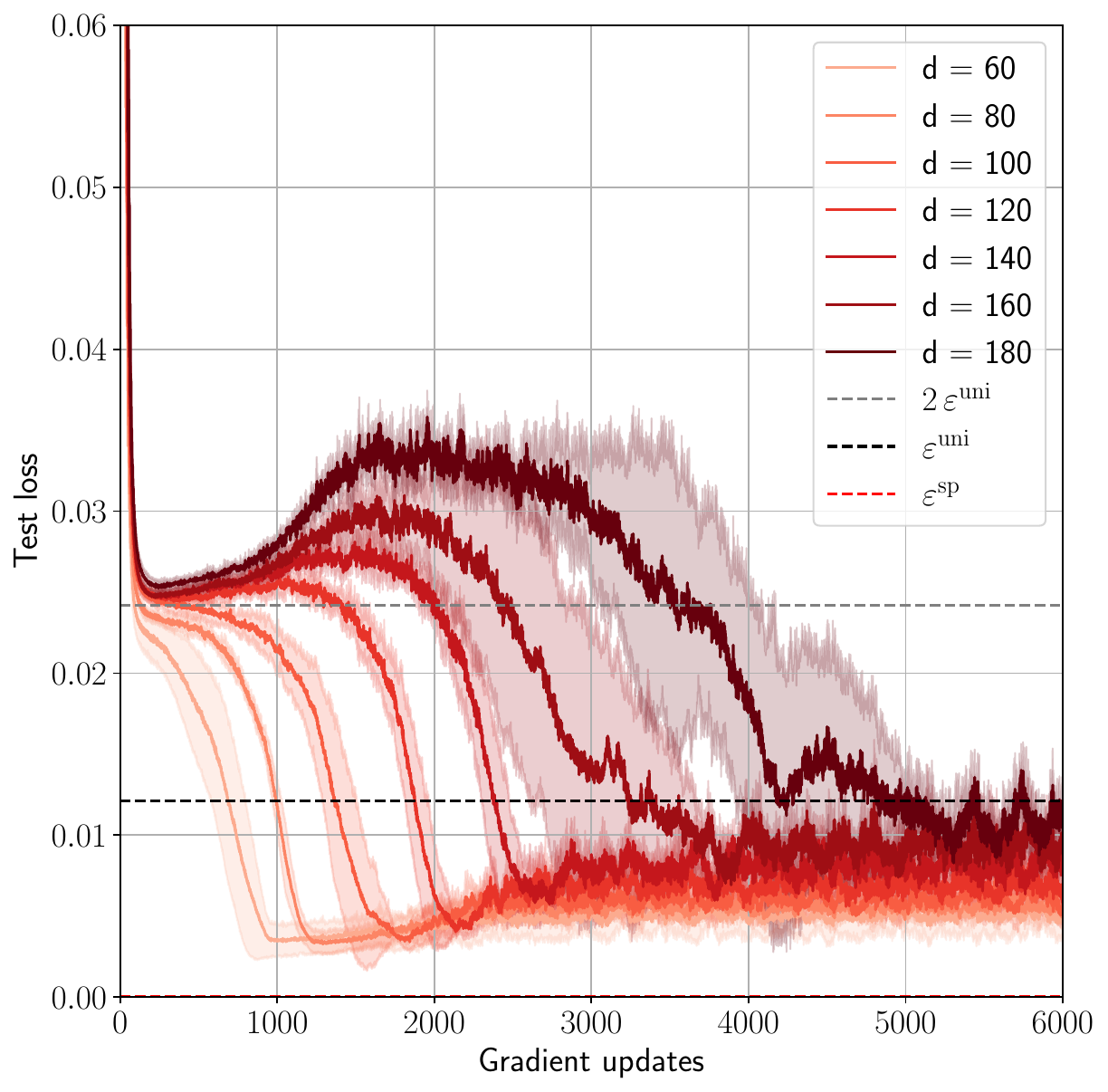}
    \includegraphics[width=0.325\linewidth,trim={0.9cm 0 0 0},clip]{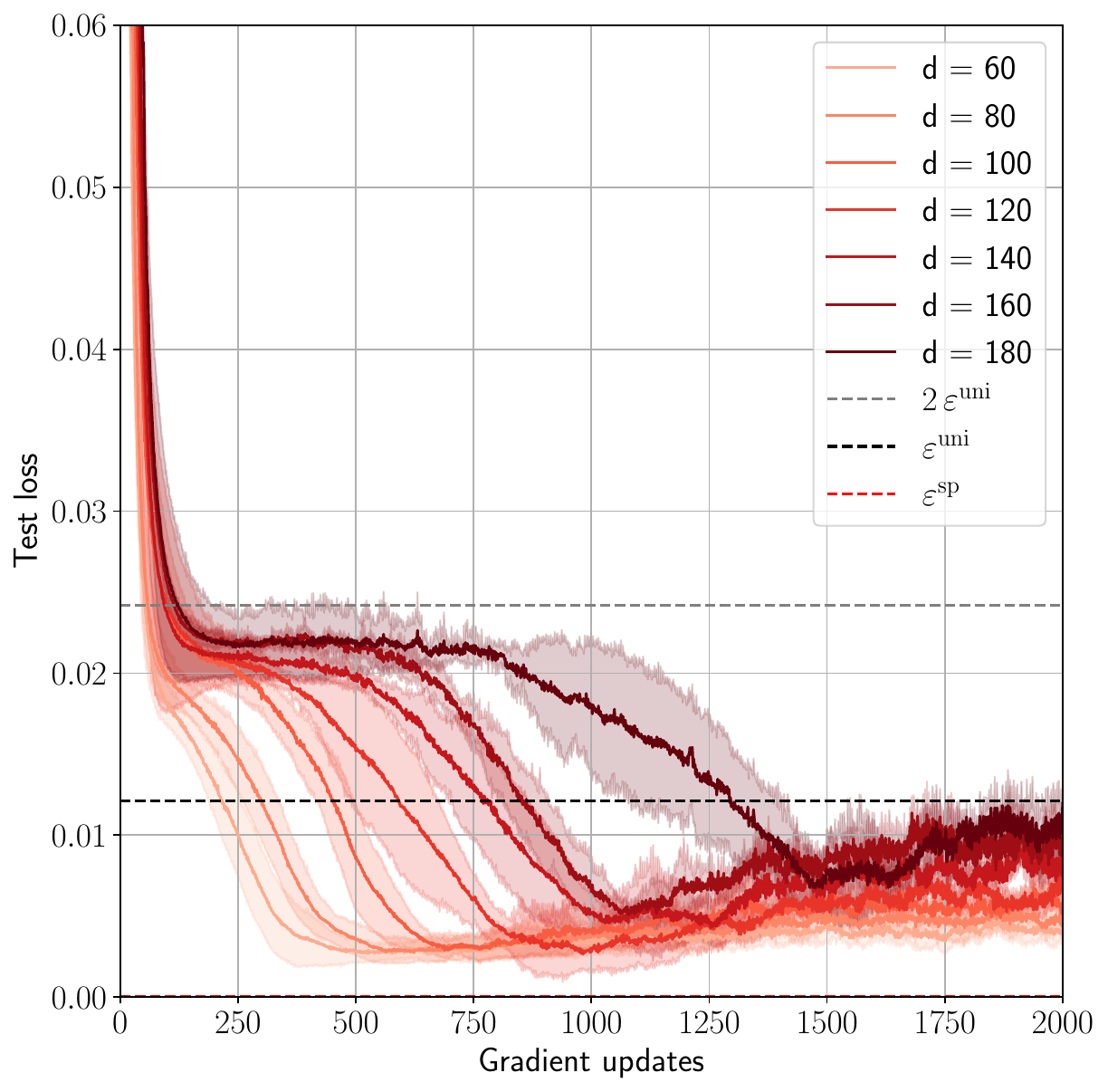}
    \includegraphics[width=0.32\linewidth,trim={0.9cm 0 0 0},clip]{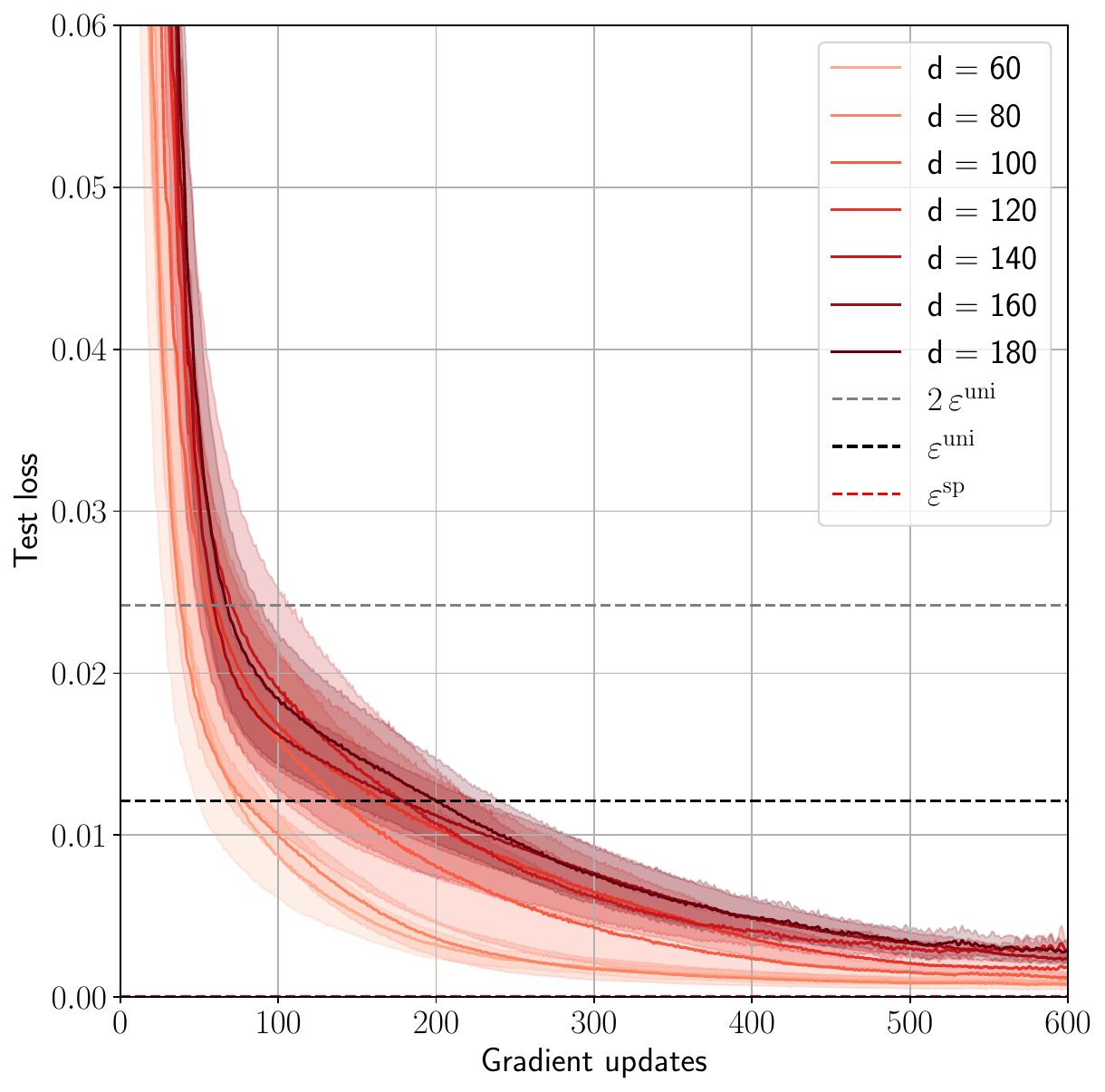}
    \caption{Trajectories of the generalisation error of neural networks trained with ADAM at fixed batch size $B=\lfloor n/4\rfloor$, for ReLU activation with parameters $\Delta = 10^{-4}$, $\gamma=0.5$ and $\alpha=5.0 > \alpha_{\rm sp}$.
    \textbf{Left}: Constant read-outs.
    \textbf{Centre}: Rademacher read-outs.
    \textbf{Right}: Gaussian read-outs. Read-outs are kept fixed (and equal to the teacher's) in all cases during training. Points on the solid lines are obtained by averaging over 5 teacher instances, and shaded regions around them correspond to one standard deviation.
    }
    \label{fig:hardness_adam_runs}
\end{figure}

We now provide empirical evidence concerning the computational complexity to find the specialisation solution we discussed in the main. We test two algorithms that can find it in testable computational time: ADAM with optimised batch size for every dimension tested (the learning rate is automatically tuned), and Hamiltonian Monte Carlo (HMC), both trying to infer a two-layer teacher network with Gaussian inner weights. 

\paragraph{ADAM}
We focus on ReLU activation, $\alpha=5.0>\alpha_{\rm sp}$ ($\alpha_{\rm sp}\approx 0.5$ in all the cases we report), $\gamma=0.5$ and Gaussian output channel with low label noise ($\Delta=10^{-4}$), so that the specialisation solution exhibits a very low generalisation error. We test the learned model at each gradient update measuring the generalisation error with a moving average of 10 steps to smoothen the curves. Fixing a threshold $ \varepsilon^{\rm opt} < \varepsilon^* < \varepsilon^{\rm uni}$, we define $t^*(d)$ the time (in gradient updates) needed for the algorithm to cross the threshold for the first time. We optimise over different batch sizes $B_p$ as follows: we define them as $B_p = \left\lfloor \frac{n}{2^p} \right\rfloor,\quad p = 2,3,\dots,\lfloor \log_2(n) \rfloor-1$. Then for each batch size, the student network is trained until the moving average of the test loss drops below $\varepsilon^*$ and thus outperforms the universal solution; we have checked that in such a scenario, the student ultimately gets close to the performance of the specialisation solution. The batch size that requires the least gradient updates is selected. We used the ADAM routine implemented in PyTorch. 

We test different distributions for the read-out weights (kept fixed to $\bv^0$ during training of the inner weights). We report all the values of $t^*(d)$ in Fig.~\ref{fig:hardness_adam} for various dimensions $d$ at fixed $(\alpha,\gamma)$, providing an exponential fit $t^*(d) = \exp(a d + b)$ (left panel) and a power-law fit $t^*(d) = a d^b $ (right panel). We report the $\chi^2$ test for the fits in Table~\ref{tab:adam}.
We observe that for constant and Rademacher read-outs, the exponential fit is more compatible with the experiments, while for Gaussian read-outs the comparison is inconclusive.

In Fig.~\ref{fig:hardness_adam_runs}, we report the test loss of ADAM as a function of the gradient updates used for training, for various dimensions and choice of the read-out distribution (as before, the read-outs are not learned but fixed to the teacher's). Here, we fix a batch size for simplicity. For both the cases of constant ($\bv=\bm{1}$) and Rademacher read-outs (left and centre panels), the model experiences plateaux in performance increasing with the system size, in accordance with the observation of exponential complexity we reported above. The plateaux happen at values of the test loss comparable with twice the value for the Bayes error predicted by the universal branch of our theory (remember the relationship between Gibbs and Bayes errors reported in App.~\ref{app:gen_err}). The curves are smoother for the case of Gaussian read-outs.

\paragraph{Hamiltonian Monte Carlo}
The experiment is performed for the polynomial activation $\sigma_3(x) = \He_2(x)/\sqrt 2 + \He_3(x)/6$ with parameters $\Delta = 0.1$, $\gamma=0.5$ and $\alpha=1.0 > \alpha_{\rm sp}$. Our HMC, implemented with Tensorflow Probability, consists of $4000$ iterations for constant read-outs, or $2000$ iterations for Rademacher and Gaussian read-outs. Each iteration is adaptive (with initial step size of $0.01$) and uses $10$ leapfrog steps. Instead of measuring the Gibbs error, whose relationship with $\varepsilon^{\rm opt}$ holds only at equilibrium (see the last remark in App.~\ref{app:gen_err}), we measured the teacher-student $q_2$-overlap which is meaningful at any HMC step and is informative about the learning. For a fixed threshold $q_2^*$ and dimension $d$, we measure $t^*(d)$ as the number of HMC iterations needed for the $q_2$-overlap between the uninformative HMC sample and the teacher weights $\bW^0$ to go beyond the threshold. This criterion is again enough to assess that the student outperforms the universal solution and will get close to the specialisation one at convergence.

\begin{figure}[pt]
\begin{center}
\centerline{
\includegraphics[width=.49\linewidth,trim={0 0 0 0},clip]{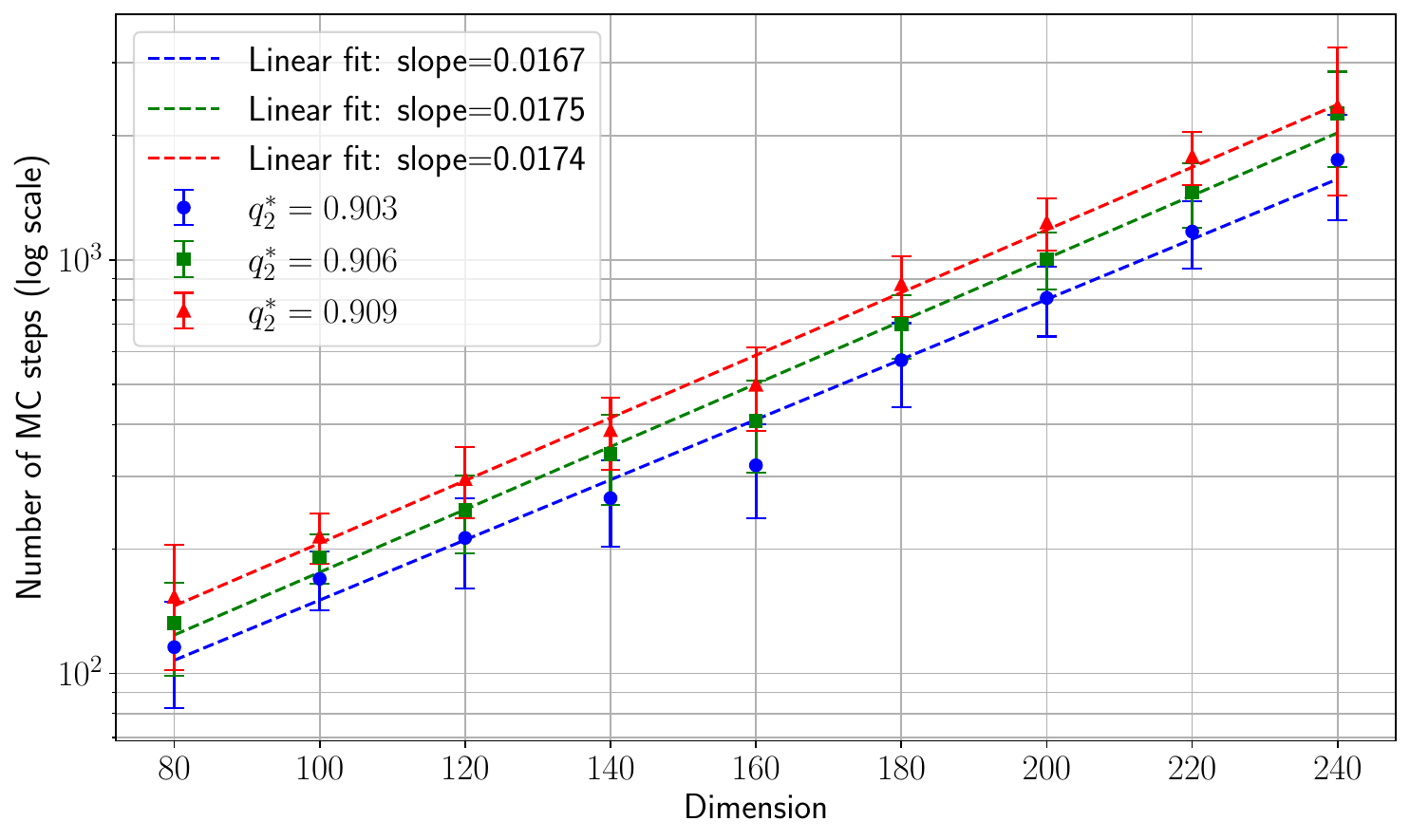}
\includegraphics[width=.49\linewidth,trim={0 0 0 0},clip]{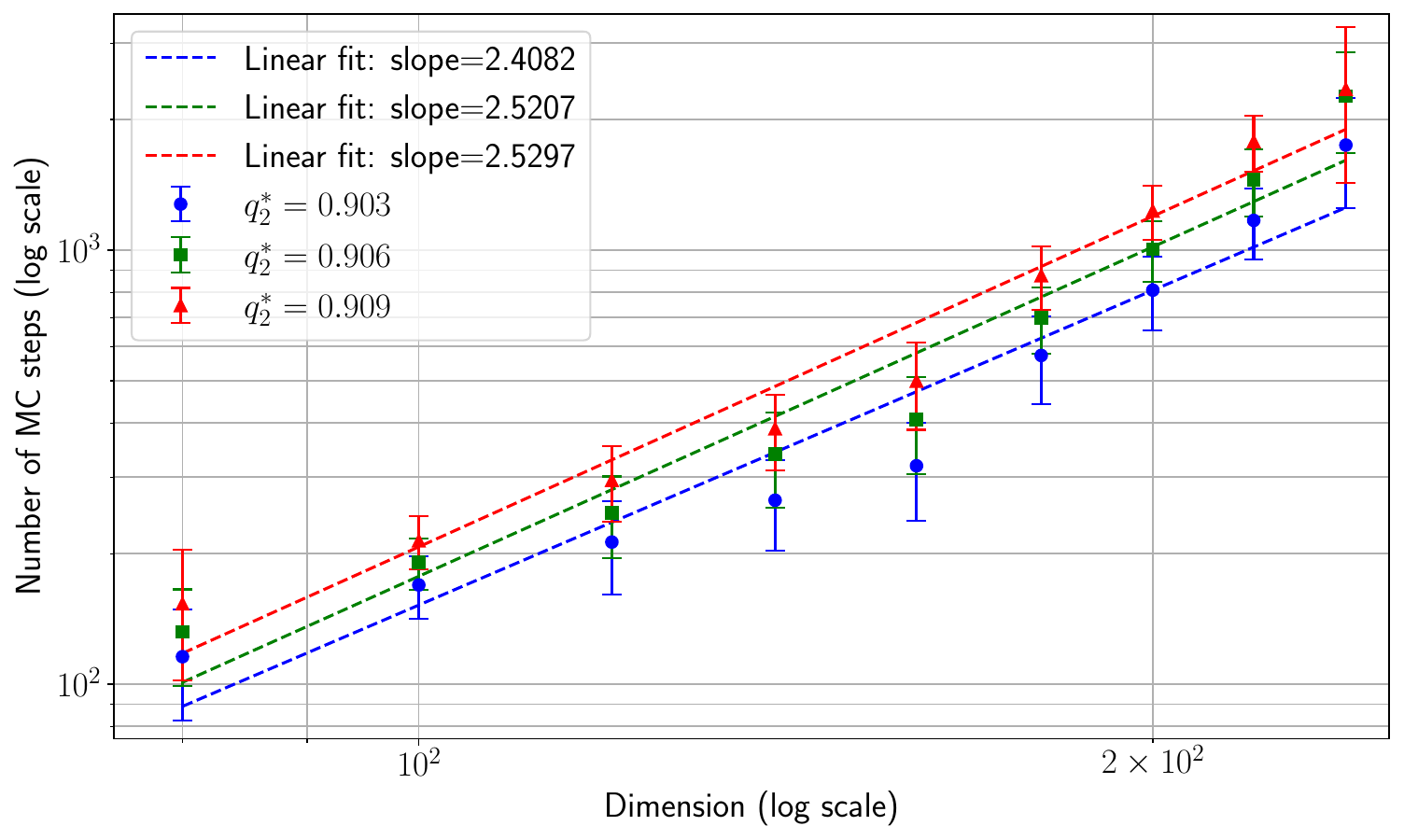}
}
\vspace{1pt} 
\centerline{
\includegraphics[width=.49\linewidth,trim={0 0 0 0},clip]{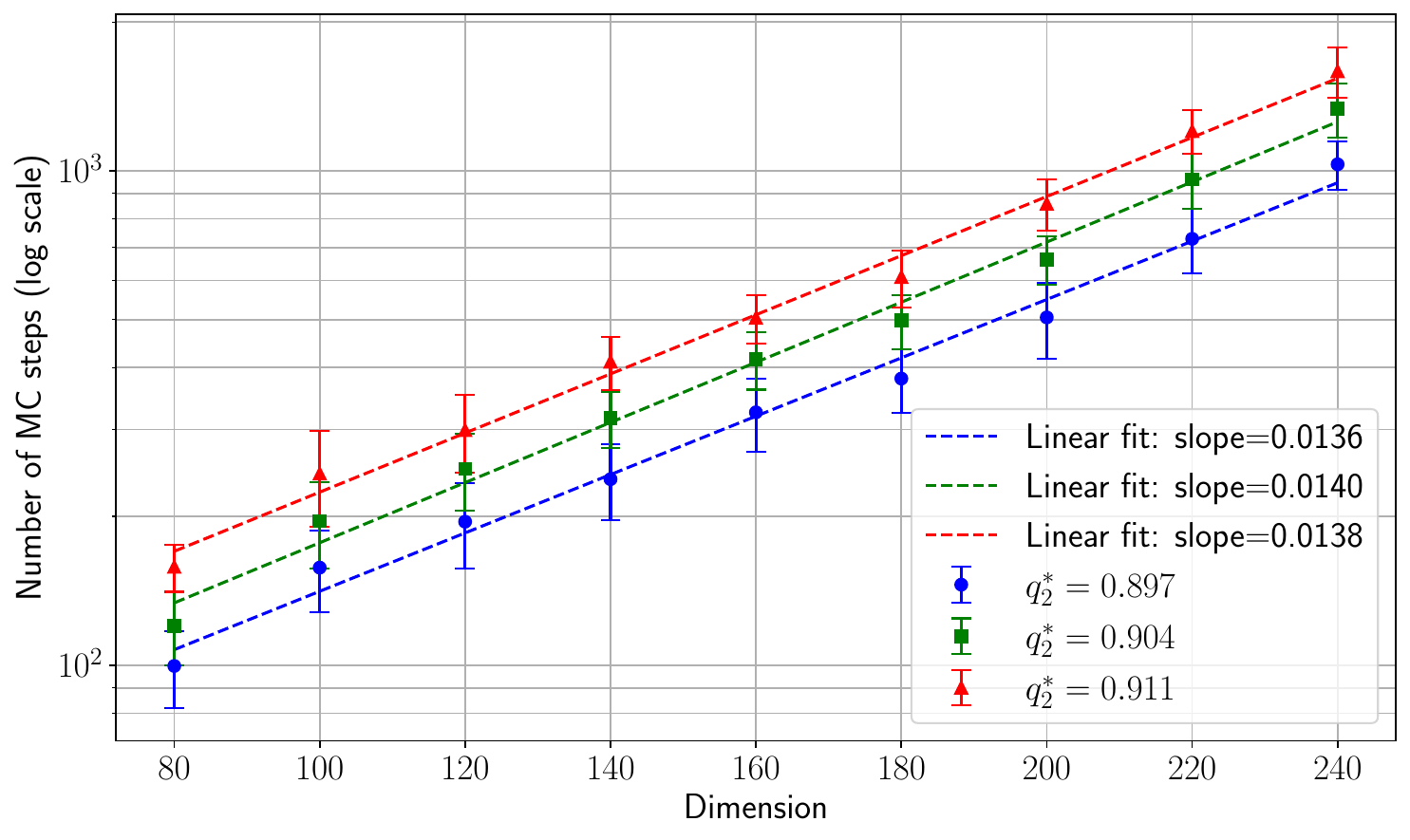}
\includegraphics[width=.49\linewidth,trim={0 0 0 0},clip]{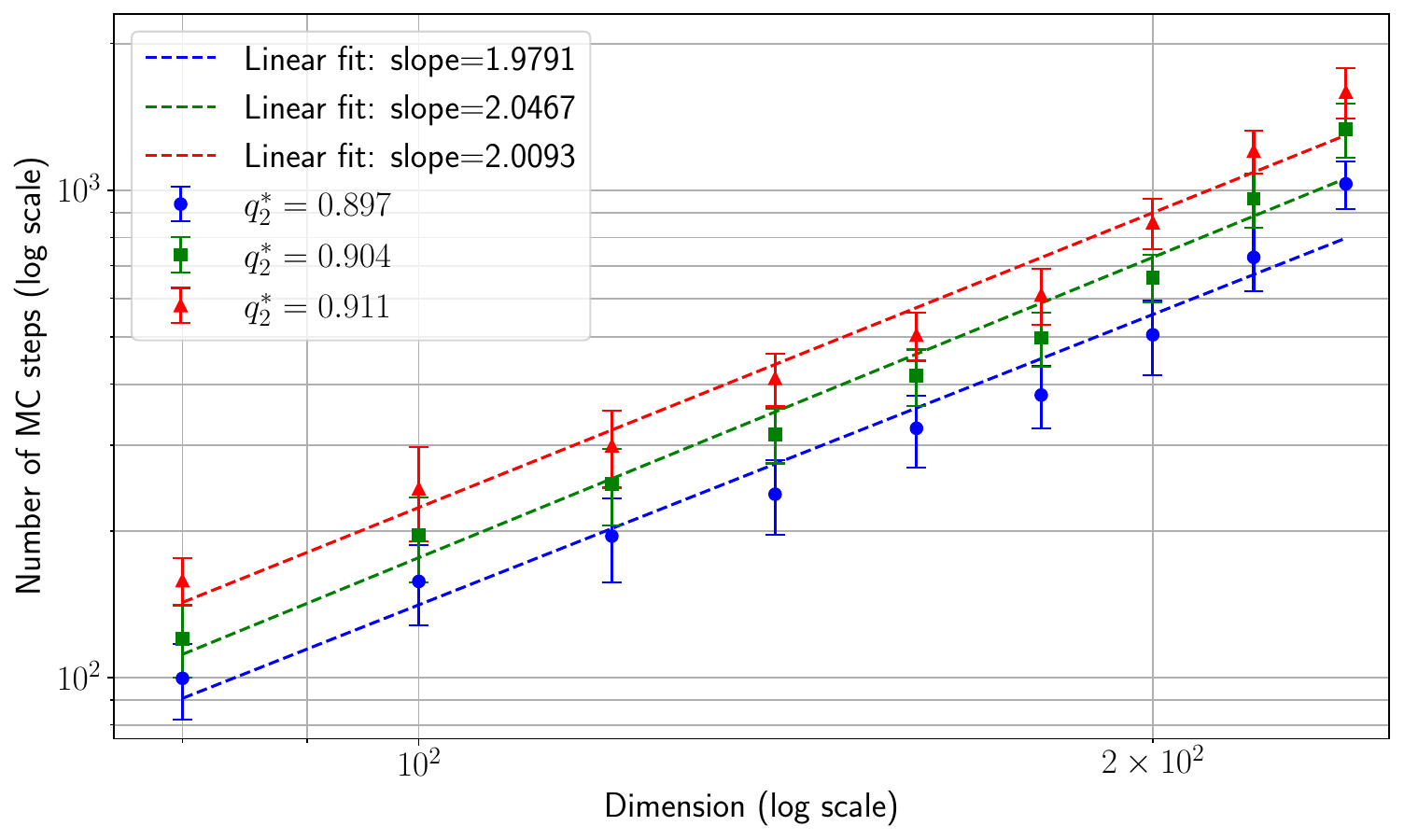}
}
\vspace{1pt} 
\centerline{
\includegraphics[width=.49\linewidth,trim={0 0 0 0},clip]{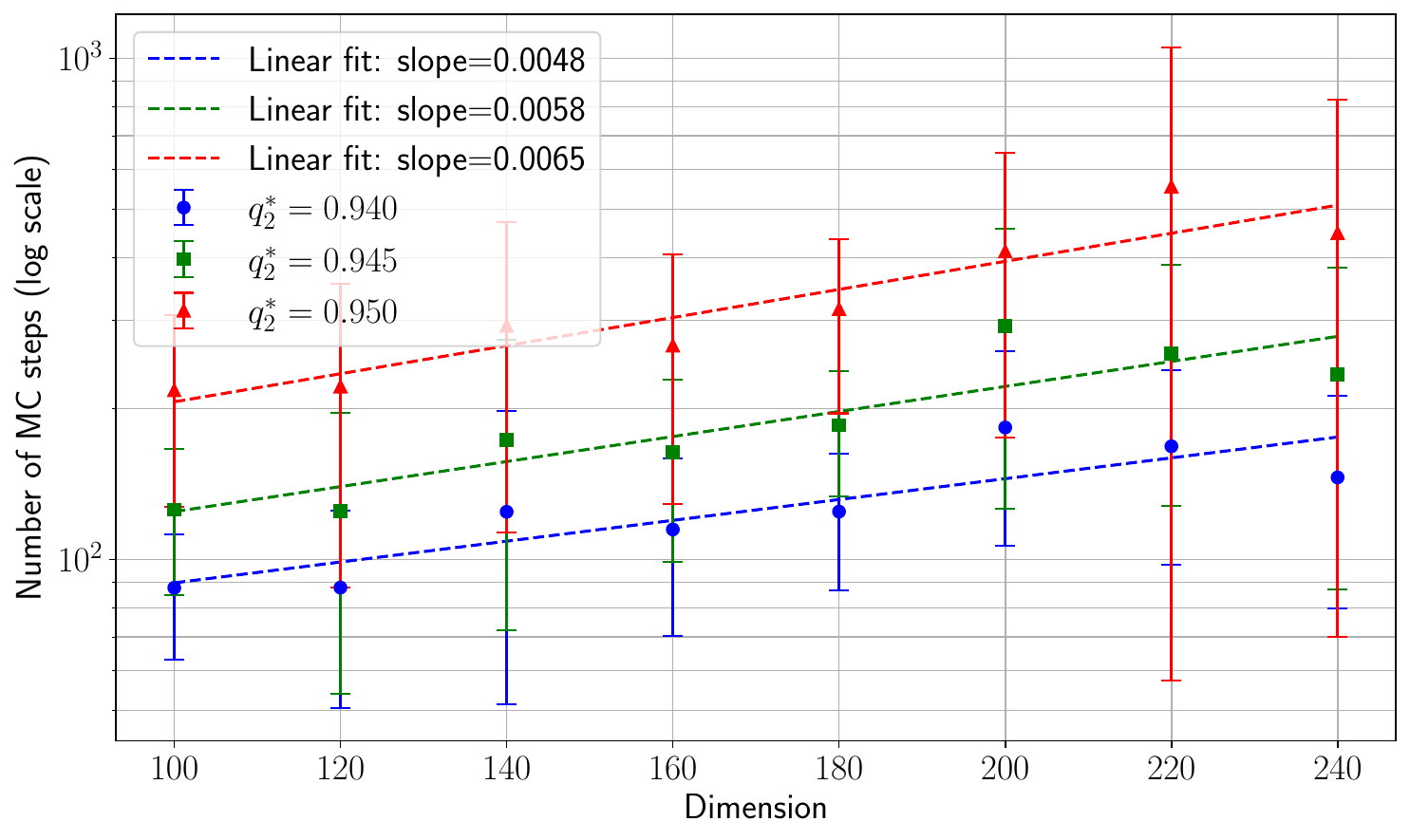}
\includegraphics[width=.49\linewidth,trim={0 0 0 0},clip]{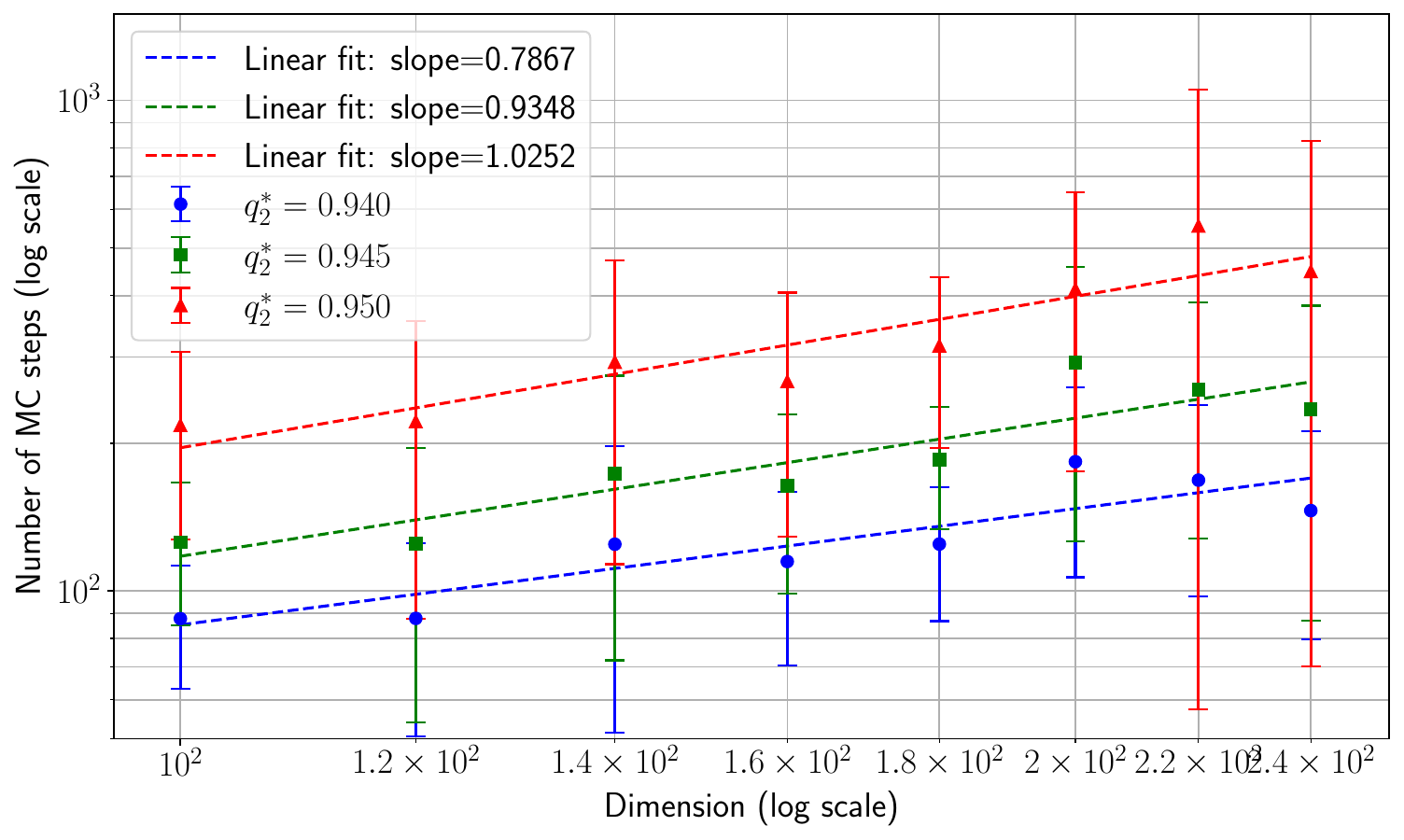}
}
\vspace{-10pt}
    \caption{
    Semilog (\textbf{Left}) and log-log (\textbf{Right}) plots of the number of Hamiltonian Monte Carlo steps needed to achieve an overlap $q_2^*>q_2^{\rm uni}$, that certifies the universal solution is outperformed. 
    The dataset was generated from a teacher with polynomial activation $\sigma_3(x) = \He_2(x)/\sqrt 2 + \He_3(x)/6$ and parameters $\Delta = 0.1$, $\gamma=0.5$ and $\alpha=1.0>\alpha_{\rm sp}$ ($=0.790,0.678,0.933$ for constant, Rademacher and Gaussian read-outs respectively).
    Student weights sampled using HMC with $4000$ iterations for constant read-outs (\textbf{Top row}, for which $q_2^{\rm uni}=0.883$), or $2000$ iterations for Rademacher (\textbf{Center row}, with $q_2^{\rm uni}=0.868$) and Gaussian read-outs (\textbf{Bottom row}, for which $q_2^{\rm uni}=0.903$). Each iteration is adaptative (with initial step size of $0.01$) and uses $10$ leapfrog steps. $q_2^{\rm sp}=0.941$ in the three cases. The read-outs are kept fixed during training. 
    Points are obtained averaging over 10 teacher instances with error bars representing the standard deviation.
    }
    \label{fig:hardness_HMC}
\end{center}
\vskip -0.3in
\end{figure}

\begin{figure}[t!!!]
    \centering
    \includegraphics[width=0.34\linewidth,trim={0.3cm 0 0.3cm 0},clip]{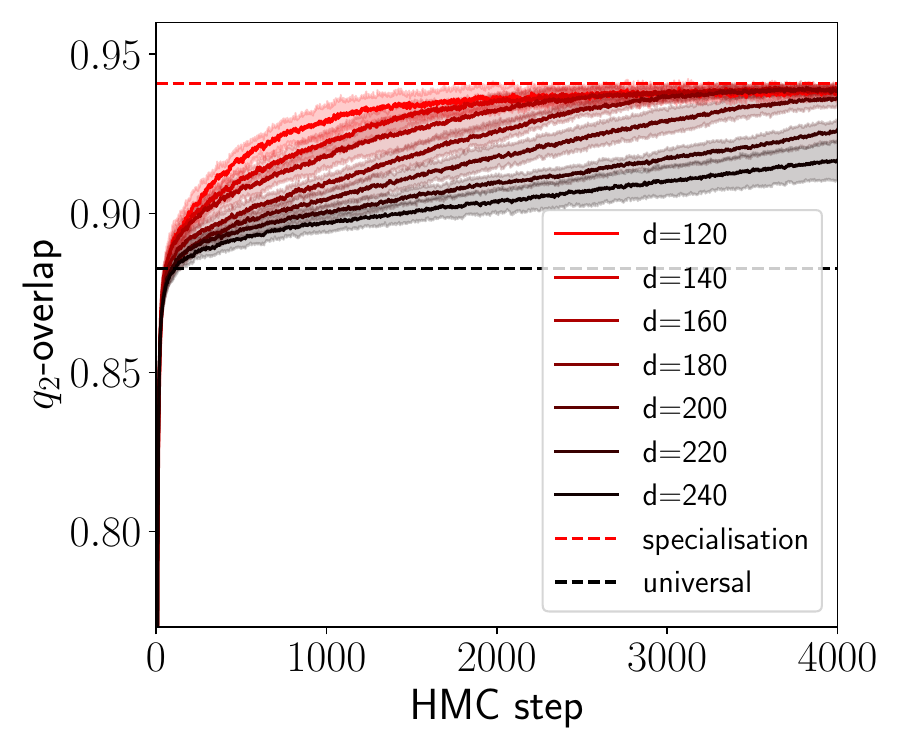}
    \includegraphics[width=0.325\linewidth,trim={1.1cm 0 0.3cm 0},clip]{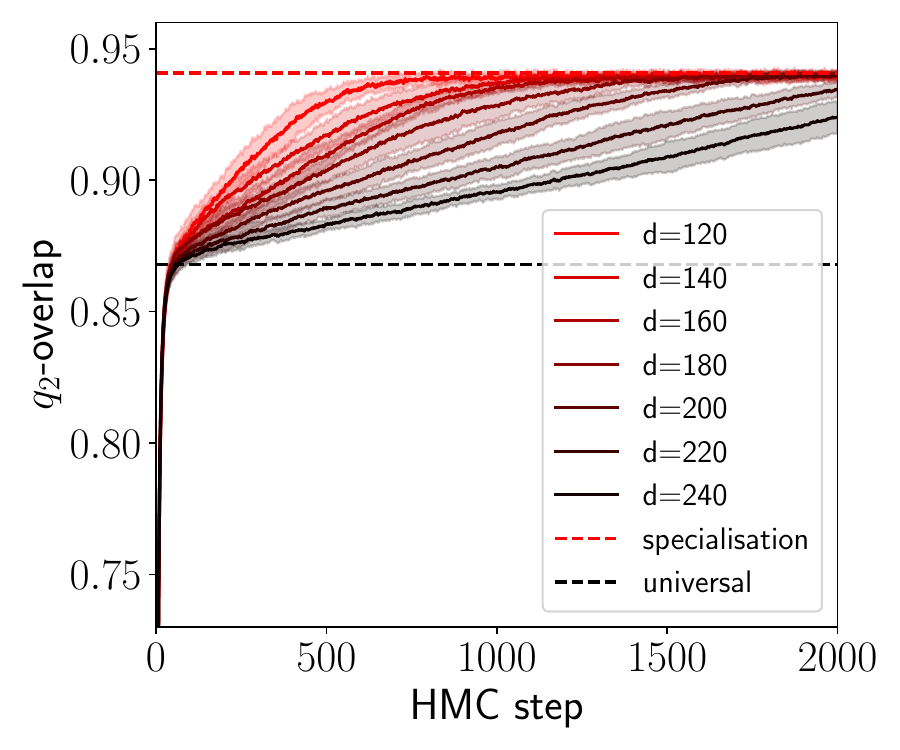}
    \includegraphics[width=0.325\linewidth,trim={1.1cm 0 0.3cm 0},clip]{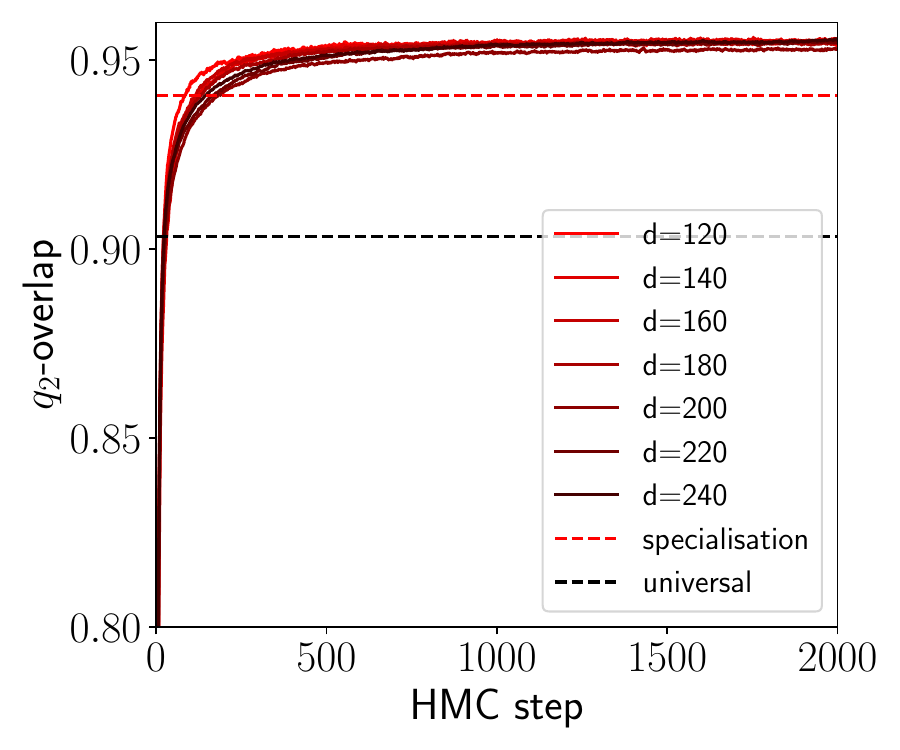}
    \caption{Trajectories of the overlap $q_2$ in HMC runs for the polynomial activation $\sigma_3(x)= \He_2(x)/\sqrt 2 + \He_3(x)/6$ with parameters $\Delta = 0.1$, $\gamma=0.5$ and $\alpha=1.0 > \alpha_{\rm sp}$ ($=0.790,0.678,0.933$ for constant, Rademacher and Gaussian read-outs respectively), as explained in the text.
    \textbf{Left}: Constant read-outs.
    \textbf{Centre}: Rademacher read-outs.
    \textbf{Right}: Gaussian read-outs. Read-outs are kept fixed (and equal to the teacher's ones) in all cases during training. Points on the solid lines are obtained by averaging over 10 teacher instances, and shaded regions around them correspond to one standard deviation.
    Notice that the $y$-axes are limited for better visualisation. For the left and centre plot, any threshold (horizontal line in the plot) between the universal and specialisation value for $q_2$ crosses the curves in points $t^*(d)$ more compatible with an exponential fit (see Fig.~\ref{fig:hardness_HMC} and Table~\ref{tab:HMC}, where these fits are reported and $\chi^2$-tested). For the cases of constant and Rademacher read-outs, both the value of the overlap at which the dynamics slows down (predicted by the universal branch) and the one at which the runs ultimately converge (predicted, for this choice of control parameters, by the specialisation branch) are in quantitative agreement with the theoretical predictions (horizontal lines, left and centre panels). The prediction of $q_2^{\rm sp}$ is off by $\approx 1 \%$ in the case of Gaussian read-outs (right panel).
    }
    \label{fig:hardness_HMC_runs}
\end{figure}

As before, we test constant, Rademacher and Gaussian read-outs, getting to the same conclusions: while for constant and Rademacher read-outs exponential time is more compatible with the observations, the experiments remain inconclusive for Gaussian read-outs (see Fig.~\ref{fig:hardness_HMC}). We report in Fig.~\ref{fig:hardness_HMC_runs} the values of the overlap $q_2$ measured along the HMC runs for different dimensions. While constant and Rademacher read-outs, both more compatible with an exponential fit, converge sharply to the overlap predicted by the specialisation solution, the Gaussian case is off by $\approx 1\%$. Whether this is a finite size effect (we did observe that simulations with continuous readout weights exhibit larger fluctuations), or an effect not taken into account by the current theory is an interesting question requiring further investigation.

\begin{table}[pb]
    \centering
    \begin{tabular}{lc|c|c|c|c|c|c|}
            &        & \multicolumn{3}{c|}{$\chi^2$ exponential fit} & \multicolumn{3}{c|}{$\chi^2$ power law fit}\\
        Read-outs    & & \multicolumn{3}{c|}{} & \multicolumn{3}{c|}{} \\
        \hline
         Constant & ($q_2^*\in \{ 0.903, 0.906, 0.909\}$) &   $\bm{2.22}$ & $\bm{1.47}$ & $\bm{1.14}$ &$8.01$ &$7.25$ &$6.35$  \\
         Rademacher& ($q_2^*\in \{0.897 ,0.904 ,0.911 \}$)& $\bm{1.88}$ & $\bm{2.12}$ & $\bm{1.70}$ &$8.10$ &$7.70$ &$8.57$ \\
         Gaussian& ($q_2^*\in \{0.940 ,0.945 ,0.950 \}$)& $0.66$&$\bm{0.44}$&$\bm{0.26}$  & $\bm{0.62}$&$0.53$ &$0.39$\\
    \end{tabular}
    \caption{$\chi^2$ test for exponential and power-law fits for the time needed by Hamiltonian Monte Carlo to reach the thresholds $q_2^*$, for various priors on the read-outs. For a given row, we report three values of the $\chi^2$ test per hypothesis, corresponding with the thresholds $q_2^*$ on the left, in the order given. Fits are displayed in \figurename~\ref{fig:hardness_HMC}. Smaller values of $\chi^2$ (in bold, for given threshold and read-outs) indicate a better compatibility with the hypothesis.}
    \label{tab:HMC}
\end{table}

\end{document}